\documentclass[runningheads]{llncs}

\usepackage[mobile]{eccv}

\usepackage{eccvabbrv}

\usepackage{graphicx}
\usepackage{booktabs}
\usepackage{wrapfig}
\usepackage{tabularx} %
\usepackage{multirow} %
\usepackage{pgfplots} %
\pgfplotsset{compat=1.18}

\usepackage{siunitx} %
\usepackage[table]{xcolor} %
\usepackage{pifont}

\usepackage[accsupp]{axessibility}  %
\usepackage{adjustbox}

\usepackage{hyperref}

\usepackage{orcidlink}

\newcommand{\hobs}[1][\text{H}]{\mathbf{X}_{#1}}

\newcommand{\fobs}[1][\text{T}]{\mathbf{X}_{#1}}

\newcommand{\xmark}{\ding{55}}

\newcommand{\ourmodel}{\textsc{TriO}}
\newcommand{\ourimagelabels}{\textsc{ISL}}
\newcommand{\ourcombinedlabels}{\textsc{CSL}}
\newcommand{\nameourdataset}{\textsc{LRR}}
\newcommand{\nameourdatasetfull}{\textit{Long Range RADAR}}

\definecolor{tablecolor}{rgb}{0.9, 0.9, 1.0} 

\newcommand{\histcam}{\mathbf{C}_{\text{H}}}
\newcommand{\futcam}{\mathbf{C}_{\text{T}}}

\newcommand{\histlidar}{\mathbf{L}_{\text{H}}}
\newcommand{\futlidar}{\mathbf{L}_{\text{T}}}

\newcommand{\histradar}{\mathbf{R}_{\text{H}}}
\newcommand{\futradar}{\mathbf{R}_{\text{T}}}

\newcommand{\fcam}{f_{\text{c}}} %
\newcommand{\flidarradar}{f_{\text{lr}}} %

\newcommand{\zc}{\mathbf{Z}_{\text{c}}}     %
\newcommand{\zlr}{\mathbf{Z}_{\text{lr}}}     %

\newcommand{\occloss}{\mathcal{L}_{\text{occ}}}
\newcommand{\flowloss}{\mathcal{L}_{\text{flow}}}
\newcommand{\groundloss}{\mathcal{L}_{\text{seg}}}
\newcommand{\lidarloss}{\mathcal{L}_{\text{lidar}}}

\definecolor{groundblue}{RGB}{31,119,180}
\definecolor{otherorange}{RGB}{255,127,14}
\usepackage{printlen}

\begin{document}

\title{TriO: Tri-Modal Unsupervised Occupancy \\ World Model for Anything Perception} 

\titlerunning{TriO: Tri-Modal Unsupervised Occupancy World Model}

\author{Quinlan Sykora*\inst{1, 2} \and
Sourav Biswas*\inst{1, 2} \and
Christopher Diehl*\inst{1} \and \\
Andrew Cunningham\inst{1} \and
Thomas Gilles\inst{1} \and
Raquel Urtasun\inst{1, 2}}

\authorrunning{Q.~Sykora et al.}

\institute{Waabi \and
University of Toronto,
\email{\{qsykora,sbiswas,cdiehl,acunningham,tgilles,urtasun\}@waabi.ai}\\
} 
\def\thefootnote{*}\footnotetext{denotes equal contribution}\def\thefootnote{\arabic{footnote}}

\maketitle

\begin{abstract}
We present TriO, a multi-modal unsupervised world model that predicts 4D occupancy, obstacle segmentation, flow and LiDAR. In contrast to prior work, TriO utilizes three distinct sensor modalities (camera, LiDAR, and RADAR) as both inputs and sources of self-supervision, eliminating the need for additional human annotations. Thanks to its novel supervision, the model is able to segment any occupancy from the drivable surface, overcoming the limitations of existing open-set methods in handling long-tail objects. TriO achieves state-of-the-art results in multiple 3D and 4D tasks, including occupancy, flow, and LiDAR prediction, as well as zero-shot road obstacle segmentation across multiple datasets such as Argoverse 2, and Spotting the Unexpected. 
\end{abstract}

\begin{figure}[ht]
    \centering
    \includegraphics[trim=2cm 2.2cm 0cm 1.2cm, clip, width=1.0\textwidth]{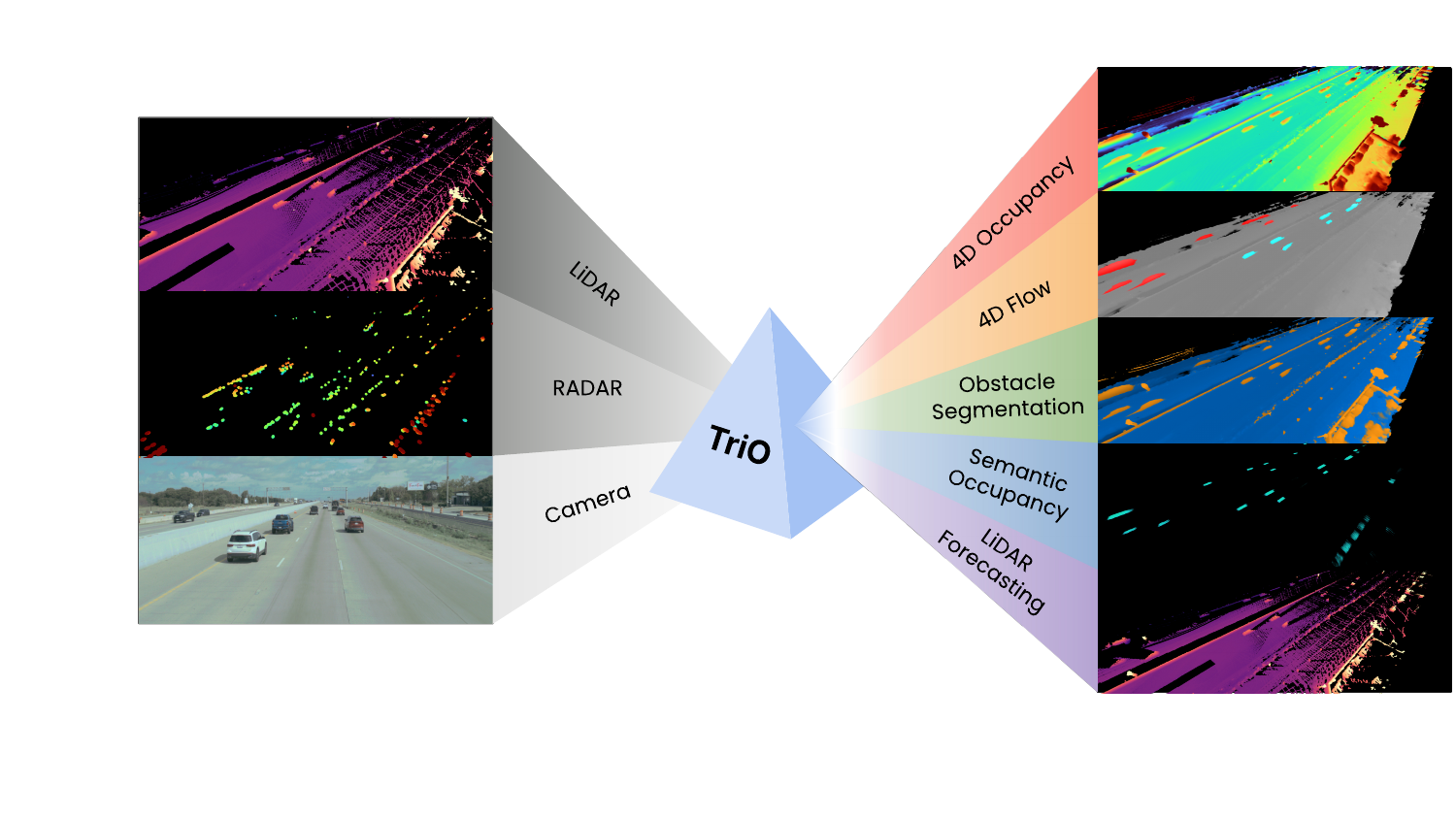}
    \caption{We present \ourmodel{}, an unsupervised tri-modal occupancy world model designed to perceive and forecast anything that could be considered path-blocking.}
\end{figure}

\section{Introduction}
\label{sec:intro}
In order for a self-driving vehicle (SDV) to safely navigate through complex environments, it must first \textit{perceive anything}, i.e. segment above ground obstacles from drivable areas at the current time, then forecast the evolution of the scene to plan its future motion. This problem has been predominantly tackled with object-based approaches \cite{casas2018intentnet,Liang_2020_CVPR,weng2020ptp,weng20203d,salzmann2020trajectron,ivanovic2018the,ivanovic2020multimodal,cui2021lookout,diehl2023corl, DeTRA2024}, which first detect objects and then forecast their future trajectories. In contrast, object-free occupancy approaches \cite{mahjourian2022occupancy,casas2021mp3,sadat2020perceive,philion2020lift,hu2021fiery,agro2023implicit,diehl2023RAL} overcome the downsides of thresholding detections and the limited representation capability of a finite set of trajectories.

However, most 3D/4D occupancy approaches \cite{gu2024dometamingdiffusionmodel, ma2024cvpr, zheng2024eccv, guo2024fsfnetenhance4doccupancy, Zuo_2025_CVPR} rely on dense voxel grid labels \cite{NEURIPS2023_Occ3D, OpenOccupancy_2023_ICCV} which are computationally expensive at high resolutions, and are constructed with human annotations (e.g., semantic LiDAR point annotations or bounding box labels). Due to the cost of the latter, and their difficulty to generate at scale, the volume of the labeled data is commonly orders of magnitude smaller than the amount of unlabeled data \cite{Agro_2024_CVPR}. Moreover, the mentioned approaches are usually supervised with a finite set of semantic classes, which do not encapsulate safety-critical rare events.

Conversely, recent occupancy approaches \cite{khurana2023point, Agro_2024_CVPR, Diehl_2025_DIO} leverage LiDAR self-supervision, pre-training on vast amounts of sensor data and fine-tuning with a small amount of labeled data for downstream applications. However, LiDAR is susceptible to blooming \cite{lidar_weaknesses}, multipath returns \cite{rife2025lidar}, and limited range. Additional sensor modalities can be of benefit for self-supervised training, and multi-sensor fusion can increase robustness to sensor failures and adverse conditions.

Other methods have leveraged RGB cameras, either as inputs \cite{Palladin_2025_ICCV}, distilling foundation model features \cite{GASP_2025}, or predicting semantic pseudo-labels \cite{QueryOcc_2025, SelfOcc_24, Li_2025_ICCV, ShelfOcc2025,  GaussianOcc_Gan_2025_ICCV, Boeder_2025_ICCV} from camera-based models. 
Furthermore, many prior works \cite{maskclippp25, FCCLIP_NIPS23, MAFTp_ECCV2024,groundedsam24,DinoText_2025_CVPR} generate semantic labels with open-vocabulary segmentation models queried with text-prompts. However, common failure modes have been shown in open-set detectors \cite{anonymous2026fantastic} and segmentors \cite{Saric_2025_ICCV}, challenging their overall robustness to rare classes. Finally, the segmentation models  require image-text pair annotations.

One generally overlooked modality is RADAR, which can enhance training by introducing radial velocity measurements, improve robustness under adverse weather conditions \cite{radar_in_rain}, and increase the observation range. While some works investigate RADAR-based 3D occupancy models using human annotations \cite{RadarOcc2024, gaussianfusionoccseamlesssensorfusion}, and others focus on RADAR-only input with self-supervision \cite{Kung_2025_ICCV} or cross-modal LiDAR supervision \cite{4DRolls_IROS2025}, none integrate all aforementioned modalities in a unsupervised tri-modal 4D world model. 

Motivated by these downsides, our goal is to develop a model that can leverage all three common sensor modalities (LiDAR, camera, and RADAR) as both input and supervision for occupancy in a self-supervised manner. Towards this goal, we propose \ourmodel: a multi-modal world model that can predict and forecast 4D continuous occupancy, obstacle segmentation, and flow at any continuous $(x,y,z,t)$ coordinate from tri-modal observations. To achieve this, we leverage each sensor as both an input and as a unique form of supervision, ultimately combining the strengths of each modality to produce the best possible representation of the world. We further introduce a novel multi-sensor fusion architecture leading to  better  forecasting capabilities. 
Notably, being pre-trained on a vast amount of raw sensor data, \ourmodel{} can be effectively leveraged for multiple downstream applications. 

Historically, to determine which entities in the scene are relevant to the self-driving task, prior work uses semantic supervision from open-vocabulary models prompted with a closed set of multiple classes \cite{maskclippp25, FCCLIP_NIPS23, MAFTp_ECCV2024,groundedsam24,DinoText_2025_CVPR}. These however do not generalize to long tail events with rare obstacles, and hence do not cover all safety-critical scenarios. We argue the most general and robust requirement for occupancy is to semantically distinguish between what is \textit{occupied but not path-blocking} (e.g., a flat surface that is traversable by the SDV) vs. \textit{occupied and path-blocking} (e.g., a rare object on the road). Towards this end, we define the task of \textbf{anything-perception} as the ability to segment any object that exists above the road from the ground surface. 

Despite demonstrating impressive open-set detection capabilities, the segmentation models of prior work used for 2D $\rightarrow$ 3D label distillation struggle to segment rare objects \cite{Saric_2025_ICCV} (the long tail), even when prompted with the ground truth class, highlighting a lack of semantic understanding and generalization in current VLMs. To circumvent this, we propose a simple yet effective unsupervised\footnote[1]{We note that Metric3Dv2 \cite{hu2024metric3dv2} was partially trained on datasets with semi-supervised human annotations. In this context, we define \textit{unsupervised} to mean not requiring direct human annotations. For more discussion refer to the supplementary.} \textit{geometric image/LiDAR segmentation} method that leverages representations from image depth model Metric3Dv2 \cite{hu2024metric3dv2}, and the LiDAR segmentation heuristic from Patchwork++\cite{patchworkpp}. Our method detects any obstruction, leading to significant improvements in generalizable unknown object segmentation. Notably, in contrast to other vision-language based approaches which require at least textual image descriptions \cite{groundedsam24, Talk2Dino_2025_ICCV, DinoText_2025_CVPR, FCCLIP_NIPS23, siglip2_2025} or (semantic) mask labels \cite{Kirillov_2023_ICCV, maskclippp25, FCCLIP_NIPS23, MAFTp_ECCV2024}, our obstacle segmentation model relies solely on geometric properties rather than semantic annotations. \\

\textbf{Contributions:} \textbf{1)} \ourmodel{} uniquely uses all three common sensor modalities (camera, LiDAR, RADAR), both as input and for self-supervision. \textbf{2)} We introduce a novel image/LiDAR-based obstacle segmentation technique that generates pseudo-labels for distillation into \ourmodel{}, without requiring additional human labels in our training pipeline, while maintaining robustness and generalizability. \textbf{3)} We propose a novel architecture fusing all modalities leading to better long-term forecasting.
\textbf{4)} \ourmodel{} achieves state-of-the-art performance on a variety of tasks including point cloud forecasting, obstacle segmentation, and both 4D/3D occupancy and flow forecasting, validated on three real-world autonomous driving datasets (\textit{Argoverse2}, \textit{Spotting the Unexpected}, \nameourdatasetfull{}). Overall, the novelty of our contribution is %
the unique integration of all three sensor modalities as inputs, and as source of self-supervision for our output representations.

\section{Related Work}
\label{seq:related_work} 
\textbf{Semantic Occupancy.} 
The task of \textit{scene completion} \cite{LMP2006, PossionSurface2006, PCN2018, Zimmermann2017} involves reconstructing a complete scene from sparse sensor measurements. \textit{Semantic scene completion} \cite{sscIJCV2021, ScanNet2017, yan2021sparse, SemanticKITTI} further extends this by aiming to infer semantic classes for each completed region, a task closely related to \textit{semantic occupancy prediction}.  Contemporary methods in this domain \cite{gu2024dometamingdiffusionmodel, ma2024cvpr, zheng2024eccv, guo2024fsfnetenhance4doccupancy, Zuo_2025_CVPR} typically focus on predicting 3D or 4D occupancy voxel grids with their associated semantic labels. These approaches generally require dense voxel-level supervision \cite{NEURIPS2023_Occ3D, OpenOccupancy_2023_ICCV}, which is often derived from costly human-annotated labels, such as 3D bounding boxes or semantic point clouds. More recently, several approaches \cite{MonoScene, OccNerf23, QueryOcc_2025, Boeder_2025_ICCV} have proposed distilling 2D pseudo-semantic labels from VLMs. However, these VLMs themselves rely on large-scale human annotations in the form of image-text pairs, and supervising with a closed set of classes inherently restricts generalization in open-world scenarios. 
\\
\textbf{Anything Perception.} %
Open-vocabulary segmentation methods perform classification via text prompts at inference time, overcoming the limitations of fixed, closed-set label spaces.  In the 2D domain, two primary paradigms have emerged: one leverages open-set detectors like GroundingDINO \cite{groundingdino2024eccv} or OWLv2 \cite{OWLv2neurips23} to prompt mask predictors such as SAM \cite{Kirillov_2023_ICCV}, while another \cite{FCCLIP_NIPS23, MAFTp_ECCV2024, maskclippp25} predicts mask proposals and classifies them by computing the similarity between mask embeddings and CLIP-style \cite{CLIP_2021} text embeddings. 
These paradigms have been extended to the 3D domain through methods like SAL \cite{sal2024eccv}, which distills CLIP features for LiDAR segmentation. Similarly, recent open-vocabulary occupancy methods \cite{wang2024distillnerf, zhao2025shelfgaussian, GausTR_2025_CVPR, Li_2025_ICCV} distill these features into voxel representations to enable open-world prompting. 
Despite their flexibility, these methods face two critical challenges: (1) they remain fundamentally dependent on models pre-trained on vast image-text and per-pixel mask annotations, and (2) they exhibit significant performance degradation for rare classes \cite{Saric_2025_ICCV, anonymous2026fantastic}. 
In contrast, \ourmodel{}s supervision relies on geometric cues from different sensors, enabling a more reliable segmentation of any object relative to the drivable surface. 
While ground segmentation has been extensively studied for LiDAR point clouds \cite{GndNet2020, SurveyLiDAR_ground_segmentation, patchworkpp, GroundGrid2024}, its application to occupancy remains limited. For example, \cite{Dual_Evidental_TITV__24} utilizes surface normals from stereo cameras and LiDAR to estimate ground regions \textit{in current time BEV (2D)} occupancy maps, yet is constrained by the range of current measurements and segmentation networks that require pixel-level human annotations. In comparison, we distill fused obstacle segmentation estimates from LiDAR and monocular camera data and \ourmodel{} has a unique ability of \textit{forecasting these segmentations in 4D}.\\
\textbf{Unsupervised Occupancy.}
Early non-learning-based methods \cite{DOGMA_ICRA14, Nuss2017, Diehl2020} employed classical filtering techniques to estimate occupancy and flow without human annotations. With the advent of deep learning, 4D-Occ \cite{khurana2023point} proposed a LiDAR-only self-supervised approach to forecast 4D occupancy. However, its explicit 4D voxel grid representation suffers from significant memory and runtime inefficiencies, often forcing a trade-off between spatial resolution and reconstruction quality. To address these limitations, recent implicit occupancy methods \cite{Agro_2024_CVPR, GASP_2025, Diehl_2025_DIO, QueryOcc_2025, Palladin_2025_ICCV} represent scenes in continuous space-time, leveraging self-supervision from LiDAR or camera inputs. Departing from implicit representations, SceneDino \cite{Jevtic_2025_ICCV} predicts 3D geometry and feature fields from monocular input, achieving semantic scene completion through clustering and unsupervised distillation. Other recent works, such as \cite{4DRolls_IROS2025}, adopt cross-modal strategies using RADAR as input with LiDAR-based supervision. In contrast to all prior work, \ourmodel{} utilizes camera, LiDAR, and RADAR data for both input and self-supervision, exploiting the inherent physical synergies between sensors to perform 4D forecasting of occupancy, segmentation, and flow. We show an extensive tabular comparison and further related works in the supplementary.

\section{Unsupervised Multi-Modal Occupancy-Flow Method}
\label{sec:method}

To learn a robust representation of the world, we leverage the complementary physical properties of the standard SDV sensor suite: active sensors (LiDAR and RADAR) operating on Time-of-Flight principles, and passive cameras. While LiDAR provides high-precision 3D geometry capable of capturing free-space \cite{Agro_2024_CVPR} and object surfaces \cite{patchworkpp}, its sparse returns degrade at distance and in adverse weather. In contrast, RADAR uses longer wavelengths that penetrate fog and rain, and provide instantaneous radial velocity via the Doppler effect. Passive RGB cameras offer dense, high-resolution semantic and textural information, but lack native depth. To overcome this limitation, we extract dense depth and surface normal estimates using recent vision foundation models \cite{hu2024metric3dv2}, enabling the reliable segmentation of obstacles. 
In summary, we combine the complementary synergies of each sensor modality and propose \ourmodel{}, a 4D world model that learns geometry, dynamics, and object segmentation while remaining unsupervised.

\subsection{Unsupervised Task}
\label{sec:method_unsupervised_task}

To define our training objective, let $T \in \mathbb{N}$ denote the future time-horizon and $H \in \mathbb{N}$ the historical observation time steps. For notational clarity, in the following we define $\mathbf{S}_{\text{H}} = \mathbf{S}_{-H,\dots,0}$ as the historical sensor observations at time $-H,\dots,0$ and $\mathbf{S}_{\text{T}}$ as the future observations at time $0,\dots,T$. Here, $\mathbf{S}$ is a placeholder for either multi-view camera images $\mathbf{C} \in \mathbb{R}^{N_{\text{C}} \times W \times H \times 3}$, LiDAR points $\mathbf{L} \in \mathbb{R}^{N_{\text{L}} \times d_{\text{L}}}$, or RADAR targets $\mathbf{R} \in \mathbb{R}^{N_{\text{R}} \times d_{\text{R}}}$. $N_{\text{C}}$ denotes the number of camera images with width $W$ and height $H$. Further, $N_{\text{L}}$, $N_{\text{R}}$, $d_{\text{L}}$, $d_{\text{R}}$ are the number of LiDAR/RADAR points and their feature dimensionalities, respectively.  Finally, we define the combined historical and future sensor observations as $\hobs = (\histcam, \histlidar, \histradar)$ and $\fobs = (\futcam, \futlidar, \futradar)$, respectively. Our goal is to learn a model:
\begin{equation}
\mathbf{y} = (o, s, \mathbf{f}) = f_{\theta}(\hobs, \mathbf{q}),
\end{equation}
parameterized by $\theta$, that can predict occupancy probability $o \in [0,1]$, obstacle segmentation probability $s \in [0,1]$ and 3D flow $\mathbf{f} \in \mathbb{R}^3$ at \textit{query point} $\mathbf{q} = (q_x, q_y, q_z, q_t)$ in \textit{continuous} 4D space-time ($x, y, z, t$). 

Geometric occupancy and 3D flow offer a simple representation that captures important characteristics of real-world environments (being occupied by a physical object and its velocity). However, a motion planner must distinguish between what is path-blocking (e.g., a child sitting on the street) and what is traversable (e.g., a drivable slope), motivating the prediction of obstacle segmentation $s$. To train our method, we require labels for each query point $\mathbf{q}$. We construct supervisory signals (pseudo-labels) as $\bar{\mathbf{y}} = g(\fobs, \mathbf{q})$, which are used in the loss function $\mathcal{L}(\mathbf{y}, \bar{\mathbf{y}})$.

In the remainder of this section, we first detail the architecture of $f_{\theta}$ in \cref{sec:method_architecture}. We then detail the function $g$ by introducing our segmentation method in \cref{sec:method_ground_seg} and how pseudo-labels are generated from all three sensor modalities in \cref{sec:method_self_supervision}. The training procedure is described in \cref{sec:method_training}.
\subsection{Unsupervised Multi-Modal Occupancy-Flow Model}
\label{sec:method_architecture}
\begin{figure}[t]
    \centering
    \includegraphics[trim=0cm 3cm 0cm 3cm, clip, width=1.0\textwidth]{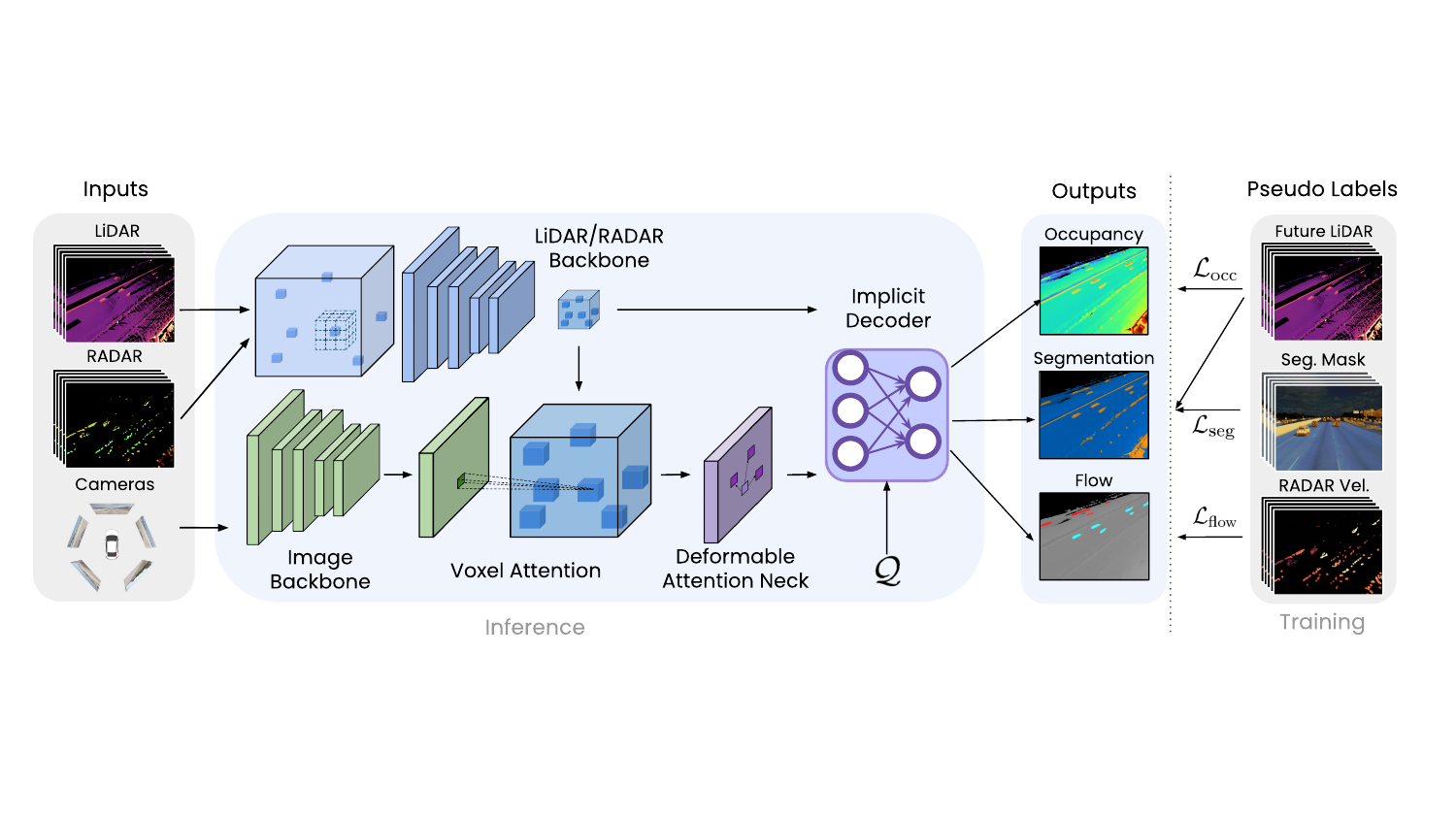}
    \caption{\ourmodel{}'s multi-modal architecture fusing LiDAR, RADAR, and camera inputs via voxel-attention. An implicit decoder generates occupancy, segmentation, and flow for a set of query points $\mathcal{Q}$ supervised by pseudo-labels derived from sensor data.}
    \label{fig:model_architecture}
\end{figure}

Our model follows an encoder-decoder structure and the overall architecture is shown in \cref{fig:model_architecture}. First, we encode the measurements of each sensor with two encoders, one for LiDAR-RADAR ($\flidarradar$) and one for camera ($\fcam$):
\begin{equation}
    \zc = \fcam(\histcam), \quad \zlr = \flidarradar(\histlidar, \histradar).
\end{equation}
The camera encoder $\fcam$ is a pretrained version of ResNet50 \cite{he2016deep}, which only takes in the most recent camera frame. The LiDAR-RADAR encoder discretizes locations at the voxel level of multiple historical frames, and encodes them as features in a 3D sparse feature map. The features from the LiDAR  
$\mathbf{l} = [x, y, z, i, \Delta t]^\intercal \in \mathbb{R}^5$ (including intensity $i$ and relative 
timestep $\Delta t$) and RADAR points 
$\mathbf{r} = [x, y, z, v_r, \sigma, \Delta t]^\intercal \in \mathbb{R}^6$ (including range rate $v_r$, RADAR cross section $\sigma$, and $\Delta t$) are each encoded through separate shallow MLPs.

The latent representations $\zc$ and $\zlr$ are then fused by allowing the sparse LiDAR features at the coarsest level to attend to the image features through a voxel attention mechanism ($f_\text{voxelatt}$), similar to what is done in \cite{Li_2023_CVPR}, giving $\mathcal{Z}_0=[\mathbf{V}_{2x}, \mathbf{V}_{4x}, \mathbf{V}_{8x}, \mathbf{M}_{16x}]$, with feature volumes $\mathbf{V}_{2x}, \mathbf{V}_{4x}, \mathbf{V}_{8x}$ and $\mathbf{M}_{16x}$ having been downsampled by 2x, 4x, 8x, and 16x respectively. We then densify the coarsest feature map $\mathbf{M}_{16x}$ and pass it through several deformable attention layers ($f_\text{defatt}$) to increase the effective receptive field of the model.  Finally, we use the same deformable attention multi-resolution header proposed by \cite{Diehl_2025_DIO} for each of the model's predictions. We describe the specific details in the supplementary. The process is summarized as:
\begin{equation}
    \mathcal{Z}_0 = f_\text{voxelatt}(\zc, \zlr) \rightarrow \mathcal{Z}_1 = f_\text{defatt}(\mathcal{Z}_0) \rightarrow (o, \mathbf{f}, s) = f_\text{dec}(\mathcal{Z}_1, \zlr, \mathbf{q}).
\end{equation}
\subsection{Obstacle Segmentation Model}
\label{sec:method_ground_seg}
\begin{figure}[tbp]
    \centering
    \includegraphics[trim=0cm 4.5cm 0cm 0.5cm, clip, width=1.0\textwidth]{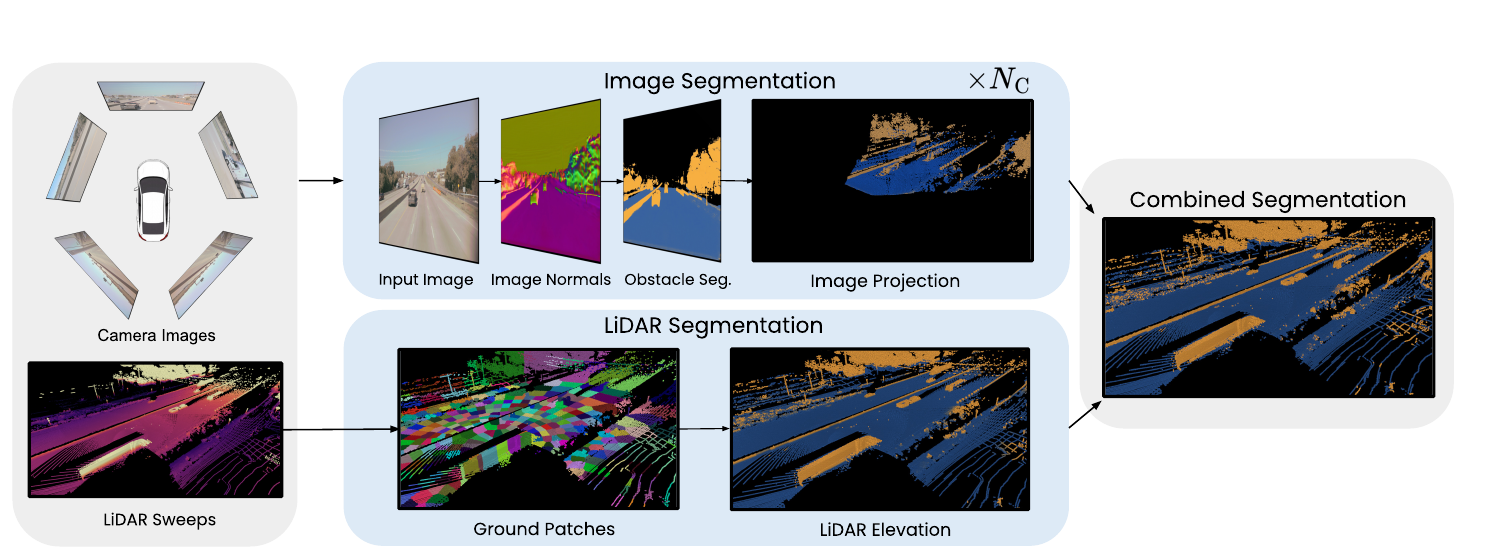}
    \caption{\textcolor{orange}{\textbf{Obstacle}} and \textcolor{blue}{\textbf{ground}} pseudo-label generation process.}
    \label{fig:method_seg_model}
\end{figure}

To provide supervision for obstacle segmentation $s$, we propose a novel method for extracting obstacle/ground segmentation labels from image and LiDAR features in a probabilistic manner, combining the strength of each sensor modality. This approach ensures that the objects which are distinctly above ground are captured by the LiDAR segmentation, which is robust due to the 3D information it receives, while also detecting smaller close-to-the-ground obstacles that may only be observable in image space. The overall process is illustrated in \cref{fig:method_seg_model}. \\
\textbf{Image Obstacle Segmentation.} We propose using an image foundation model to produce estimates of normals and depths of the image for each camera.
Specifically, given a camera image $\mathbf{I} \in \mathbb{R}^{H \times W \times 3}$ 
we apply Metric3Dv2 \cite{hu2024metric3dv2} to obtain per-pixel metric depth $d \in \mathbb{R}^{H \times W}$, depth confidence $c \in \mathbb{R}^{H \times W}$, and surface normals $\hat{\mathbf{n}} \in \mathbb{R}^{H \times W \times 3}$ with associated confidence $c_n \in \mathbb{R}^{H \times W}$. We then follow a similar process to \cite{depth_normal_seg} to produce a segmentation of the obstacles relative to the rest of the scene based on the normals and depth information. In contrast to prior work, which produce discrete labels, we output continuous probabilities for obstacles/ground ($p^{\text{image}}_{\text{obst}}$), in addition to probabilities of the validity of the estimation ($p^{\text{image}}_{\text{valid}}$). Producing these continuous image segmentation labels (\ourimagelabels{}) enables the downstream model to learn a more calibrated representation. \\
\textbf{LiDAR Obstacle Segmentation.} 
We leverage ideas from Patchwork++ \cite{patchworkpp} for LiDAR segmentation to discretize the point cloud into polar patches, where for each patch $j$, a ground plane is characterized by its centroid $\mathbf{c}_j \in \mathbb{R}^3$ and surface normal $\hat{\mathbf{n}}_j \in \mathbb{R}^3$. For a LiDAR return $\mathbf{l} \in \mathbb{R}^3$, we identify its corresponding patch in the $xy$-plane. We then get the elevation $e\in \mathbb{R}$ as the point to plane distance using the patch normal, and map this into continuous probabilities for the obstacle  $p^{\text{LiDAR}}_{\text{obst}}$ and the valid distribution $p^{\text{LiDAR}}_{\text{valid}}$. \\
\textbf{Combined Segmentation.} To leverage the complementary strengths of both modalities, we employ a confidence-driven fusion using the per-point validity and obstacle probabilities $p_{\text{valid}}$, $p_{\text{obst}}$ from both image (projected onto LiDAR) and LiDAR sources. Formally, the combined probability is
\begin{equation}
\begin{aligned}
    \mathbf{p}^{\text{comb}} &= 
    \begin{cases} 
        \mathbf{p}^{\text{image}}, & \text{if } (p^{\text{LiDAR}}_{\text{valid}} < 0.5) \;\wedge\; (p^{\text{image}}_{\text{valid}} > p^{\text{LiDAR}}_{\text{valid}}) \\
        \mathbf{p}^{\text{LiDAR}}, & \text{otherwise}
    \end{cases}
\end{aligned}
\end{equation}
with $\mathbf{p}^{\textsc{source}} = \bigl[p^{\textsc{source}}_{\text{valid}}, p^{\textsc{source}}_{\text{obst}}\bigr]^\intercal$, $\textsc{source} \in \{\text{comb}, \text{image},\, \text{LiDAR}\}$. The probability vector $\mathbf{p}^{\text{comb}}$ describes the combined soft labels (\ourcombinedlabels) to supervise $s$. By prioritizing image labels for low-confidence LiDAR segmentation results, this approach captures small visual details while maintaining LiDAR’s robustness to points that are clearly distinct from the ground plane.
Additional details of our segmentation method are in the supplementary.

\subsection{Multi-Modal Self-Supervision}
\label{sec:method_self_supervision}
We will now outline the pseudo-labels used to supervise \ourmodel{}. These can all be represented by a 4D query point $\mathbf{q}$, and a label $\bar{y}$,
summarized as $\bar{\mathbf{y}} = (\mathbf{q}, \bar{y})$. \\
\textbf{Occupancy.}
We generate dense occupancy pseudo-labels without any human annotation, following the methodology of UnO \cite{Agro_2024_CVPR}, DIO \cite{Diehl_2025_DIO}, and GASP \cite{GASP_2025}.
For a LiDAR return $\mathbf{l}$ we assume the continuous ray segment between the sensor position and $\mathbf{l}$ as the unoccupied region $\mathcal{R}^-$, and a short segment immediately behind $\mathbf{l}$ along the ray direction as the occupied region $\mathcal{R}^+$. The sets of occupied and unoccupied points in the scene are $\mathcal{O}^+ = \bigcup_{\mathbf{l} \in \mathbf{L}} \mathcal{R}^+$ and $\mathcal{O}^- = \bigcup_{\mathbf{l} \in \mathbf{L}} \mathcal{R}^-$, respectively. The set of $2N$ sampled points with occupancy pseudo-labels is
\begin{align}
    \mathcal{Y}_{\text{occ}} &= \{(\mathbf{q}, 1) \mid \mathbf{q} \in \mathcal{O}^+\} \cup \{(\mathbf{q}, 0) \mid \mathbf{q} \in \mathcal{O}^-\}.
\end{align}
\textbf{Obstacle Segmentation.}
In order to train for the segmentation class, we revisit the pseudo-labels generated in \cref{sec:method_ground_seg}. Recall that this formulation provides a combined probability vector $\mathbf{p}^{\text{comb}}$ for each point $\mathbf{r} \in \mathcal{P}_{\text{sweep}}$ in LiDAR sweep $\mathcal{P}_{\text{sweep}}$, containing both the validity and obstacle probabilities. We therefore use the locations of these LiDAR points with their time of emission ($x,y,z,t$) as coordinates for our query $\mathbf{q}$, and their corresponding combined probabilities as the soft labels. The set of points with pseudo-labels $\mathcal{Y}_{\text{seg}}$ is 
\begin{equation}
    \mathcal{Y}_{\text{seg}} = \{(\mathbf{q}, \mathbf{p}^{\text{comb}}(\mathbf{q})) \mid \mathbf{q} \in \mathcal{P}_{\text{sweep}}\}.
\end{equation}
\textbf{Instantaneous Flow.}
In order to train for unsupervised instantaneous flow, we use the Doppler readings from the RADAR sensor as our supervision source. As RADAR points can be noisy, we first define a refined set of RADAR points $\mathcal{R}_{\text{clean}}$ where the measurement uncertainty is below a certain threshold. Due to the inherent characteristic that RADAR only captures velocity in the radial direction, our pseudo-labels must include the sensor velocity at the time of measurement $\mathbf{v}_{\text{sensor}}$, the unit radial direction of the RADAR point $\mathbf{d}_{\text{radar}}$, and the scalar Doppler velocity $v_{\text{doppler}}$. We define the flow pseudo-label set $\mathcal{Y}_{\text{flow}}$ as:
\begin{equation}
    \mathcal{Y}_{\text{flow}} = \{(\mathbf{q}, (v_{\text{doppler}}, \mathbf{v}_{\text{sensor}}, \mathbf{d}_{\text{radar}})) \mid \mathbf{q} \in \mathcal{R}_{\text{clean}}\}.
\end{equation}
\subsection{Training}
\label{sec:method_training}
We train our model with a multi-task loss for occupancy ($\occloss$), flow ($\flowloss$), and obstacle probability ($\groundloss$)
\begin{equation}
    \mathcal{L} = \lambda_1 \occloss + \lambda_2 \flowloss + \lambda_3 \groundloss 
\end{equation}
using the pseudo-labels defined by $\mathcal{Y}_{\text{occ}}$, $\mathcal{Y}_{\text{seg}}$, and $\mathcal{Y}_{\text{flow}}$. Let \ourmodel{}'s occupancy, obstacle probability, and flow output at a query point $\mathbf{q}$ be denoted as $f_{\theta, \text{occ}}(\mathbf{q}) \in [0,1]$, $f_{\theta, \text{seg}}(\mathbf{q})\in [0,1]$, and $f_{\theta, \text{flow}}(\mathbf{q})\in \mathbb{R}^3$ respectively. For notational simplicity, we drop the conditioning on the input $\hobs$.
The occupancy loss is the binary cross-entropy loss over all query points:
\begin{align}
    \occloss &= \frac{1}{|\mathcal{Y}_{\text{occ}}|} \sum_{(\mathbf{q}, \bar{y}_{\text{occ}}) \in \mathcal{Y}_{\text{occ}}} \text{BCE}\left(f_{\theta,\text{occ}}(\mathbf{q}),\; \bar{y}_{\text{occ}}\right).
    \label{eq:occ_loss}
\end{align}
The obstacle probability loss is a soft focal loss (SFocalLoss):
\begin{align}
    \groundloss &= \frac{1}{|\mathcal{Y}_{\text{seg}}|} \sum_{(\mathbf{q}, p^{\text{comb}}(\mathbf{q})) \in \mathcal{Y}_{\text{seg}}} \text{SFocalLoss}\left(f_{\theta,\text{seg}}(\mathbf{q}),\; p^{\text{comb}}_{\text{obst}}(\mathbf{q}),\; p^{\text{comb}}_{\text{valid}}(\mathbf{q})\right).
    \label{eq:ground_loss}
\end{align}
In contrast to $\occloss$, which uses hard labels $\bar{y}_{\text{occ}}\in \{0,1\}$, $\groundloss$ operates on soft targets $p^{\text{comb}}_{\text{obst}} \in [0,1]$, with each point's contribution scaled by its label confidence $p^{\text{comb}}_{\text{valid}}$. We find this soft parameterization to be important for generating a well-calibrated model and elaborate further in \cref{sec:exp}. Full loss definitions are provided in the supplementary material.
Finally, the flow supervision is determined as follows. We begin by moving the model's instantaneous 3D flow prediction $f_{\theta, \text{flow}}(\mathbf{q})$ into the sensor's frame of reference by subtracting the sensor's velocity $\mathbf{v}_{\text{sensor}}$, yielding the relative flow $\mathbf{f}_{\text{rel}}(\mathbf{q}) = f_{\theta, \text{flow}}(\mathbf{q}) - \mathbf{v}_{\text{sensor}}$. We then project this relative flow onto the radial direction vector $\mathbf{d}_{\text{radar}}$ to obtain the predicted radial velocity: $f_{\text{rad}}(\mathbf{q}) = \mathbf{d}_{\text{radar}} \cdot \mathbf{f}_{\text{rel}}(\mathbf{q})$. Finally, we supervise this value using the Doppler measurements with an $L_1$ loss over the cleaned RADAR points $\mathcal{R}_{\text{clean}}$:
\begin{equation}
    \flowloss = \frac{1}{|\mathcal{R}_{\text{clean}}|}\sum_{\mathbf{q} \in \mathcal{R}_{\text{clean}}} \left| f_{\text{rad}}(\mathbf{q}) - v_{\text{doppler}} \right|.
\end{equation}

\section{Experiments}
\label{sec:exp}
In this section, we investigate the following research questions while comparing against the state-of-the-art: \textbf{Q1}: Can \ourmodel{} reliably perceive unknown objects in a zero-shot, open-set manner? \textbf{Q2}: Does \ourmodel{}'s architecture lead to a better 4D geometric representation? \textbf{Q3}: Does unsupervised multi-modal pre-training improve downstream task performance via fine-tuning?\\ 
\textbf{Datasets.} 
We use the \textit{Argoverse 2} (AV2) \cite{wilson2023argoverse} \textit{Sensor} split (700 train/ 150 val sequences)
and evaluate on the AV2 Occupancy Forecasting leaderboard.
The dataset \textit{Spotting the Unexpected} (STU) \cite{STU_cvpr2025} uniquely provides 3D LiDAR point labels for rarely seen objects on the road, aiming to test models on anything perception.
We utilize all 22 available sequences with labels from STU (train and val) for evaluation. Note, STU is only for \textit{zero-shot evaluation} -- i.e. we train \ourmodel{} and baselines on AV2 and apply it without any fine-tuning on STU, containing different sensor characteristics. 
Given the limited label range and no RADAR in previous datasets, we use an unreleased dataset \nameourdatasetfull{} (\nameourdataset{}) where we label bounding boxes up to 350m in front of the SDV. \nameourdataset{} has 2280 sequences (highway and urban). See supplementary for details.

\subsection{Zero-Shot Anything Perception (Q1)}
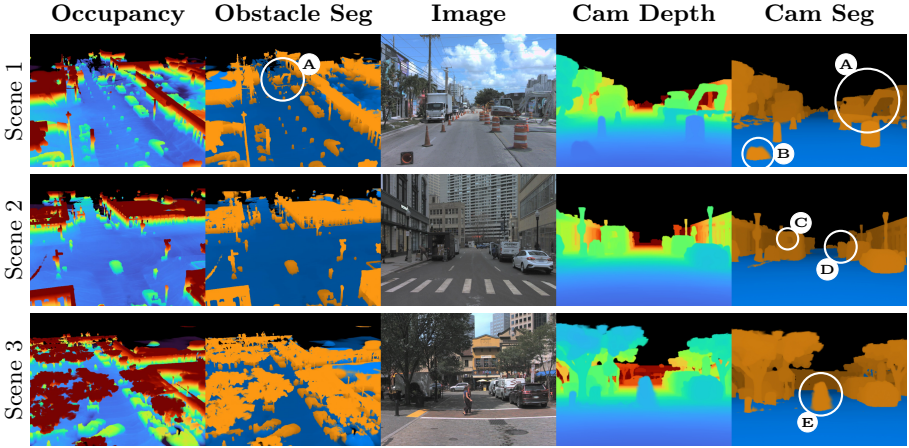
\begin{figure*}[t]
    \centering
    \begin{tikzpicture}
        \pgfmathsetlengthmacro{\imw}{0.95 \linewidth}
        \pgfmathsetlengthmacro{\sep}{1mm}
        \pgfmathsetlengthmacro{\ysep}{1mm}

        \node[inner sep=0pt, outer sep=0, anchor=east] (scene-1) at  (0,0) {\includegraphics[width=\imw]{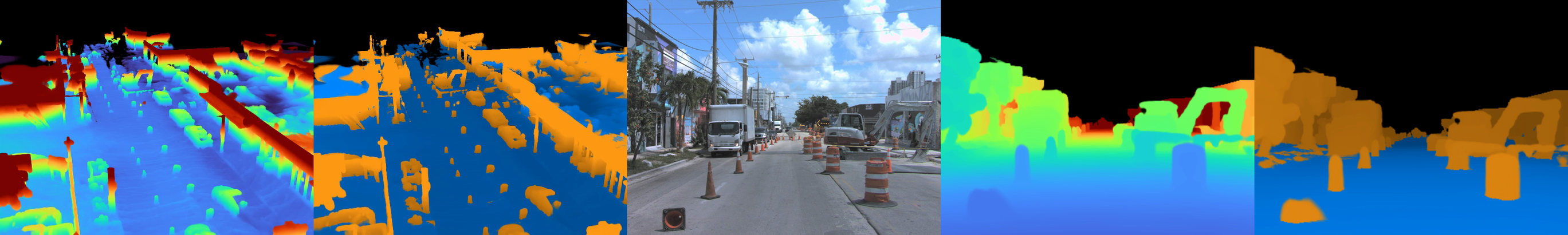}};
        \node[inner sep=0pt, outer sep=0, anchor=north] (scene-2) at  ([yshift=-3pt]scene-1.south) {\includegraphics[width=\imw]{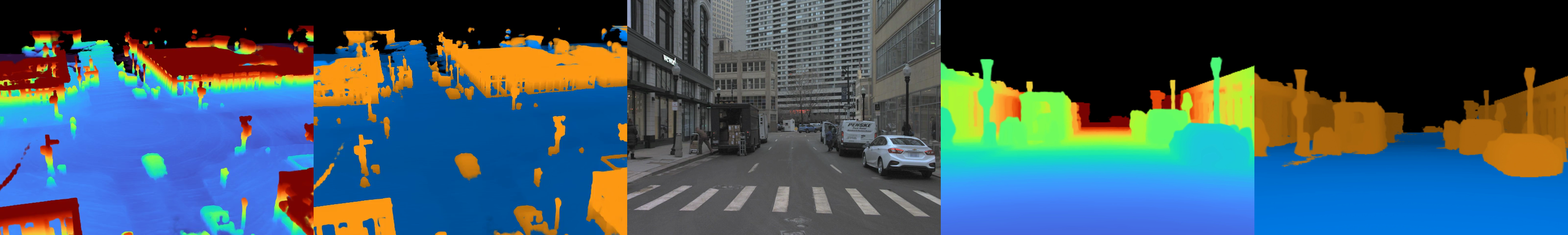}};
        \node[inner sep=0pt, outer sep=0, anchor=north] (scene-3) at  ([yshift=-3pt]scene-2.south) {\includegraphics[width=\imw]{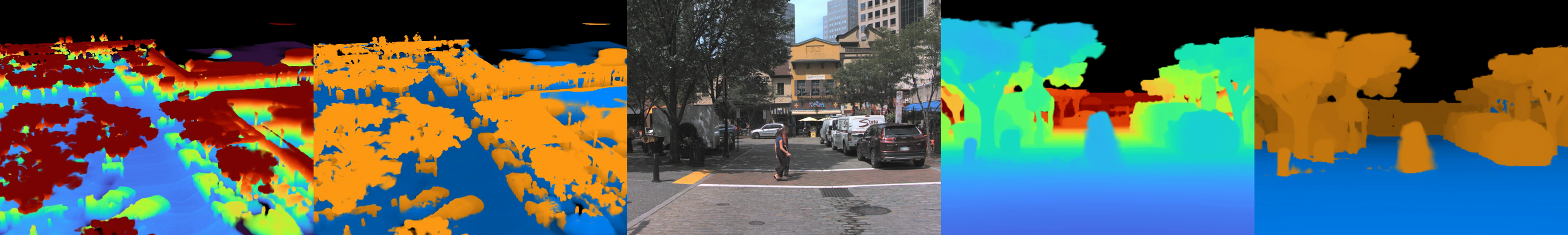}};

        \node[anchor = south, rotate=90] at (scene-1.west) {Scene 1};
        \node[anchor = south, rotate=90] at (scene-2.west) {Scene 2};
        \node[anchor = south, rotate=90] at (scene-3.west) {Scene 3};

        \node[anchor=south, font=\small\bfseries] at ([xshift=-0.4*\imw]scene-1.north) {Occupancy};
        \node[anchor=south, font=\small\bfseries] at ([xshift=-0.2*\imw]scene-1.north) {Obstacle Seg};
        \node[anchor=south, font=\small\bfseries] at ([xshift=0.0*\imw]scene-1.north) {Image};
        \node[anchor=south, font=\small\bfseries] at ([xshift=0.2*\imw]scene-1.north) {Cam Depth};
        \node[anchor=south, font=\small\bfseries] at ([xshift=0.4*\imw]scene-1.north) {Cam Seg};

        \draw[white, thick] ([xshift=-70pt, yshift=8pt]scene-1.center) circle (8pt);
        \node[black, fill=white, draw=white, font=\bfseries\tiny, circle, inner sep=0.5pt, minimum size=4pt] 
            at ([xshift=-60pt, yshift=13.66pt]scene-1.center) {A};

        \draw[white, thick] ([xshift=150pt, yshift=0pt]scene-1.center) circle (12pt);
        \node[black, fill=white, draw=white, font=\bfseries\tiny, circle, inner sep=0.5pt, minimum size=6pt] 
            at ([xshift=142pt, yshift=13.66pt]scene-1.center) {A};

        \draw[white, thick] ([xshift=109pt, yshift=-20pt]scene-1.center) circle (6pt);
        \node[black, fill=white, draw=white, font=\bfseries\tiny, circle, inner sep=0.5pt, minimum size=6pt] 
            at ([xshift=118.5pt, yshift=-20pt]scene-1.center) {B};

        \draw[white, thick] ([xshift=120pt, yshift=-52pt]scene-1.center) circle (4pt);
        \node[black, fill=white, draw=white, font=\bfseries\tiny, circle, inner sep=0.5pt, minimum size=6pt] 
            at ([xshift=125pt, yshift=-46pt]scene-1.center) {C};

        \draw[white, thick] ([xshift=140pt, yshift=-55pt]scene-1.center) circle (6pt);
        \node[black, fill=white, draw=white, font=\bfseries\tiny, circle, inner sep=0.5pt, minimum size=6pt] 
            at ([xshift=135pt, yshift=-63.5pt]scene-1.center) {D};
        
        \draw[white, thick] ([xshift=132.5pt, yshift=-110.5pt]scene-1.center) circle (8pt);
        \node[black, fill=white, draw=white, font=\bfseries\tiny, circle, inner sep=0.5pt, minimum size=6pt] 
            at ([xshift=127.5pt, yshift=-121pt]scene-1.center) {E};

    \end{tikzpicture}
\caption{
    Visualization of \ourmodel{} on AV2. 
    Each row shows (left to right): perspective-view occupancy, perspective-view obstacle segmentation, 
    camera image, rendered depth, and rendered obstacle segmentation. 
    \ourmodel{} captures key elements of the scenes: \textbf{A)} excavator, \textbf{B)} fallen cone, \textbf{C)} inside of truck, \textbf{D)} open car door, \textbf{E)} pedestrians.
}
    \label{fig:occ_seg_present_av2}    
\end{figure*}
\label{sec:stu_experiments}
We evaluate our method’s ability to perceive unusual, path-blocking objects on STU. First, we assess pseudo-label quality by comparing our \ourimagelabels{} and \ourcombinedlabels{} against STU ground truth. Using the correspondence between image and LiDAR labels in STU, we benchmark our soft pseudo-labels against open-set 2D and 3D baselines on equal footing. We also provide qualitatives results on AV2 in \cref{fig:occ_seg_present_av2}.
\\
\begin{table*}[t]
    \label{tab:stu_combination}
    \centering
    \footnotesize
    \setlength\tabcolsep{2pt}
    
    \begin{minipage}[t]{0.47\textwidth}
        \centering
        \caption{Pseudo-label model segmentation performance on STU. MCPP refers to MaskCLIP++.  }
        \label{tab:pseudo_labels}
        \resizebox{\linewidth}{!}{%
        \begin{tabular}{@{}l ccc@{}}
        \toprule 
        Method & mIoU $\uparrow$ & F1 $\uparrow$ & Recall $\uparrow$ \\ 
        \midrule
        MCPP + SAM        & 22.2 & 38.5 & 51.8 \\ 
        MCPP + SAM (Oracle)    & 22.2 & 38.3 & 51.9 \\ 
        MCPP + FC-CLIP         & 24.8 & 30.3  & 43.9 \\
        MCPP + FC-CLIP (Oracle)     & 27.4 & 35.1  & 52.4 \\
        MCPP + MAFT+           & 25.6 & 34.0  & 51.6 \\ 
        MCPP + MAFT+ (Oracle)       & 28.0 & 37.8 & 59.5 \\ 
        GroundedSAM                   & 22.0 & 27.9 & 35.3 \\ 
        GroundedSAM (Oracle)                     & 26.8 & 18.7 & 55.2 \\ 
        OWLv2 + SAM                & 20.2 & 21.9 & 23.4 \\ 
        OWLv2 + SAM (Oracle)                     & 28.6 & 29.5 & 57.2 \\ 
        Patchwork++              & 42.1 & 51.4 & 62.8 \\ 
        \rowcolor{tablecolor} \ourimagelabels{} (Ours)                                               & 38.6 & 35.0 & 76.3 \\ 
        \rowcolor{tablecolor} 
        \ourcombinedlabels{} (Ours)                                   & \textbf{44.6} & \textbf{57.8} & \textbf{79.9} \\ 
        \bottomrule
        \end{tabular}}
    \end{minipage}%
    \hfill 
    \begin{minipage}[t]{0.48\textwidth}
        \centering
        \caption{Semantic occupancy prediction performance of distilled methods. $\dagger$ is QueryOcc+ with  LiDAR input. 
        }
        \label{tab:distilled_occupancy}
        \resizebox{\linewidth}{!}{%
        \begin{tabular}{@{}l cccc@{}}
        \toprule 
        Method & mIoU $\uparrow$ & AP $\uparrow$ & F1 $\uparrow$ & Rec. $\uparrow$ \\ 
        \midrule
        UnO (\ourimagelabels{})                                    & 40.1 & 7.4  & 11.6 & 23.5 \\ 
        DiO (\ourimagelabels{})                                   & 49.7 & 29.3 & 34.3 & 50.2 \\ 
        UnO (\ourcombinedlabels{})                                 & 40.6 & 23.1 & 29.8 & 25.5 \\ 
        DiO (\ourcombinedlabels{})                                 & 49.8 & 74.8 & 72.9 & 65.6 \\ 
        \midrule
        TriO (MCPP + SAM)                        & 27.9 & 25.7 & 27.5 & 24.0 \\ 
        TriO (MCPP + FC-CLIP)                    & 23.1 & 2.2 & 5.2 & 56.7 \\ 
        TriO (MCPP + MAFT+)                      & 20.6 & 10.2 & 14.8 & 10.0  \\ 
        TriO (GroundedSAM)$\dagger$                             & 50.4 & 50.1 & 52.1 & 54.8 \\ 
        TriO (OWLv2 + SAM)                             & 50.1 & 72.3 & 70.1 & 68.8 \\ 
        TriO (Patchwork++)                             & 39.4 & 74.8 & 70.8 & 59.5 \\ 
        \rowcolor{tablecolor} 
        TriO (\ourimagelabels{}) (Ours)                                  & 41.8 & 57.1 & 53.7 & 59.9 \\ 
        \rowcolor{tablecolor} 
        TriO (\ourcombinedlabels{} Hard) (Ours)                    & 50.3 & 76.1 & 71.7 & 64.5 \\ 
        \rowcolor{tablecolor} 
        TriO (\ourcombinedlabels{} Soft) (Ours)                    & \textbf{54.9} & \textbf{81.6} & \textbf{77.7} & \textbf{71.3} \\ 
        \bottomrule
        \end{tabular}}
    \end{minipage}
\end{table*}
\textbf{Anything Perception Baselines.} We benchmark against state-of-the-art open-vocabulary methods, including \textit{detect-then-segment} (GroundedSAM \cite{groundedsam24}, OWLv2 \cite{OWLv2neurips23}+SAM \cite{Kirillov_2023_ICCV}) and \textit{segment-then-classify} (MaskCLIP++ \cite{maskclippp25} with FC-CLIP \cite{FCCLIP_NIPS23}, MAFT+ \cite{MAFTp_ECCV2024}, or SAM) approaches. Following \cite{sal2024eccv}, we use standard prompts for ground ($\mathcal{P}_{\text{ground}}$) and non-ground ($\mathcal{P}_{\text{obstacle}}$) classes; see supplementary material for details.  To establish an upper performance bound, we include an (Oracle) variant where scenes are manually inspected and specific unknown object classes are appended to $\mathcal{P}_{\text{obstacle}}$. Finally, we also include Patchwork++ \cite{patchworkpp} as a LiDAR-based baseline.
For the second set of experiments, we train \ourmodel{}'s segmentation objective ($\groundloss$) using each of the pseudo-label from the baselines. This allows us to measure how effectively open-world knowledge can be distilled into the downstream world model, and in turn how well this model generalizes when evaluated on a different domain. Notably, our model under GroundedSAM supervision serves as a LiDAR-enhanced version of QueryOcc+ \cite{QueryOcc_2025}. Finally, we compare against LiDAR-only models (UnO, DiO) trained with \ourimagelabels{} and \ourcombinedlabels{}. \\
\begin{figure*}[t]
    \centering
    \begin{tikzpicture}[
        green check/.style={
            circle, fill=ForestGreen!80!black, 
            draw=white, line width=0.5pt,
            opacity=0.6,
            inner sep=0.75pt, minimum size=6pt,
            path picture={
                \draw[white, opacity=0.6, line width=0.6pt, line cap=round, line join=round]
                ([xshift=-1.25pt,yshift=-0.25pt]path picture bounding box.center) --
                ([xshift=-0.25pt,yshift=-1.25pt]path picture bounding box.center) --
                ([xshift=1.5pt,yshift=1.25pt]path picture bounding box.center);
            }
        },
        red x/.style={
            circle, fill=red!80!black,
            draw=white, line width=0.5pt,
            opacity=0.6,
            inner sep=0.75pt, minimum size=6pt,
            path picture={
                \draw[white, opacity=0.6, line width=0.6pt, line cap=round]
                ([xshift=-1.25pt,yshift=-1.25pt]path picture bounding box.center) --
                ([xshift=1.25pt,yshift=1.25pt]path picture bounding box.center)
                ([xshift=-1.25pt,yshift=1.25pt]path picture bounding box.center) --
                ([xshift=1.25pt,yshift=-1.25pt]path picture bounding box.center);
            }
        },
        badge/.style={anchor=south east}
    ]
        \pgfmathsetlengthmacro{\imw}{0.95 \linewidth}
        \pgfmathsetlengthmacro{\sep}{1mm}
        \pgfmathsetlengthmacro{\ysep}{1mm}

        \node[inner sep=0pt, outer sep=0, anchor=east] (scene-1) at  (0,0) {\includegraphics[width=\imw]{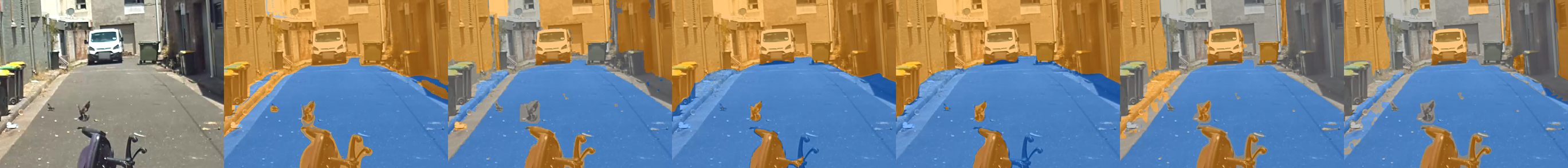}};
        \node[inner sep=0pt, outer sep=0, anchor=north] (scene-2) at  ([yshift=-3pt]scene-1.south) {\includegraphics[width=\imw]{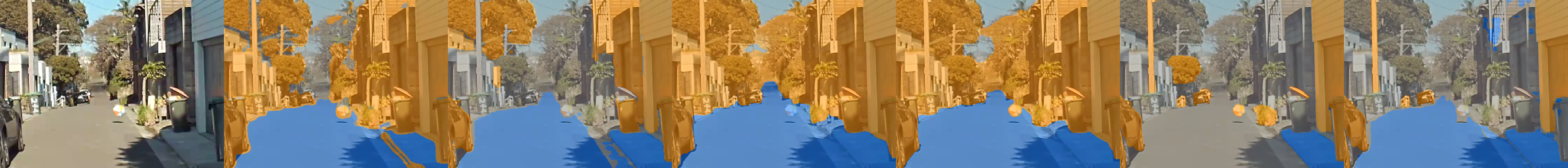}};
        \node[inner sep=0pt, outer sep=0, anchor=north] (scene-3) at  ([yshift=-3pt]scene-2.south) {\includegraphics[width=\imw]{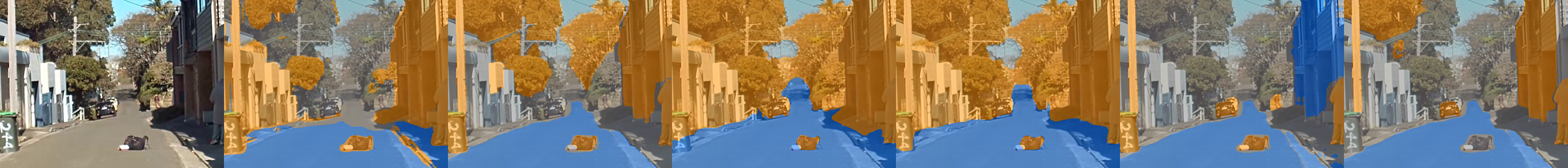}};

        \node[anchor = south, rotate=90] at (scene-1.west) {Scene 1};
        \node[anchor = south, rotate=90] at (scene-2.west) {Scene 2};
        \node[anchor = south, rotate=90] at (scene-3.west) {Scene 3};
        
        \node[anchor = south, align=center, font=\scriptsize] at ([xshift=-0.429*\imw]scene-1.north) {Reference};
        \node[anchor = south, align=center, font=\scriptsize] at ([xshift=-0.286*\imw]scene-1.north) {\ourimagelabels (Ours)};
        \node[anchor = south, align=center, font=\scriptsize] at ([xshift=-0.143*\imw]scene-1.north) {MCPP +\\SAM};
        \node[anchor = south, align=center, font=\scriptsize] at ([xshift=0*\imw]scene-1.north) {MCPP +\\FC-CLIP};
        \node[anchor = south, align=center, font=\scriptsize] at ([xshift=0.143*\imw]scene-1.north) {MCPP +\\MAFT+};
        \node[anchor = south, align=center, font=\scriptsize] at ([xshift=0.286*\imw]scene-1.north) {Grounded\\SAM};
        \node[anchor = south, align=center, font=\scriptsize] at ([xshift=0.429*\imw]scene-1.north) {OWLv2 +\\SAM};

        \node[badge, green check]       at ([xshift=-0.2143*\imw - 2pt, yshift=2pt]scene-1.south) {};
        \node[badge, red x]       at ([xshift=-0.0714*\imw - 2pt, yshift=2pt]scene-1.south) {};
        \node[badge, red x] at ([xshift=0.0714*\imw - 2pt, yshift=2pt]scene-1.south) {};
        \node[badge, red x] at ([xshift=0.2143*\imw - 2pt, yshift=2pt]scene-1.south) {};
        \node[badge, red x]       at ([xshift=0.3571*\imw - 2pt, yshift=2pt]scene-1.south) {};
        \node[badge, red x] at ([xshift=0.5000*\imw - 2pt, yshift=2pt]scene-1.south) {};

        \node[badge, green check]       at ([xshift=-0.2143*\imw - 2pt, yshift=2pt]scene-2.south) {};
        \node[badge, red x] at ([xshift=-0.0714*\imw - 2pt, yshift=2pt]scene-2.south) {};
        \node[badge, red x] at ([xshift=0.0714*\imw - 2pt, yshift=2pt]scene-2.south) {};
        \node[badge, green check]       at ([xshift=0.2143*\imw - 2pt, yshift=2pt]scene-2.south) {};
        \node[badge, red x]       at ([xshift=0.3571*\imw - 2pt, yshift=2pt]scene-2.south) {};
        \node[badge, red x] at ([xshift=0.5000*\imw - 2pt, yshift=2pt]scene-2.south) {};

        \node[badge, green check] at ([xshift=-0.2143*\imw - 2pt, yshift=2pt]scene-3.south) {};
        \node[badge, green check]       at ([xshift=-0.0714*\imw - 2pt, yshift=2pt]scene-3.south) {};
        \node[badge, green check] at ([xshift=0.0714*\imw - 2pt, yshift=2pt]scene-3.south) {};
        \node[badge, green check] at ([xshift=0.2143*\imw - 2pt, yshift=2pt]scene-3.south) {};
        \node[badge, green check] at ([xshift=0.3571*\imw - 2pt, yshift=2pt]scene-3.south) {};
        \node[badge, red x]       at ([xshift=0.5000*\imw - 2pt, yshift=2pt]scene-3.south) {};

        \draw[white, thick] ([xshift=-104pt, yshift=-22pt]scene-1.north) circle (7pt);
        \node[black, fill=white, draw=white, font=\bfseries\tiny, circle, inner sep=0.5pt, minimum size=6pt] 
            at ([xshift=-99pt, yshift=-12pt]scene-1.north) {A};

        \draw[white, thick] ([xshift=-47pt, yshift=-31pt]scene-1.north) circle (7pt);
        \node[black, fill=white, draw=white, font=\bfseries\tiny, circle, inner sep=0.5pt, minimum size=6pt] 
            at ([xshift=-39pt, yshift=-23pt]scene-1.north) {B};

        \draw[white, thick] ([xshift=-93pt, yshift=-62pt]scene-1.north) circle (4pt);
        \node[black, fill=white, draw=white, font=\bfseries\tiny, circle, inner sep=0.5pt, minimum size=6pt] 
            at ([xshift=-99pt, yshift=-68pt]scene-1.north) {C};

        \draw[white, thick] ([xshift=-90pt, yshift=-107pt]scene-1.north) circle (5pt);
        \node[black, fill=white, draw=white, font=\bfseries\tiny, circle, inner sep=0.5pt, minimum size=6pt] 
            at ([xshift=-99pt, yshift=-107pt]scene-1.north) {D};

    \end{tikzpicture}
    \caption{
        Segmentation results on the STU validation set. Each scene contains an unexpected obstacle: \textbf{A)} two pigeons and \textbf{B)} chair, \textbf{C)} ball, and \textbf{D)} bookbag. An obstacle segmentation is correct if the road is labeled traversable (\textcolor{blue}{\textbf{blue}}) and the obstacle is fully segmented as an obstacle (\textcolor{orange}{\textbf{orange}}). \ourimagelabels{} robustly segments long-tail obstacles.
    }
    \label{fig:cameramodels_STU}    
\end{figure*}
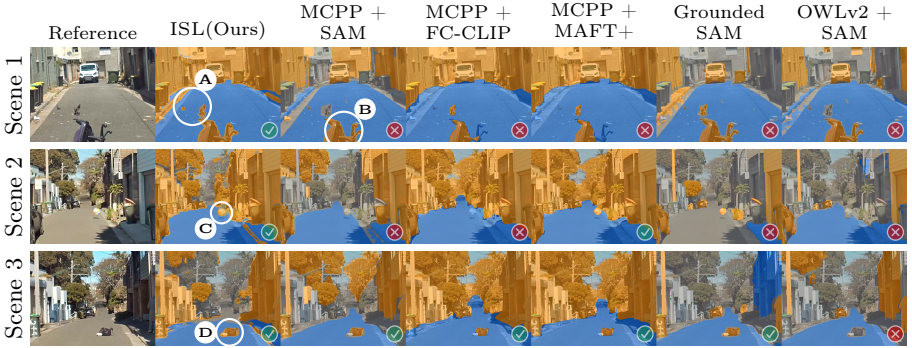
\textbf{Anything Perception Metrics.}
We report \textit{mIoU} (mean Intersection over Union) \cite{NEURIPS2023_Occ3D, OpenOccupancy_2023_ICCV} for semantic predictions and \textit{F1} score to evaluate pseudo-label quality at a fixed operating point without unfairly advantaging soft probability methods. We emphasize \textit{Recall} to capture safety-critical missed detections, while Average Precision (\textit{AP}) evaluates continuous probabilities in distilled occupancy models. See supplementary for further details.\\
\textbf{Comparison of Pseudo-Label Methods.}
Our generated pseudo-labels (\ourcombinedlabels) achieve the best balance of detecting the obstacles (measured by recall) on the road without too many false positive detections (F1 and mIoU) as shown in \cref{tab:pseudo_labels}. 
Notably, this strong performance is not solely reliant on multi-modal fusion; even our image-only pseudo-labels (\ourimagelabels) surpass all evaluated baselines in terms of recall. We also find that text-based segmentation baselines, even when prompted with the exact text description of the objects in the road, struggle to generalize to these rarely seen classes, reinforcing the findings of \cite{Saric_2025_ICCV}. We provide visual evidence for that claim in \cref{fig:cameramodels_STU} and the supplementary. In contrast, our model, relying on geometry, is able to generalize in a more robust manner. \\
\textbf{Semantic Occupancy Evaluation.}
As shown in \cref{tab:distilled_occupancy} \ourmodel{} performs the best after the obstacle labels have been distilled into our occupancy model. We also find that using soft probabilistic labels (\ourcombinedlabels{} Soft) tends to result in better performance than when trained with thresholded discretized labels (\ourcombinedlabels{} Hard) due to better calibration (see AP, mIoU). \ourmodel{} improves also over LiDAR-only methods highlighting the benefit of using multi-modal sensor information.

\begin{figure*}[t]

    \centering
    \begin{tikzpicture}
        \pgfmathsetlengthmacro{\imw}{0.115\linewidth}
        \node[inner sep=0pt, outer sep=0, anchor=west ]  (60-60-t0-lidar)     at (0,0)                              {\adjustbox{bgcolor=black}{\includegraphics[width=\imw]{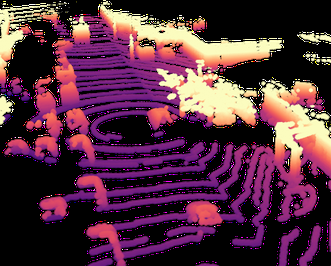}}};
        \node[inner sep=0pt, outer sep=0, anchor=north ] (60-60-t0-image)     at ([yshift=-1pt]60-60-t0-lidar.south) {\includegraphics[width=\imw]{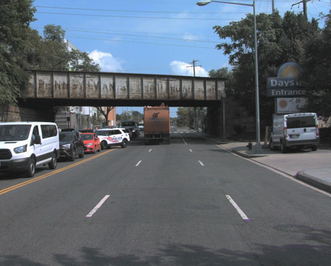}};
        \node[inner sep=0pt, outer sep=0, anchor=west ]  (60-60-occ-0)        at ([xshift=2pt]60-60-t0-lidar.east)   {\includegraphics[width=\imw]{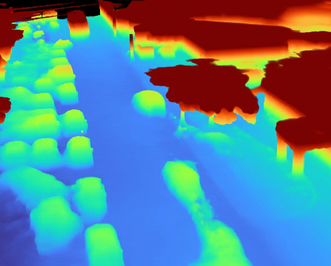}};
        \node[inner sep=0pt, outer sep=0, anchor=west ]  (60-60-occ-1)        at (60-60-occ-0.east)                  {\includegraphics[width=\imw]{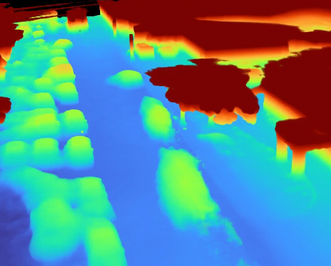}};
        \node[inner sep=0pt, outer sep=0, anchor=west ]  (60-60-occ-2)        at (60-60-occ-1.east)                  {\includegraphics[width=\imw]{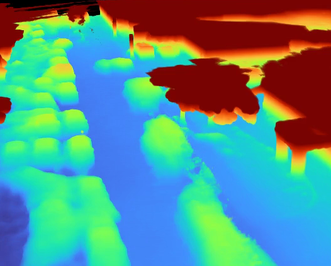}};
        \node[inner sep=0pt, outer sep=0, anchor=north ] (flow_60-60-gt-0)    at ([yshift=-1pt]60-60-occ-0.south)    {\includegraphics[width=\imw]{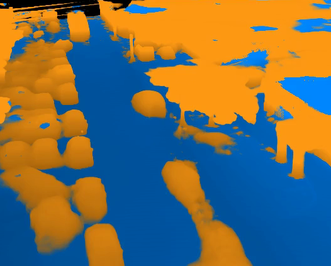}};
        \node[inner sep=0pt, outer sep=0, anchor=west ]  (flow_60-60-gt-1)    at (flow_60-60-gt-0.east)              {\includegraphics[width=\imw]{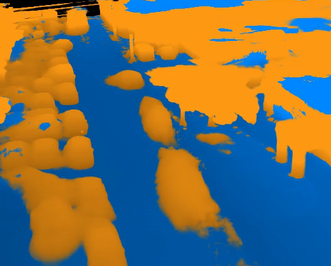}};
        \node[inner sep=0pt, outer sep=0, anchor=west ]  (flow_60-60-gt-2)    at (flow_60-60-gt-1.east)              {\includegraphics[width=\imw]{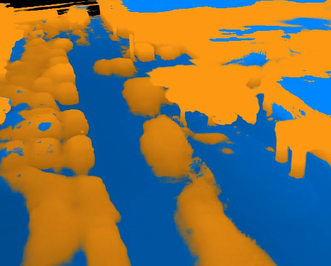}};

        \node[inner sep=0pt, outer sep=0, anchor=west ]  (148-50-t0-lidar)    at ([xshift=5pt]60-60-occ-2.east)      {\adjustbox{bgcolor=black}{\includegraphics[width=\imw]{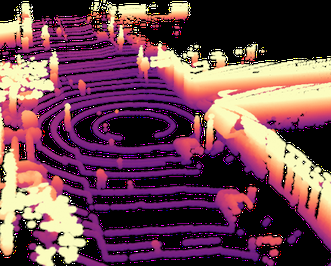}}};
        \node[inner sep=0pt, outer sep=0, anchor=north ] (148-50-t0-image)    at ([yshift=-1pt]148-50-t0-lidar.south) {\includegraphics[width=\imw]{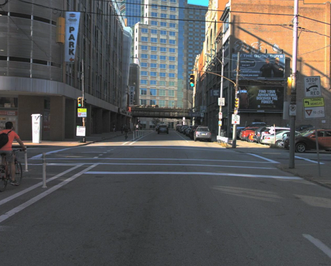}};
        \node[inner sep=0pt, outer sep=0, anchor=west ]  (148-50-occ-0)       at ([xshift=2pt]148-50-t0-lidar.east)  {\includegraphics[width=\imw]{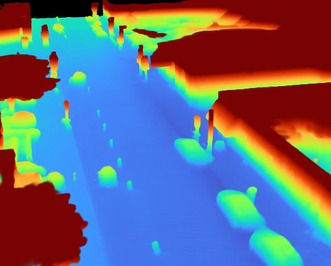}};
        \node[inner sep=0pt, outer sep=0, anchor=west ]  (148-50-occ-1)       at (148-50-occ-0.east)                 {\includegraphics[width=\imw]{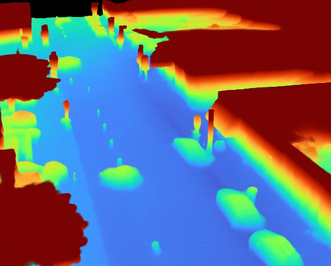}};
        \node[inner sep=0pt, outer sep=0, anchor=west ]  (148-50-occ-2)       at (148-50-occ-1.east)                 {\includegraphics[width=\imw]{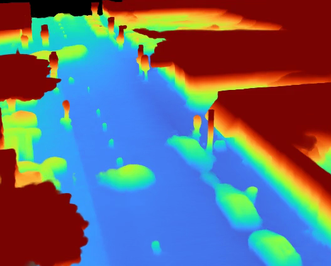}};
        \node[inner sep=0pt, outer sep=0, anchor=north ] (flow_148-50-gt-0)   at ([yshift=-1pt]148-50-occ-0.south)   {\includegraphics[width=\imw]{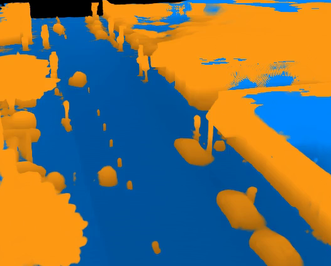}};
        \node[inner sep=0pt, outer sep=0, anchor=west ]  (flow_148-50-gt-1)   at (flow_148-50-gt-0.east)             {\includegraphics[width=\imw]{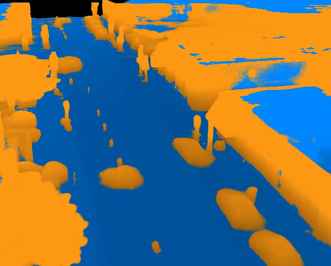}};
        \node[inner sep=0pt, outer sep=0, anchor=west ]  (flow_148-50-gt-2)   at (flow_148-50-gt-1.east)             {\includegraphics[width=\imw]{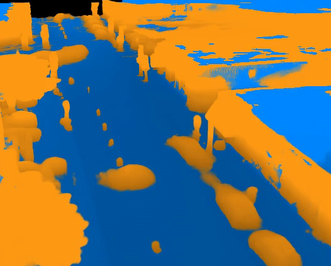}};

        \node[anchor = south] (t0-label-1)   at (60-60-t0-lidar.north)    {\small $\SI{0.0}{s}$};
        \node[anchor = south] at (60-60-occ-0.north)                      {\small $+\SI{0.6}{s}$};
        \node[anchor = south] (mid-1) at (60-60-occ-1.north)              {\small $+\SI{1.8}{s}$};
        \node[anchor = south] at (60-60-occ-2.north)                      {\small $+\SI{3.0}{s}$};

        \node[anchor = south] (t0-label-2)   at (148-50-t0-lidar.north)   {\small $\SI{0.0}{s}$};
        \node[anchor = south] at (148-50-occ-0.north)                     {\small $+\SI{0.6}{s}$};
        \node[anchor = south] (mid-2) at (148-50-occ-1.north)             {\small $+\SI{1.8}{s}$};
        \node[anchor = south] at (148-50-occ-2.north)                     {\small $+\SI{3.0}{s}$};

        \node[anchor = south] at ([yshift=-4pt]mid-1.north) {\small Example 1};
        \node[anchor = south] at ([yshift=-4pt]mid-2.north) {\small Example 2};

        \draw[white, thick] ([xshift=-10pt, yshift=6pt]148-50-occ-0) circle (4pt);
        \node[black, fill=white, draw=white, font=\bfseries\tiny, circle, inner sep=0.5pt, minimum size=6pt] 
            at ([xshift=-2.0pt, yshift=6pt]148-50-occ-0) {A};

        \draw[white, thick] ([xshift=-11pt, yshift=10pt]148-50-occ-2) circle (4pt);
        \node[black, fill=white, draw=white, font=\bfseries\tiny, circle, inner sep=0.5pt, minimum size=6pt] 
            at ([xshift=-3.0pt, yshift=10pt]148-50-occ-2) {B};

    \end{tikzpicture}
    \caption{
        Scene forecasting results of \ourmodel{} on AV2.  LiDAR and camera sensor inputs are displayed at present time for reference. \ourmodel{}'s forecasts maintain coherent obstacle segmentation throughout time, distinguishing occupied from free space even as agents move. \ourmodel{} captures a bicyclist \textbf{A)} and predicts its future multimodality \textbf{B)}. 
    }
    \label{fig:occ_seg_forecasting}
\end{figure*}
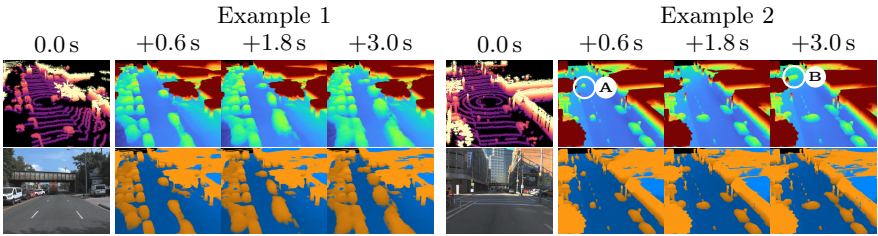

\subsection{Geometric Occupancy/ LiDAR Point Cloud Forecasting (Q2)} %
\label{sec:exp_lidar_forecasting}
We evaluate \ourmodel{} on the AV2 Occupancy Forecasting benchmark; the goal of this benchmark is to determine how \ourmodel{} performs on occupancy prediction (visualized in \cref{fig:occ_seg_forecasting}), evaluated via LiDAR point cloud forecasting. Following prior work \cite{khurana2023point,Agro_2024_CVPR,Diehl_2025_DIO}, after training the world model $f_{\theta}$, we freeze all parameters $\theta$ and introduce a LiDAR forecasting header $f_{\psi}$ to predict LiDAR depth. We provide more details in the supplementary. \\
\textbf{LiDAR Forecasting Baselines.} We compare our method against the state-of-the-art from the Argoverse 2 leaderboard. Notably, 4D-Occ \cite{khurana2023point} serves as a representative unsupervised baseline for explicit occupancy prediction via voxel grid forecasting. In contrast, UnO \cite{Agro_2024_CVPR} and DiO \cite{Diehl_2025_DIO}, also perform implicit occupancy prediction. However, unlike \ourmodel{}, these methods do not incorporate camera input, representing architectural differences in sensor modality.\\
\textbf{LiDAR Forecasting Metrics:} For point cloud forecasting we use well-established metrics of the AV2 Occupancy Forecasting leaderboard, namely  Chamfer distance (CD), Near Field Chamfer Distance (\textit{NFCD}), depth L1 error (\textit{L1}), depth relative \textit{L1} error (AbsRel) as our metrics. \\
\textbf{Comparison against state-of-the-art.}
\cref{tab:supp-full-leaderboard} compares \ourmodel{} against all the AV2  leaderboard baselines. \ourmodel{} is best in the L1 and AbsRel metrics, while competitive with baselines in other metrics. This underlines \ourmodel{}'s ability to accurately forecast occupancy and its effectiveness as a world model.

\begin{table}[t]
    \setlength\tabcolsep{2pt} %
    \centering
    \footnotesize
    \caption{LiDAR forecasting on the Argoverse 2 leaderboard.}
    \label{tab:supp-full-leaderboard}
    \resizebox{0.7\textwidth}{!}{%
    \begin{tabular}{@{}l cccccc@{}}
    \toprule 
     & L1 (\SI{}{\meter}) $\downarrow$ & AbsRel (\SI{}{\percent}) $\downarrow$ & NFCD (\SI{}{\meter\squared}) $\downarrow$ & CD (\SI{}{\meter\squared}) $\downarrow$ \\ 
    \midrule       
     RayTracing  & 4.88 & 35.00 & 3.62 & 17.03 \\ 
     $\text{occformer}\_\text{ep15}$ & 3.57 & 22.00  & 3.35  & 91.61 \\
     4D-Occ        & 3.22 & 19.00 & 2.45 & 72.74 \\ 
     occformer & 2.97 & 17.00  & 2.03  & 71.33 \\
     Progressive ARM & 2.32 & 13.00  & 1.81  & 71.41 \\
     UnO                            & 2.24 & 12.00  & 0.86  & \textbf{8.10} \\ 
     DIO                     & 2.11    & 11.00     & \textbf{0.85}     & 13.96    \\ 
     \rowcolor{tablecolor}  \ourmodel        & \textbf{1.82}           & \textbf{10.63}   &  0.93    & 10.90       \\ 
    \bottomrule
    \end{tabular}
    }
    
\end{table}

\subsection{Fine-tuning for Downstream Occupancy Tasks (Q3)}  
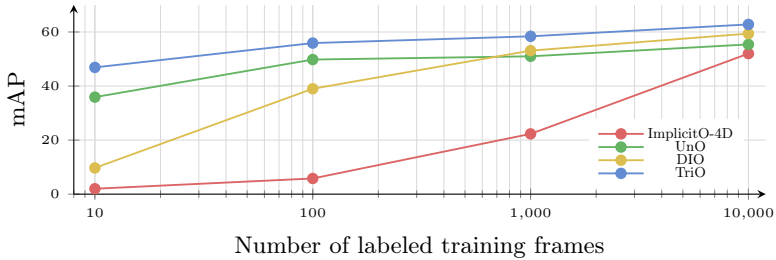
\begin{figure}[th]
    \centering

    \begin{tikzpicture}
    \definecolor{clr1}{RGB}{220,100,100}  %
    \definecolor{clr2}{RGB}{100,180,100}  %
    \definecolor{clr3}{RGB}{220,190,80}   %
    \definecolor{clr4}{RGB}{100,140,210}  %

    \begin{axis}[
        xlabel=Number of labeled training frames,
        ylabel=mAP,
        xmode=log,
        log ticks with fixed point,
        legend pos=south east,
        legend style={nodes={scale=0.6, transform shape}, draw=none, row sep=-2pt},
        width=0.75\linewidth,
        height=2.5cm,
        scale only axis,
        axis lines=left,
        label style={font=\scriptsize},
        tick label style={font=\tiny},
        xlabel near ticks,
        ylabel near ticks,
        xmin=8,
        xmax=12000,
        ymin=0,
        ymax=70,
        ytick={0,20,40,60},
        grid=both,
        grid style={line width=.1pt, draw=gray!30},
    ]

    \addplot[color=clr1, mark=*, mark size=1.8pt, line width=0.7pt] coordinates {
        (10, 2.0) (100, 5.8) (1000, 22.3) (10000, 52.0)
    };
    \addlegendentry{ImplicitO-4D}

    \addplot[color=clr2, mark=*, mark size=1.8pt, line width=0.7pt] coordinates {
        (10, 35.9) (100, 49.8) (1000, 51.0) (10000, 55.4)
    };
    \addlegendentry{UnO}

    \addplot[color=clr3, mark=*, mark size=1.8pt, line width=0.7pt] coordinates {
        (10, 9.7) (100, 39.0) (1000, 53.1) (10000, 59.4)
    };
    \addlegendentry{DIO}

    \addplot[color=clr4, mark=*, mark size=1.8pt, line width=0.7pt] coordinates {
        (10, 46.9) (100, 55.9) (1000, 58.4) (10000, 62.8)
    };
    \addlegendentry{TriO}

    \end{axis}
    \end{tikzpicture}
    \caption{
        Occupancy fine-tuning under a varying number of labelled frames.
    }
    \label{fig:scaling_exps}
\end{figure}

We show the versatility of the \ourmodel{} pre-trained feature extractor by evaluating its performance on various downstream tasks. We follow \cite{GASP_2025, Agro_2024_CVPR} and finetune the model to forecast semantic 3D (AV2) and semantic 2D BEV (\nameourdataset{}) occupancy and flow over time supervised with bounding box labels for the vehicle class. 
\\
\textbf{Downstream Occupancy Baselines.} For fine-tuning experiments, we pre-train UnO and DiO baselines via LiDAR self-supervision then fine-tune on 2D or 3D bounding box labels. We also compare against ImplicitO \cite{agro2023implicit} which does not utilize unsupervised pre-training. \\
\textbf{Downstream Occupancy Metrics.}
We use well-established metrics to evaluate the performance of the different methods on these downstream tasks by following prior work \cite{mahjourian2022occupancy, agro2023implicit,Agro_2024_CVPR,Diehl_2025_DIO}. We report mAP, SoftIoU \cite{mahjourian2022occupancy} for occupancy and End Point Error (EPE) for flow. For \nameourdataset{}, we also evaluate each metric at over 200m in front of the SDV. We hypothesize that RADAR specifically having a longer effective range is able to improve our model performance at these distances, where other sensor readings may be too sparse to be consistently useful. This emphasizes the use case of \nameourdataset{} as only it is able to provide validation labels at a distance of 350m in front of the SDV.\\
\textbf{Semantic Occupancy and Flow Forecasting.} 
\cref{tab:downstream_argo_long_range_split} and \cref{tab:ablation_long_range_split} show the BEV and 3D occupancy and flow forecasting results on AV2 and \nameourdataset{}. Our method outperforms all the baselines when finetuned to the supervised occupancy task. The drop in EPE performance in \cref{tab:downstream_argo_long_range_split} compared to DIO is explained as TriO on AV2 has no flow supervision (due to the absence of RADAR data) during pre-training, whereas DIO does. Notably, our model outperforms the baselines significantly at range. The results show an improvement of 6.4 AP over the state-of-the-art showcasing the benefits of using RADAR at long-range.\\
\textbf{Ablation Study.} 
We evaluate the contributions of pre-training, sensor modalities, and loss objectives in \cref{tab:ablation_long_range_split}. Unsupervised pre-training provides consistent gains across all metrics, validating the world model objectives. Omitting the radar loss leads to degradation in flow metrics across all ranges. Conversely, adding camera inputs improves far range occupancy performance at the expense of EPE; this suggests that single-frame camera features, which lack temporal context, may introduce noise into the flow estimation process. Removing RADAR as an input modality significantly degrades performance on long range metrics, supporting our hypothesis that this sensor modality directly lends itself to long range tasks.\\
\textbf{Scaling Labeled Data.}
In \cref{fig:scaling_exps} we scale the number of labeled training samples when fine-tuning pre-trained models (\ourmodel{}, UnO, DIO) and models trained from scratch (ImplicitO-4D). \ourmodel{} outperforms at all levels of supervision, and with $10^2$ labeled training examples, reaches similar performance to the strongest baselines with $10^4$ examples. \ourmodel{} is \textit{two-orders of magnitude} more sample efficient, underscoring the generalizability of our method. 
Due to its novel obstacle segmentation supervision \ourmodel{} better understands obstacles and free space.
This underlines the strong representation learned during pre-training and \ourmodel{}'s ability to encode the geometry, semantics, and dynamics of the scene.
\begin{table*}[t]
    \centering
    \footnotesize
    \setlength\tabcolsep{2pt}
    \caption{Fine-tuning downstream performance on AV2 and \nameourdataset{}.}
    \label{tab:downstream_argo_long_range_split}
    \resizebox{\linewidth}{!}{%
    \begin{tabular}{@{}l ccc ccc ccc@{}}
    \toprule
     & \multicolumn{3}{c}{AV2} & \multicolumn{6}{c}{\nameourdataset{}} \\
    \cmidrule(lr){2-4} \cmidrule(lr){5-10}
     & & & & \multicolumn{3}{c}{All} & \multicolumn{3}{c}{200m+} \\
    \cmidrule(lr){5-7} \cmidrule(lr){8-10}
    Method & mAP $\uparrow$ & Soft-IoU $\uparrow$ & EPE $\downarrow$ & mAP $\uparrow$ & Soft-IoU $\uparrow$ & EPE $\downarrow$ & mAP $\uparrow$ & Soft-IoU $\uparrow$ & EPE $\downarrow$ \\
    \midrule
    Implicit-O & 45.3 & 18.9 & 3.04 & 49.1 & 22.6 & 5.69 & 29.5 & 10.7 & 10.2 \\
    UnO & 59.0 & 29.2 & 2.04 & 51.6 & 21.2 & 4.92 & 27.2 & 8.30 & 8.29 \\
    DIO & 57.6 & 26.2 & \textbf{1.43} & 58.0 & 31.5 & 3.01 & 36.2 & 13.8 & 4.40 \\
    \rowcolor{tablecolor}
    \ourmodel{} & \textbf{63.3} & \textbf{31.9} & 1.97 & \textbf{60.5} & \textbf{35.7} & \textbf{2.86} & \textbf{42.6} & \textbf{21.4} & \textbf{4.01} \\
    \bottomrule
    \end{tabular}}
\end{table*}

\begin{table*}[]
    \centering
    \footnotesize
    \setlength\tabcolsep{2pt}
    \caption{Ablation studies of \ourmodel{} on \nameourdataset{}.}
    \label{tab:ablation_long_range_split}
    \resizebox{0.8\textwidth}{!}{%
    \begin{tabular}{@{}l ccc ccc@{}}
    \toprule
     & \multicolumn{3}{c}{All} & \multicolumn{3}{c}{200m+} \\
    \cmidrule(lr){2-4} \cmidrule(lr){5-7}
    Method & mAP $\uparrow$ & Soft-IoU $\uparrow$ & EPE $\downarrow$ & mAP $\uparrow$ & Soft-IoU $\uparrow$ & EPE $\downarrow$ \\
    \midrule
    \ourmodel{} (no pretraining) & 59.7 & 33.9 & 4.09 & 41.0 & 19.8 & 4.70 \\
    \ourmodel{} (no radar input) & 57.7 & 31.9 & 3.04 & 34.5 & 14.3 & 4.59 \\
    \ourmodel{} (no camera input) & \textbf{60.5} & \textbf{35.7} & \textbf{2.68} & 42.2 & 20.9 & \textbf{3.64} \\
    \ourmodel{} (no radar loss) & \textbf{60.5} & 35.6 & 4.35 & \textbf{42.6} & 21.3 & 6.02 \\
    \rowcolor{tablecolor}
    \ourmodel{} & \textbf{60.5} & \textbf{35.7} & 2.86 & \textbf{42.6} & \textbf{21.4} & 4.01 \\
    \bottomrule
    \end{tabular}}
\end{table*}

\subsubsection{Limitations, Discussion, and Future Work.}
Currently, \ourmodel{} is limited to using RADAR for flow self-supervision, but future work could add occupancy supervision from RADAR. While our model uses RADAR to generalize occupancy predictions beyond the LiDAR range, RADAR can also be used as additional occupancy supervision in adverse weather to robustify predictions. This requires techniques beyond the line-of-sight assumption  done in \cite{Kung_2025_ICCV} due to nature of RADAR returns.
Furthermore, the ability to perceive and forecast the movement of rare objects is an understudied problem in the SDV space. We see a general need for additional larger scale benchmarks focusing on the task of \textit{Anything Perception} to ensure the overall robustness of many pre-existing systems.

\section{Conclusion}
We present \ourmodel{}, a novel unsupervised tri-modal occupancy model that learns to predict 4D occupancy, obstacle semantics, and flow by leveraging the complementary physical synergies of camera, LiDAR, and RADAR data. It achieves state-of-the-art performance across multiple tasks, including zero-shot anything perception, LiDAR forecasting, and 3D/4D occupancy prediction, while being unsupervised and independent of additional costly human annotations. 
We observe failure-cases of open-vocabulary methods for rare classes, and show that \ourmodel{} outperforms these methods in open-world scenarios. Notably, it achieves robust performance in perceiving anything relative to the drivable surface, including rare classes, showcasing its potential for a scalable deployment of SDVs.

\bibliographystyle{splncs04}
\bibliography{main}

@String(CVPR= {IEEE Conf. Comput. Vis. Pattern Recog.})

@String(ICCV= {Int. Conf. Comput. Vis.})

@String(ECCV= {Eur. Conf. Comput. Vis.})

@String(AAAI = {AAAI})

@String(CVPR  = {CVPR})

@String(ICCV  = {ICCV})

@String(ECCV  = {ECCV})

@InProceedings{Agro_2024_CVPR,
    author    = {Agro, Ben and Sykora, Quinlan and Casas, Sergio and Gilles, Thomas and Urtasun, Raquel},
    title     = {UnO: Unsupervised Occupancy Fields for Perception and Forecasting},
    booktitle = {Proceedings of the IEEE/CVF Conference on Computer Vision and Pattern Recognition (CVPR)},
    month     = {June},
    year      = {2024},
    pages     = {14487-14496}
}

@inproceedings{khurana2023point,
  title     = {Point Cloud Forecasting as a Proxy for 4D Occupancy Forecasting},
  author    = {Khurana, Tarasha and Hu, Peiyun and Held, David and Ramanan, Deva},
  booktitle = {Proceedings of the IEEE/CVF Conference on Computer Vision and Pattern Recognition},
  pages     = {1116--1124},
  year      = {2023}
}

@inproceedings{khurana2022differentiable,
  title        = {Differentiable raycasting for self-supervised occupancy forecasting},
  author       = {Khurana, Tarasha and Hu, Peiyun and Dave, Achal and Ziglar, Jason and Held, David and Ramanan, Deva},
  booktitle    = {European Conference on Computer Vision},
  pages        = {353--369},
  year         = {2022},
  organization = {Springer}
}

@inproceedings{agro2023implicit,
  title     = {Implicit Occupancy Flow Fields for Perception and Prediction in Self-Driving},
  author    = {Agro, Ben and Sykora, Quinlan and Casas, Sergio and Urtasun, Raquel},
  booktitle = {Proceedings of the IEEE/CVF Conference on Computer Vision and Pattern Recognition},
  pages     = {1379--1388},
  year      = {2023}
}

@InProceedings{diehl2023corl,
  title = 	 {Energy-based Potential Games for Joint Motion Forecasting and Control},
  author =       {Diehl, Christopher and Klosek, Tobias and Krueger, Martin and Murzyn, Nils and Osterburg, Timo and Bertram, Torsten},
  booktitle = 	 {Proceedings of The 7th Conference on Robot Learning},
  pages = 	 {3112--3141},
  year = 	 {2023},
}

@ARTICLE{diehl2023RAL,
  author={Diehl, Christopher and Sievernich, Timo Sebastian and Krüger, Martin and Hoffmann, Frank and Bertram, Torsten},
  journal={IEEE Robotics and Automation Letters}, 
  title={Uncertainty-Aware Model-Based Offline Reinforcement Learning for Automated Driving}, 
  year={2023},
  pages={1167-1174},
}

@article{mahjourian2022occupancy,
  title     = {Occupancy flow fields for motion forecasting in autonomous driving},
  author    = {Mahjourian, Reza and Kim, Jinkyu and Chai, Yuning and Tan, Mingxing and Sapp, Ben and Anguelov, Dragomir},
  journal   = {IEEE Robotics and Automation Letters},
  volume    = {7},
  number    = {2},
  pages     = {5639--5646},
  year      = {2022},
  publisher = {IEEE}
}

@inproceedings{casas2021mp3,
  title     = {Mp3: A unified model to map, perceive, predict and plan},
  author    = {Casas, Sergio and Sadat, Abbas and Urtasun, Raquel},
  booktitle = {Proceedings of the IEEE/CVF Conference on Computer Vision and Pattern Recognition},
  pages     = {14403--14412},
  year      = {2021}
}

@article{sscIJCV2021,
  author       = {Luis Roldao and
                  Raoul de Charette and
                  Anne Verroust{-}Blondet},
  title        = {3D Semantic Scene Completion: a Survey},
  booktitle    = {International Journal on Computer Vision},
  pages        = {1978–-2005},
  year         = {2022},
  organization = {Springer}
}

@article{wilson2023argoverse,
  title   = {Argoverse 2: Next generation datasets for self-driving perception and forecasting},
  author  = {Wilson, Benjamin and Qi, William and Agarwal, Tanmay and Lambert, John and Singh, Jagjeet and Khandelwal, Siddhesh and Pan, Bowen and Kumar, Ratnesh and Hartnett, Andrew and Pontes, Jhony Kaesemodel and others},
  journal = {arXiv preprint arXiv:2301.00493},
  year    = {2023}
}

@inproceedings{casas2018intentnet,
  title     = {Intentnet: Learning to predict intention from raw sensor data},
  author    = {Casas, Sergio and Luo, Wenjie and Urtasun, Raquel},
  booktitle = {CoRL},
  year      = {2018}
}

@inproceedings{Liang_2020_CVPR,
  author    = {Liang, Ming and Yang, Bin and Zeng, Wenyuan and Chen, Yun and Hu, Rui and Casas, Sergio and Urtasun, Raquel},
  title     = {PnPNet: End-to-End Perception and Prediction With Tracking in the Loop},
  booktitle = {CVPR},
  year      = {2020}
}

@article{weng2020ptp,
  title   = {Ptp: Parallelized tracking and prediction with graph neural networks and diversity sampling},
  author  = {Weng, Xinshuo and Yuan, Ye and Kitani, Kris},
  journal = {RA-L},
  year    = {2021}
}

@inproceedings{weng20203d,
  title     = {3d multi-object tracking: A baseline and new evaluation metrics},
  author    = {Weng, Xinshuo and Wang, Jianren and Held, David and Kitani, Kris},
  booktitle = {IROS},
  year      = {2020}
}

@inproceedings{salzmann2020trajectron,
  title     = {Trajectron++: Dynamically-feasible trajectory forecasting with heterogeneous data},
  author    = {Salzmann, Tim and Ivanovic, Boris and Chakravarty, Punarjay and Pavone, Marco},
  booktitle = {ECCV},
  year      = {2020}
}

@inproceedings{ivanovic2018the,
  title     = {The trajectron: Probabilistic multi-agent trajectory modeling with dynamic spatiotemporal graphs},
  author    = {Ivanovic, Boris and Pavone, Marco},
  booktitle = {ICCV},
  year      = {2019}
}

@article{ivanovic2020multimodal,
  title     = {Multimodal deep generative models for trajectory prediction: A conditional variational autoencoder approach},
  author    = {Ivanovic, Boris and Leung, Karen and Schmerling, Edward and Pavone, Marco},
  journal   = {RA-L},
  year      = {2020},
  publisher = {IEEE}
}

@inproceedings{cui2021lookout,
  title     = {Lookout: Diverse multi-future prediction and planning for self-driving},
  author    = {Cui, Alexander and Casas, Sergio and Sadat, Abbas and Liao, Renjie and Urtasun, Raquel},
  booktitle = {ICCV},
  year      = {2021}
}

@inproceedings{sadat2020perceive,
  title     = {Perceive, predict, and plan: Safe motion planning through interpretable semantic representations},
  author    = {Sadat, Abbas and Casas, Sergio and Ren, Mengye and Wu, Xinyu and Dhawan, Pranaab and Urtasun, Raquel},
  booktitle = {ECCV},
  year      = {2020}
}

@inproceedings{philion2020lift,
  title     = {Lift, splat, shoot: Encoding images from arbitrary camera rigs by implicitly unprojecting to 3d},
  author    = {Philion, Jonah and Fidler, Sanja},
  booktitle = {ECCV},
  year      = {2020}
}

@inproceedings{hu2021fiery,
  title     = {FIERY: Future Instance Prediction in Bird's-Eye View From Surround Monocular Cameras},
  author    = {Hu, Anthony and Murez, Zak and Mohan, Nikhil and Dudas, Sof{\'\i}a and Hawke, Jeffrey and Badrinarayanan, Vijay and Cipolla, Roberto and Kendall, Alex},
  booktitle = {ICCV},
  year      = {2021}
}

@inproceedings{hess2023masked,
  title     = {Masked autoencoder for self-supervised pre-training on lidar point clouds},
  author    = {Hess, Georg and Jaxing, Johan and Svensson, Elias and Hagerman, David and Petersson, Christoffer and Svensson, Lennart},
  booktitle = {Proceedings of the IEEE/CVF Winter Conference on Applications of Computer Vision},
  pages     = {350--359},
  year      = {2023}
}

@article{yang2023unipad,
  title   = {UniPAD: A Universal Pre-training Paradigm for Autonomous Driving},
  author  = {Yang, Honghui and Zhang, Sha and Huang, Di and Wu, Xiaoyang and Zhu, Haoyi and He, Tong and Tang, Shixiang and Zhao, Hengshuang and Qiu, Qibo and Lin, Binbin and others},
  journal = {arXiv preprint arXiv:2310.08370},
  year    = {2023}
}

@inproceedings{he2016deep,
  title     = {Deep residual learning for image recognition},
  author    = {He, Kaiming and Zhang, Xiangyu and Ren, Shaoqing and Sun, Jian},
  booktitle = {Proceedings of the IEEE conference on computer vision and pattern recognition},
  pages     = {770--778},
  year      = {2016}
}

@inproceedings{lin2017feature,
  title     = {Feature pyramid networks for object detection},
  author    = {Lin, Tsung-Yi and Doll{\'a}r, Piotr and Girshick, Ross and He, Kaiming and Hariharan, Bharath and Belongie, Serge},
  booktitle = {Proceedings of the IEEE conference on computer vision and pattern recognition},
  pages     = {2117--2125},
  year      = {2017}
}

@InProceedings{Kirillov_2023_ICCV,
    author    = {Kirillov, Alexander and Mintun, Eric and Ravi, Nikhila and Mao, Hanzi and Rolland, Chloe and Gustafson, Laura and Xiao, Tete and Whitehead, Spencer and Berg, Alexander C. and Lo, Wan-Yen and Dollar, Piotr and Girshick, Ross},
    title     = {Segment Anything},
    booktitle = {Proceedings of the IEEE/CVF International Conference on Computer Vision (ICCV)},
    month     = {October},
    year      = {2023},
    pages     = {4015-4026}
}

@INPROCEEDINGS{Diehl2020,
  author={Diehl, Christopher and Feicho, Eduard and Schwambach, Alexander and Dammeier, Thomas and Mares, Eric and Bertram, Torsten},
  booktitle={2020 IEEE 23rd International Conference on Intelligent Transportation Systems (ITSC)}, 
  title={Radar-based Dynamic Occupancy Grid Mapping and Object Detection}, 
  year={2020},
  volume={},
  number={},
  pages={1-6},
}

@inproceedings{RadarOcc2024,
 author = {Ding, Fangqiang and Wen, Xiangyu and Zhu, Yunzhou and Li, Yiming and Lu, Chris Xiaoxuan},
 booktitle = {Advances in Neural Information Processing Systems},
 doi = {10.52202/079017-3222},
 editor = {A. Globerson and L. Mackey and D. Belgrave and A. Fan and U. Paquet and J. Tomczak and C. Zhang},
 pages = {101589--101617},
 publisher = {Curran Associates, Inc.},
 title = {RadarOcc: Robust 3D Occupancy Prediction with 4D Imaging Radar},
 url = {https://proceedings.neurips.cc/paper_files/paper/2024/file/b81d83165e3145a2e7d33bb5e33ea913-Paper-Conference.pdf},
 volume = {37},
 year = {2024}
}

@inproceedings{QuAD2024,
      title={QuAD: Query-based Interpretable Neural Motion Planning for Autonomous Driving}, 
      author={Sourav Biswas and Sergio Casas and Quinlan Sykora and Ben Agro and Abbas Sadat and Raquel Urtasun},
      booktitle={ICRA},
      year={2022}, 
}

@misc{gu2024dometamingdiffusionmodel,
      title={DOME: Taming Diffusion Model into High-Fidelity Controllable Occupancy World Model}, 
      author={Songen Gu and Wei Yin and Bu Jin and Xiaoyang Guo and Junming Wang and Haodong Li and Qian Zhang and Xiaoxiao Long},
      year={2024},
      eprint={2410.10429},
      archivePrefix={arXiv},
      primaryClass={cs.CV},
      url={https://arxiv.org/abs/2410.10429}, 
}

@inproceedings{ma2024cvpr,
	author = {Junyi Ma and Xieyuanli Chen and Jiawei Huang and Jingyi Xu and Zhen Luo and Jintao Xu and Weihao Gu and Rui Ai and Hesheng Wang},
	title = {{Cam4DOcc: Benchmark for Camera-Only 4D Occupancy Forecasting in Autonomous Driving Applications}},
	booktitle = {Proc.~of the IEEE/CVF Conf.~on Computer Vision and Pattern Recognition (CVPR)},
	year = {2024}
}

@inproceedings{zheng2024eccv,
    title={OccWorld: Learning a 3D Occupancy World Model for Autonomous Driving},
    author={Zheng, Wenzhao and Chen, Weiliang and Huang, Yuanhui and Zhang, Borui and Duan, Yueqi and Lu, Jiwen },
    booktitle={ECCV},
    year={2024}, 
}

@misc{guo2024fsfnetenhance4doccupancy,
      title={FSF-Net: Enhance 4D Occupancy Forecasting with Coarse BEV Scene Flow for Autonomous Driving}, 
      author={Erxin Guo and Pei An and You Yang and Qiong Liu and An-An Liu},
      year={2024},
      eprint={2409.15841},
      archivePrefix={arXiv},
      primaryClass={cs.CV},
      url={https://arxiv.org/abs/2409.15841}, 
}

@INPROCEEDINGS{DeTRA2024,
      title={DeTra: A Unified Model for Object Detection and Trajectory Forecasting}, 
      author={Sergio Casas and Ben Agro and Jiageng Mao and Thomas Gilles and Alexander Cui and Thomas Li and Raquel Urtasun},
      year={2024},
      booktitle={ECCV},
}

@INPROCEEDINGS{PCN2018,
  author={Yuan, Wentao and Khot, Tejas and Held, David and Mertz, Christoph and Hebert, Martial},
  booktitle={2018 International Conference on 3D Vision (3DV)}, 
  title={PCN: Point Completion Network}, 
  year={2018},
  volume={},
  number={},
  pages={728-737},
}

@inproceedings{LMP2006,
author = {Nealen, Andrew and Igarashi, Takeo and Sorkine, Olga and Alexa, Marc},
title = {Laplacian mesh optimization},
year = {2006},
booktitle = {Proceedings of the 4th International Conference on Computer Graphics and Interactive Techniques in Australasia and Southeast Asia},
pages = {381–389},
numpages = {9},

}

@inproceedings{PossionSurface2006,
author = {Kazhdan, Michael and Bolitho, Matthew and Hoppe, Hugues},
title = {Poisson surface reconstruction},
year = {2006},
publisher = {Eurographics Association},
address = {Goslar, DEU},
booktitle = {Proceedings of the Fourth Eurographics Symposium on Geometry Processing},
pages = {61–70},
numpages = {10},
}

@INPROCEEDINGS{Zimmermann2017,
  author={Zimmermann, Karel and Petrícek, Tomáš and Šalanský, Vojtech and Svoboda, Tomáš},
  booktitle={2017 IEEE International Conference on Computer Vision (ICCV)}, 
  title={Learning for Active 3D Mapping}, 
  year={2017},
  volume={},
  number={},
  pages={1548-1556},}

@INPROCEEDINGS{ScanNet2017,
  author={Dai, Angela and Chang, Angel X. and Savva, Manolis and Halber, Maciej and Funkhouser, Thomas and Nießner, Matthias},
  booktitle={2017 IEEE Conference on Computer Vision and Pattern Recognition (CVPR)}, 
  title={ScanNet: Richly-Annotated 3D Reconstructions of Indoor Scenes}, 
  year={2017},
  volume={},
  number={},
  pages={2432-2443},
}

@inproceedings{yan2021sparse,
  title={Sparse Single Sweep LiDAR Point Cloud Segmentation via Learning Contextual Shape Priors from Scene Completion},
  author={Yan, Xu and Gao, Jiantao and Li, Jie and Zhang, Ruimao and Li, Zhen and Huang, Rui and Cui, Shuguang},
  booktitle={Proceedings of the AAAI Conference on Artificial Intelligence},
  volume={35},
  number={4},
  pages={3101--3109},
  year={2021}
}

@INPROCEEDINGS{SemanticKITTI,
  author={Behley, Jens and Garbade, Martin and Milioto, Andres and Quenzel, Jan and Behnke, Sven and Stachniss, Cyrill and Gall, Jürgen},
  booktitle={2019 IEEE/CVF International Conference on Computer Vision (ICCV)}, 
  title={SemanticKITTI: A Dataset for Semantic Scene Understanding of LiDAR Sequences}, 
  year={2019},
  volume={},
  number={},
  pages={9296-9306},
}

@article{SubmanifoldSparseConvNet,
  title={Submanifold Sparse Convolutional Networks},
  author={Graham, Benjamin and van der Maaten, Laurens},
  journal={arXiv preprint arXiv:1706.01307},
  year={2017}
}

@misc{zhu2018deformableconvnetsv2deformable,
      title={Deformable ConvNets v2: More Deformable, Better Results}, 
      author={Xizhou Zhu and Han Hu and Stephen Lin and Jifeng Dai},
      year={2018},
      eprint={1811.11168},
      archivePrefix={arXiv},
      primaryClass={cs.CV},
      url={https://arxiv.org/abs/1811.11168}, 
}

@misc{liu2024selfsupervisedmultiframeneuralscene,
      title={Self-Supervised Multi-Frame Neural Scene Flow}, 
      author={Dongrui Liu and Daqi Liu and Xueqian Li and Sihao Lin and Hongwei xie and Bing Wang and Xiaojun Chang and Lei Chu},
      year={2024},
      eprint={2403.16116},
      archivePrefix={arXiv},
      primaryClass={cs.CV},
      url={https://arxiv.org/abs/2403.16116}, 
}

@ARTICLE{Flow4DRAL2025,

  author={Kim, Jaeyeul and Woo, Jungwan and Shin, Ukcheol and Oh, Jean and Im, Sunghoon},

  journal={IEEE Robotics and Automation Letters}, 

  title={Flow4D: Leveraging 4D Voxel Network for LiDAR Scene Flow Estimation}, 

  year={2025},

  volume={},

  number={},

  pages={1-8},
}

@InProceedings{Zuo_2025_CVPR,
    author    = {Zuo, Sicheng and Zheng, Wenzhao and Huang, Yuanhui and Zhou, Jie and Lu, Jiwen},
    title     = {GaussianWorld: Gaussian World Model for Streaming 3D Occupancy Prediction},
    booktitle = {Proceedings of the IEEE/CVF Conference on Computer Vision and Pattern Recognition (CVPR)},
    month     = {June},
    year      = {2025},
    pages     = {6772-6781}
}

@inproceedings{NEURIPS2023_Occ3D,
 author = {Tian, Xiaoyu and Jiang, Tao and Yun, Longfei and Mao, Yucheng and Yang, Huitong and Wang, Yue and Wang, Yilun and Zhao, Hang},
 booktitle = {Advances in Neural Information Processing Systems},
 editor = {A. Oh and T. Naumann and A. Globerson and K. Saenko and M. Hardt and S. Levine},
 pages = {64318--64330},

 title = {Occ3D: A Large-Scale 3D Occupancy Prediction Benchmark for Autonomous Driving},
 volume = {36},
 year = {2023}
}

@InProceedings{OpenOccupancy_2023_ICCV,
    author    = {Wang, Xiaofeng and Zhu, Zheng and Xu, Wenbo and Zhang, Yunpeng and Wei, Yi and Chi, Xu and Ye, Yun and Du, Dalong and Lu, Jiwen and Wang, Xingang},
    title     = {OpenOccupancy: A Large Scale Benchmark for Surrounding Semantic Occupancy Perception},
    booktitle = {Proceedings of the IEEE/CVF International Conference on Computer Vision (ICCV)},
    month     = {October},
    year      = {2023},
    pages     = {17850-17859}
}

@InProceedings{Boeder_2025_ICCV,
    author    = {Boeder, Simon and Gigengack, Fabian and Risse, Benjamin},
    title     = {GaussianFlowOcc: Sparse and Weakly Supervised Occupancy Estimation using Gaussian Splatting and Temporal Flow},
    booktitle = {Proceedings of the IEEE/CVF International Conference on Computer Vision (ICCV)},
    month     = {October},
    year      = {2025},
    pages     = {24943-24954}
}

@misc{GASP_2025,
      title={GASP: Unifying Geometric and Semantic Self-Supervised Pre-training for Autonomous Driving}, 
      author={William Ljungbergh and Adam Lilja and Adam Tonderski. Arvid Laveno Ling and Carl Lindström and Willem Verbeke and Junsheng Fu and Christoffer Petersson and Lars Hammarstrand and Michael Felsberg},
      year={2025},
      eprint={2503.15672},
      archivePrefix={arXiv},
      primaryClass={cs.CV},
      url={https://arxiv.org/abs/2503.15672}, 
}

@misc{QueryOcc_2025,
      title={QueryOcc: Query-based Self-Supervision for 3D Semantic Occupancy}, 
      author={Adam Lilja and Ji Lan and Junsheng Fu and Lars Hammarstrand},
      year={2025},
      eprint={2511.17221},
      archivePrefix={arXiv},
      primaryClass={cs.CV},
      url={https://arxiv.org/abs/2511.17221}, 
}

@INPROCEEDINGS{SelfOcc_24,
  author={Huang, Yuanhui and Zheng, Wenzhao and Zhang, Borui and Zhou, Jie and Lu, Jiwen},
  booktitle={2024 IEEE/CVF Conference on Computer Vision and Pattern Recognition (CVPR)}, 
  title={SelfOcc: Self-Supervised Vision-Based 3D Occupancy Prediction}, 
  year={2024},
  volume={},
  number={},
  pages={19946-19956},
  doi={10.1109/CVPR52733.2024.01885}}

@InProceedings{Li_2025_ICCV,
    author    = {Li, Peizheng and Ding, Shuxiao and Zhou, You and Zhang, Qingwen and Inak, Onat and Triess, Larissa and Hanselmann, Niklas and Cordts, Marius and Zell, Andreas},
    title     = {AGO: Adaptive Grounding for Open World 3D Occupancy Prediction},
    booktitle = {Proceedings of the IEEE/CVF International Conference on Computer Vision (ICCV)},
    month     = {October},
    year      = {2025},
    pages     = {8645-8655}
}

@misc{ShelfOcc2025,
      title={ShelfOcc: Native 3D Supervision beyond LiDAR for Vision-Based Occupancy Estimation}, 
      author={Simon Boeder and Fabian Gigengack and Simon Roesler and Holger Caesar and Benjamin Risse},
      year={2025},
      eprint={2511.15396},
      archivePrefix={arXiv},
      primaryClass={cs.CV},
      url={https://arxiv.org/abs/2511.15396}, 
}

@InProceedings{GaussianOcc_Gan_2025_ICCV,
    author    = {Gan, Wanshui and Liu, Fang and Xu, Hongbin and Mo, Ningkai and Yokoya, Naoto},
    title     = {GaussianOcc: Fully Self-supervised and Efficient 3D Occupancy Estimation with Gaussian Splatting},
    booktitle = {Proceedings of the IEEE/CVF International Conference on Computer Vision (ICCV)},
    month     = {October},
    year      = {2025},
    pages     = {28980-28990}
}

@inproceedings{
anonymous2026fantastic,
title={Fantastic Tractor-Dogs and How Not to Find Them With Open-Vocabulary Detectors},
author={Anonymous},
booktitle={The Fourteenth International Conference on Learning Representations},
year={2026},
url={https://openreview.net/forum?id=jUuXNrG7wh}
}

@InProceedings{Saric_2025_ICCV,
    author    = {\v{S}ari\'c, Josip and Martinovi\'c, Ivan and Kristan, Matej and \v{S}egvi\'c, Sini\v{s}a},
    title     = {What Holds Back Open-Vocabulary Segmentation?},
    booktitle = {Proceedings of the IEEE/CVF International Conference on Computer Vision (ICCV) Workshops},
    month     = {October},
    year      = {2025},
    pages     = {4256-4266}
}

@misc{gaussianfusionoccseamlesssensorfusion,
      title={GaussianFusionOcc: A Seamless Sensor Fusion Approach for 3D Occupancy Prediction Using 3D Gaussians}, 
      author={Tomislav Pavković and Mohammad-Ali Nikouei Mahani and Johannes Niedermayer and Johannes Betz},
      year={2025},
      eprint={2507.18522},
      archivePrefix={arXiv},
      primaryClass={cs.CV},
      url={https://arxiv.org/abs/2507.18522}, 
}

@INPROCEEDINGS{4DRolls_IROS2025,
  author={Liu, Ruihan and Wu, Xiaoyi and Chen, Xijun and Hu, Liang and Lou, Yunjiang},
  booktitle={2025 IEEE/RSJ International Conference on Intelligent Robots and Systems (IROS)}, 
  title={4D-ROLLS: 4D Radar Occupancy Learning via LiDAR Supervision}, 
  year={2025},
  volume={},
  number={},
  pages={13602-13609},
  doi={10.1109/IROS60139.2025.11246471}}

@InProceedings{Kung_2025_ICCV,
    author    = {Kung, Pou-Chun and Harisha, Skanda and Vasudevan, Ram and Eid, Aline and Skinner, Katherine A.},
    title     = {RadarSplat: Radar Gaussian Splatting for High-Fidelity Data Synthesis and 3D Reconstruction of Autonomous Driving Scenes},
    booktitle = {Proceedings of the IEEE/CVF International Conference on Computer Vision (ICCV)},
    month     = {October},
    year      = {2025},
    pages     = {27596-27606}
}

@InProceedings{Palladin_2025_ICCV,
    author    = {Palladin, Edoardo and Brucker, Samuel and Ghilotti, Filippo and Narayanan, Praveen and Bijelic, Mario and Heide, Felix},
    title     = {Self-Supervised Sparse Sensor Fusion for Long Range Perception},
    booktitle = {Proceedings of the IEEE/CVF International Conference on Computer Vision (ICCV)},
    month     = {October},
    year      = {2025},
    pages     = {27498-27509}
}

@INPROCEEDINGS{Diehl_2025_DIO,
  author={Diehl, Christopher and Sykora, Quinlan and Agro, Ben and Gilles, Thomas and Casas, Sergio and Urtasun, Raquel},
  booktitle={2025 IEEE/CVF Conference on Computer Vision and Pattern Recognition (CVPR)}, 
  title={DIO: Decomposable Implicit 4D Occupancy-Flow World Model}, 
  year={2025},
  volume={},
  number={},
  pages={27456-27466},
  doi={10.1109/CVPR52734.2025.02557}}

@misc{maskclippp25,
      title={High-Quality Mask Tuning Matters for Open-Vocabulary Segmentation}, 
      author={Quan-Sheng Zeng and Yunheng Li and Daquan Zhou and Guanbin Li and Qibin Hou and Ming-Ming Cheng},
      year={2025},
      eprint={2412.11464},
      archivePrefix={arXiv},
      primaryClass={cs.CV},
      url={https://arxiv.org/abs/2412.11464}, 
}

@misc{groundedsam24,
      title={Grounded SAM: Assembling Open-World Models for Diverse Visual Tasks}, 
      author={Tianhe Ren and Shilong Liu and Ailing Zeng and Jing Lin and Kunchang Li and He Cao and Jiayu Chen and Xinyu Huang and Yukang Chen and Feng Yan and Zhaoyang Zeng and Hao Zhang and Feng Li and Jie Yang and Hongyang Li and Qing Jiang and Lei Zhang},
      year={2024},
      eprint={2401.14159},
      archivePrefix={arXiv},
      primaryClass={cs.CV},
      url={https://arxiv.org/abs/2401.14159}, 
}

@InProceedings{Talk2Dino_2025_ICCV,
    author    = {Barsellotti, Luca and Bianchi, Lorenzo and Messina, Nicola and Carrara, Fabio and Cornia, Marcella and Baraldi, Lorenzo and Falchi, Fabrizio and Cucchiara, Rita},
    title     = {Talking to DINO: Bridging Self-Supervised Vision Backbones with Language for Open-Vocabulary Segmentation},
    booktitle = {Proceedings of the IEEE/CVF International Conference on Computer Vision (ICCV)},
    month     = {October},
    year      = {2025},
    pages     = {22025-22035}
}

@InProceedings{DinoText_2025_CVPR,
    author    = {Jose, Cijo and Moutakanni, Th\'eo and Kang, Dahyun and Baldassarre, Federico and Darcet, Timoth\'ee and Xu, Hu and Li, Daniel and Szafraniec, Marc and Ramamonjisoa, Micha\"el and Oquab, Maxime and Sim\'eoni, Oriane and Vo, Huy V. and Labatut, Patrick and Bojanowski, Piotr},
    title     = {DINOv2 Meets Text: A Unified Framework for Image- and Pixel-Level Vision-Language Alignment},
    booktitle = {Proceedings of the IEEE/CVF Conference on Computer Vision and Pattern Recognition (CVPR)},
    month     = {June},
    year      = {2025},
    pages     = {24905-24916}
}

@misc{siglip2_2025,
      title={SigLIP 2: Multilingual Vision-Language Encoders with Improved Semantic Understanding, Localization, and Dense Features}, 
      author={Michael Tschannen and Alexey Gritsenko and Xiao Wang and Muhammad Ferjad Naeem and Ibrahim Alabdulmohsin and Nikhil Parthasarathy and Talfan Evans and Lucas Beyer and Ye Xia and Basil Mustafa and Olivier Hénaff and Jeremiah Harmsen and Andreas Steiner and Xiaohua Zhai},
      year={2025},
      eprint={2502.14786},
      archivePrefix={arXiv},
      primaryClass={cs.CV},
      url={https://arxiv.org/abs/2502.14786}, 
}

@inproceedings{FCCLIP_NIPS23,
 author = {Yu, Qihang and He, Ju and Deng, Xueqing and Shen, Xiaohui and Chen, Liang-Chieh},
 booktitle = {Advances in Neural Information Processing Systems},
 editor = {A. Oh and T. Naumann and A. Globerson and K. Saenko and M. Hardt and S. Levine},
 pages = {32215--32234},
 publisher = {Curran Associates, Inc.},
 title = {Convolutions Die Hard: Open-Vocabulary Segmentation with Single Frozen Convolutional CLIP},
 url = {https://proceedings.neurips.cc/paper_files/paper/2023/file/661caac7729aa7d8c6b8ac0d39ccbc6a-Paper-Conference.pdf},
 volume = {36},
 year = {2023}
}

@inproceedings{MAFTp_ECCV2024,
  title={Collaborative Vision-Text Representation Optimizing for Open-Vocabulary Segmentation},
  author={Jiao, Siyu and Zhu, Hongguang and Huang, Jiannan and Zhao, Yao and Wei, Yunchao and Humphrey, Shi},
  booktitle={European Conference on Computer Vision},
  year={2024},
}

@article{hu2024metric3dv2,
  title={Metric3d v2: A versatile monocular geometric foundation model for zero-shot metric depth and surface normal estimation},
  author={Hu, Mu and Yin, Wei and Zhang, Chi and Cai, Zhipeng and Long, Xiaoxiao and Chen, Hao and Wang, Kaixuan and Yu, Gang and Shen, Chunhua and Shen, Shaojie},
  journal={IEEE Transactions on Pattern Analysis and Machine Intelligence},
  year={2024},
  publisher={IEEE}
}

@InProceedings{Jevtic_2025_ICCV,
    author    = {Jevti\'c, Aleksandar and Reich, Christoph and Wimbauer, Felix and Hahn, Oliver and Rupprecht, Christian and Roth, Stefan and Cremers, Daniel},
    title     = {Feed-Forward SceneDINO for Unsupervised Semantic Scene Completion},
    booktitle = {Proceedings of the IEEE/CVF International Conference on Computer Vision (ICCV)},
    month     = {October},
    year      = {2025},
    pages     = {6784-6796}
}

@INPROCEEDINGS{MonoScene,
  author={Cao, Anh-Quan and de Charette, Raoul},
  booktitle={2022 IEEE/CVF Conference on Computer Vision and Pattern Recognition (CVPR)}, 
  title={MonoScene: Monocular 3D Semantic Scene Completion}, 
  year={2022},
  volume={},
  number={},
  pages={3981-3991},
  doi={10.1109/CVPR52688.2022.00396}}

@ARTICLE{OccFusion25,
  author={Ming, Zhenxing and Stephany Berrio, Julie and Shan, Mao and Worrall, Stewart},
  journal={IEEE Transactions on Intelligent Vehicles}, 
  title={OccFusion: Multi-Sensor Fusion Framework for 3D Semantic Occupancy Prediction}, 
  year={2025},
  volume={10},
  number={5},
  pages={3421-3433},
  doi={10.1109/TIV.2024.3453293}}

@misc{OccNerf23,
      title={OccNeRF: Advancing 3D Occupancy Prediction in LiDAR-Free Environments}, 
      author={Chubin Zhang and Juncheng Yan and Yi Wei and Jiaxin Li and Li Liu and Yansong Tang and Yueqi Duan and Jiwen Lu},
      year={2024},
      eprint={2312.09243},
      archivePrefix={arXiv},
      primaryClass={cs.CV},
      url={https://arxiv.org/abs/2312.09243}, 
}

@misc{zhao2025shelfgaussian,
      title={ShelfGaussian: Shelf-Supervised Open-Vocabulary Gaussian-based 3D Scene Understanding}, 
      author={Lingjun Zhao and Yandong Luo and James Hay and Lu Gan},
      year={2025},
      eprint={2512.03370},
      archivePrefix={arXiv},
      primaryClass={cs.CV},
      url={https://arxiv.org/abs/2512.03370}, 
}

@InProceedings{Li_2023_CVPR,
    author    = {Li, Yiming and Yu, Zhiding and Choy, Christopher and Xiao, Chaowei and Alvarez, Jose M. and Fidler, Sanja and Feng, Chen and Anandkumar, Anima},
    title     = {VoxFormer: Sparse Voxel Transformer for Camera-Based 3D Semantic Scene Completion},
    booktitle = {Proceedings of the IEEE/CVF Conference on Computer Vision and Pattern Recognition (CVPR)},
    month     = {June},
    year      = {2023},
    pages     = {9087-9098}
}

@INPROCEEDINGS{STU_cvpr2025,
  author={Nekrasov, Alexey and Burdorf, Malcolm and Worrall, Stewart and Leibe, Bastian and Berrio Perez, Julie Stephany},
  booktitle={2025 IEEE/CVF Conference on Computer Vision and Pattern Recognition (CVPR)}, 
  title={Spotting the Unexpected (STU): A 3D LiDAR Dataset for Anomaly Segmentation in Autonomous Driving}, 
  year={2025},
  volume={},
  number={},
  pages={11875-11885},
  doi={10.1109/CVPR52734.2025.01109}}

@inproceedings{sal2024eccv,
    title={Better Call SAL: Towards Learning to Segment Anything in Lidar},
    author={Osep, Aljosa and Meinhardt, Tim and Ferroni, Francesco and Peri, 
                Neehar and Ramanan, Deva and Leal-TaixÃ©, Laura},
    booktitle={European Conference on Computer Vision (ECCV)},
    year={2024},
}

@inproceedings{groundingdino2024eccv,
  title={Grounding dino: Marrying dino with grounded pre-training for open-set object detection},
  author={Liu, Shilong and Zeng, Zhaoyang and Ren, Tianhe and Li, Feng and Zhang, Hao and Yang, Jie and Li, Chunyuan and Yang, Jianwei and Su, Hang and Zhu, Jun and others},
  booktitle={European Conference on Computer Vision (ECCV)},
    year={2024},
}

@inproceedings{OWLv2neurips23,
 author = {Minderer, Matthias and Gritsenko, Alexey and Houlsby, Neil},
 booktitle = {Advances in Neural Information Processing Systems},
 editor = {A. Oh and T. Naumann and A. Globerson and K. Saenko and M. Hardt and S. Levine},
 pages = {72983--73007},
 publisher = {Curran Associates, Inc.},
 title = {Scaling Open-Vocabulary Object Detection},
 url = {https://proceedings.neurips.cc/paper_files/paper/2023/file/e6d58fc68c0f3c36ae6e0e64478a69c0-Paper-Conference.pdf},
 volume = {36},
 year = {2023}
}

@INPROCEEDINGS{lidar_weaknesses,
  author={Uttarkabat, Satarupa and Appukuttan, Sarath and Gupta, Kwanit and Nayak, Satyajit and Palo, Patitapaban},
  booktitle={2024 IEEE Intelligent Vehicles Symposium (IV)}, 
  title={BloomNet: Perception of Blooming Effect in ADAS using Synthetic LiDAR Point Cloud Data}, 
  year={2024},
  volume={},
  number={},
  pages={1886-1892},
  doi={10.1109/IV55156.2024.10588461}}

@inproceedings{rife2025lidar,
  author    = {Jason H. Rife and Yifan Li},
  title     = {Characterizing Lidar Range-Measurement Ambiguity due to Multiple Returns},
  booktitle = {Proceedings of the 38th International Technical Meeting of the Satellite Division of The Institute of Navigation (ION GNSS+ 2025)},
  year      = {2025},
  address   = {Baltimore, Maryland},
  month     = sep,
  pages     = {1949--1963},
  doi       = {10.33012/2025.20444},
  url       = {https://doi.org/10.33012/2025.20444}
}

@Article{radar_in_rain,
AUTHOR = {Li, Xinda and Cheng, Xu and Ju, Xinjie and Peng, Yunli and Hu, Jinzhu and Li, Jianbing},
TITLE = {Optimization on the Polarization and Waveform of Radar for Better Target Detection Performance under Rainy Condition},
JOURNAL = {Remote Sensing},
VOLUME = {16},
YEAR = {2024},
NUMBER = {14},
ARTICLE-NUMBER = {2557},
URL = {https://www.mdpi.com/2072-4292/16/14/2557},
ISSN = {2072-4292},
DOI = {10.3390/rs16142557}
}

@INPROCEEDINGS{patchworkpp,
  author={Lee, Seungjae and Lim, Hyungtae and Myung, Hyun},
  booktitle={2022 IEEE/RSJ International Conference on Intelligent Robots and Systems (IROS)}, 
  title={Patchwork++: Fast and Robust Ground Segmentation Solving Partial Under-Segmentation Using 3D Point Cloud}, 
  year={2022},
  volume={},
  number={},
  pages={13276-13283},
  doi={10.1109/IROS47612.2022.9981561}}

@article{Nuss2017,
author = {Dominik Nuss and Stephan Reuter and Markus Thom and Ting Yuan and Gunther Krehl and Michael Maile and Axel Gern and Klaus Dietmayer},
title ={A random finite set approach for dynamic occupancy grid maps with real-time application},

journal = {The International Journal of Robotics Research},
volume = {37},
number = {8},
pages = {841-866},
year = {2018},
doi = {10.1177/0278364918775523},

URL = { 
    
        https://doi.org/10.1177/0278364918775523
    
    

},
eprint = { 
    
        https://doi.org/10.1177/0278364918775523
    
    

}
}

@INPROCEEDINGS{DOGMA_ICRA14,
  author={Tanzmeister, Georg and Thomas, Julian and Wollherr, Dirk and Buss, Martin},
  booktitle={2014 IEEE International Conference on Robotics and Automation (ICRA)}, 
  title={Grid-based mapping and tracking in dynamic environments using a uniform evidential environment representation}, 
  year={2014},
  volume={},
  number={},
  pages={6090-6095},
  doi={10.1109/ICRA.2014.6907756}}

@misc{CLIP_2021,
      title={Learning Transferable Visual Models From Natural Language Supervision}, 
      author={Alec Radford and Jong Wook Kim and Chris Hallacy and Aditya Ramesh and Gabriel Goh and Sandhini Agarwal and Girish Sastry and Amanda Askell and Pamela Mishkin and Jack Clark and Gretchen Krueger and Ilya Sutskever},
      year={2021},
      eprint={2103.00020},
      archivePrefix={arXiv},
      primaryClass={cs.CV},
      url={https://arxiv.org/abs/2103.00020}, 
}

@InProceedings{GausTR_2025_CVPR,
    author    = {Jiang, Haoyi and Liu, Liu and Cheng, Tianheng and Wang, Xinjie and Lin, Tianwei and Su, Zhizhong and Liu, Wenyu and Wang, Xinggang},
    title     = {GaussTR: Foundation Model-Aligned Gaussian Transformer for Self-Supervised 3D Spatial Understanding},
    booktitle = {Proceedings of the IEEE/CVF Conference on Computer Vision and Pattern Recognition (CVPR)},
    month     = {June},
    year      = {2025},
    pages     = {11960-11970}
}

@inproceedings{
wang2024distillnerf,
title={DistillNe{RF}: Perceiving 3D Scenes from Single-Glance Images by Distilling Neural Fields and Foundation Model Features},
author={Letian Wang and Seung Wook Kim and Jiawei Yang and Cunjun Yu and Boris Ivanovic and Steven L. Waslander and Yue Wang and Sanja Fidler and Marco Pavone and Peter Karkus},
booktitle={The Thirty-eighth Annual Conference on Neural Information Processing Systems},
year={2024},
}

@misc{adjepa2025,
      title={Self-Supervised Representation Learning with Joint Embedding Predictive Architecture for Automotive LiDAR Object Detection}, 
      author={Haoran Zhu and Zhenyuan Dong and Kristi Topollai and Beiyao Sha and Anna Choromanska},
      year={2025},
      eprint={2501.04969},
      archivePrefix={arXiv},
      primaryClass={cs.RO},
      url={https://arxiv.org/abs/2501.04969}, 
}

@INPROCEEDINGS{RadarSSL_CVPR24,
  author={Hao, Yiduo and Madani, Sohrab and Guan, Junfeng and Alloulah, Mohammed and Gupta, Saurabh and Hassanieh, Haitham},
  booktitle={2024 IEEE/CVF Conference on Computer Vision and Pattern Recognition (CVPR)}, 
  title={Bootstrapping Autonomous Driving Radars with Self-Supervised Learning}, 
  year={2024},
  volume={},
  number={},
  pages={15012-15023},
  doi={10.1109/CVPR52733.2024.01422}}

@ARTICLE{Dual_Evidental_TITV__24,
  author={Richter, Sven and Bieder, Frank and Wirges, Sascha and Stiller, Christoph},
  journal={IEEE Transactions on Intelligent Vehicles}, 
  title={A Dual Evidential Top-View Representation to Model the Semantic Environment of Automated Vehicles}, 
  year={2024},
  volume={9},
  number={1},
  pages={2688-2700},
  doi={10.1109/TIV.2023.3284400}}

@ARTICLE{GroundGrid2024,
  author={Steinke, Nicolai and Goehring, Daniel and Rojas, Raúl},
  journal={IEEE Robotics and Automation Letters}, 
  title={GroundGrid: LiDAR Point Cloud Ground Segmentation and Terrain Estimation}, 
  year={2024},
  volume={9},
  number={1},
  pages={420-426},
  doi={10.1109/LRA.2023.3333233}}

@INPROCEEDINGS{GndNet2020,
  author={Paigwar, Anshul and Erkent, Oezgür and Sierra-Gonzalez, David and Laugier, Christian},
  booktitle={2020 IEEE/RSJ International Conference on Intelligent Robots and Systems (IROS)}, 
  title={GndNet: Fast Ground Plane Estimation and Point Cloud Segmentation for Autonomous Vehicles}, 
  year={2020},
  volume={},
  number={},
  pages={2150-2156},
  doi={10.1109/IROS45743.2020.9340979}}

@Article{SurveyLiDAR_ground_segmentation,
AUTHOR = {Gomes, Tiago and Matias, Diogo and Campos, André and Cunha, Luís and Roriz, Ricardo},
TITLE = {A Survey on Ground Segmentation Methods for Automotive LiDAR Sensors},
JOURNAL = {Sensors},
VOLUME = {23},
YEAR = {2023},
NUMBER = {2},
ARTICLE-NUMBER = {601},
PubMedID = {36679414},
ISSN = {1424-8220},
DOI = {10.3390/s23020601}
}

@article{Ransac1981,
author = {Fischler, Martin A. and Bolles, Robert C.},
title = {Random sample consensus: a paradigm for model fitting with applications to image analysis and automated cartography},
year = {1981},
issue_date = {June 1981},
publisher = {Association for Computing Machinery},
address = {New York, NY, USA},
volume = {24},
number = {6},
issn = {0001-0782},
url = {https://doi.org/10.1145/358669.358692},
doi = {10.1145/358669.358692},
journal = {Commun. ACM},
month = jun,
pages = {381–395},
numpages = {15}
}

@article{depth_normal_seg,
  author       = {Yunze Man and
                  Xinshuo Weng and
                  Kris M. Kitani},
  title        = {GroundNet: Segmentation-Aware Monocular Ground Plane Estimation with
                  Geometric Consistency},
  journal      = {CoRR},
  volume       = {abs/1811.07222},
  year         = {2018},
  url          = {http://arxiv.org/abs/1811.07222},
  eprinttype    = {arXiv},
  eprint       = {1811.07222},
  bibsource    = {dblp computer science bibliography, https://dblp.org}
}

@inproceedings{EulerFlowNeurIPS,
title={Neural Eulerian Scene Flow Fields},
author={Kyle Vedder and Neehar Peri and Ishan Khatri and Siyi Li and Eric Eaton and Mehmet Kemal Kocamaz and Yue Wang and Zhiding Yu and Deva Ramanan and Joachim Pehserl},
booktitle={The Thirteenth International Conference on Learning Representations},
year={2025},
url={https://openreview.net/forum?id=0CieWy9ONY}
}

@inproceedings{SeFlowECCV24,
      title={SeFlow: A Self-Supervised Scene Flow Method in Autonomous Driving}, 
      author={Qingwen Zhang and Yi Yang and Peizheng Li and Olov Andersson and Patric Jensfelt},
  booktitle    = {European Conference on Computer Vision},
  year         = {2024},
  organization = {Springer}
}

@inproceedings{
ZeroFlowECCV24,
title={ZeroFlow: Scalable Scene Flow via Distillation},
author={Kyle Vedder and Neehar Peri and Nathaniel Eliot Chodosh and Ishan Khatri and ERIC EATON and Dinesh Jayaraman and Yang Liu and Deva Ramanan and James Hays},
booktitle={The Twelfth International Conference on Learning Representations},
year={2024},
url={https://openreview.net/forum?id=FRCHDhbxZF}
}

@InProceedings{FastNSF_Li_2023_ICCV,
    author    = {Li, Xueqian and Zheng, Jianqiao and Ferroni, Francesco and Pontes, Jhony Kaesemodel and Lucey, Simon},
    title     = {Fast Neural Scene Flow},
    booktitle = {Proceedings of the IEEE/CVF International Conference on Computer Vision (ICCV)},
    month     = {October},
    year      = {2023},
    pages     = {9878-9890}
}

@InProceedings{Floxels_Hoffmann_2025_CVPR,
    author    = {Hoffmann, David T. and Raza, Syed Haseeb and Jiang, Hanqiu and Tananaev, Denis and Klingenhoefer, Steffen and Meinke, Martin},
    title     = {Floxels: Fast Unsupervised Voxel Based Scene Flow Estimation},
    booktitle = {Proceedings of the IEEE/CVF Conference on Computer Vision and Pattern Recognition (CVPR)},
    month     = {June},
    year      = {2025},
    pages     = {22328-22337}
}

@inproceedings{NFSP_Nips21,
 author = {Li, Xueqian and Kaesemodel Pontes, Jhony and Lucey, Simon},
 booktitle = {Advances in Neural Information Processing Systems},
 editor = {M. Ranzato and A. Beygelzimer and Y. Dauphin and P.S. Liang and J. Wortman Vaughan},
 pages = {7838--7851},
 publisher = {Curran Associates, Inc.},
 title = {Neural Scene Flow Prior},
 url = {https://proceedings.neurips.cc/paper_files/paper/2021/file/41263b9a46f6f8f22668476661614478-Paper.pdf},
 volume = {34},
 year = {2021}
}

@ARTICLE{SSL_Flow_4DRadar_RAL22,
  author={Ding, Fangqiang and Pan, Zhijun and Deng, Yimin and Deng, Jianning and Lu, Chris Xiaoxuan},
  journal={IEEE Robotics and Automation Letters}, 
  title={Self-Supervised Scene Flow Estimation With 4-D Automotive Radar}, 
  year={2022},
  volume={7},
  number={3},
  pages={8233-8240},
  doi={10.1109/LRA.2022.3187248}}

@InProceedings{LetOccFlow_CoRL_25,
  title = 	 {Let Occ Flow: Self-Supervised 3D Occupancy Flow Prediction},
  author =       {Liu, Yili and Mou, Linzhan and Yu, Xuan and Han, Chenrui and Mao, Sitong and Xiong, Rong and Wang, Yue},
  booktitle = 	 {Proceedings of The 8th Conference on Robot Learning},
  pages = 	 {2895--2912},
  year = 	 {2025},
  editor = 	 {Agrawal, Pulkit and Kroemer, Oliver and Burgard, Wolfram},
  volume = 	 {270},
  series = 	 {Proceedings of Machine Learning Research},
  month = 	 {06--09 Nov},
  publisher =    {PMLR},
  url = 	 {https://proceedings.mlr.press/v270/liu25e.html}
}

@inproceedings{keetha2026mapanything,
  title={{MapAnything}: Universal Feed-Forward Metric {3D} Reconstruction},
  author={Nikhil Keetha and Norman M\"{u}ller and Johannes Sch\"{o}nberger and Lorenzo Porzi and Yuchen Zhang and Tobias Fischer and Arno Knapitsch and Duncan Zauss and Ethan Weber and Nelson Antunes and Jonathon Luiten and Manuel Lopez-Antequera and Samuel Rota Bul\`{o} and Christian Richardt and Deva Ramanan and Sebastian Scherer and Peter Kontschieder},
  booktitle={International Conference on 3D Vision (3DV)},
  year={2026},
  organization={IEEE}
}

@InProceedings{Leroij24,
author="Palmer, Patrick
and Kr{\"u}ger, Martin
and Sch{\"u}tte, Stefan
and Altendorfer, Richard
and Adam, Ganesh
and Bertram, Torsten",
editor="Leonardis, Ale{\v{s}}
and Ricci, Elisa
and Roth, Stefan
and Russakovsky, Olga
and Sattler, Torsten
and Varol, G{\"u}l",
title="LEROjD: Lidar Extended Radar-Only Object Detection",
booktitle="Computer Vision -- ECCV 2024",
year="2025",
publisher="Springer Nature Switzerland",
address="Cham",
pages="379--396",
isbn="978-3-031-73027-6"
}

\clearpage
\newpage

\clearpage
\setcounter{section}{0}
\renewcommand{\thesection}{\Alph{section}}

\begin{center}
\Large\bfseries Supplementary Materials \\[0.5em]
\large TriO: Tri-Modal Unsupervised Occupancy World Model for Anything Perception
\end{center}
\vspace{1em}

This supplementary material provides additional related work, implementation details, the experimental setup, and further results.

{
\setcounter{tocdepth}{2}
\small
\section*{Contents}
\makeatletter
\providecommand{\authcount}[1]{}
\providecommand{\orcidlinkX}[3]{}
\let\orig@contentsline\contentsline
\renewcommand{\contentsline}[4]{%
  \def\tmp@type{#1}%
  \def\tmp@title{title}%
  \def\tmp@author{author}%
  \ifx\tmp@type\tmp@title\else
    \ifx\tmp@type\tmp@author\else
      \orig@contentsline{#1}{#2}{#3}{#4}%
    \fi
  \fi
}
\@starttoc{toc}
\let\contentsline\orig@contentsline
\makeatother
}

\section{Extended Related Works}

\begin{table}[t]
\centering
\caption{Comparison of \ourmodel{} against the most closely related methods. 
* Only DIO-$\varnothing$ does not use human labels. 
$^\dagger$QueryOcc+ uses both camera and LiDAR supervision. 
$^\ddagger$The used segmentation models rely annotations like per pixel semantic masks or image-text pairs annotations. $^\diamond$ Continuous Gaussian coordinates discretized via voxelization}
\label{tab:method_comparison_trio}
\resizebox{\textwidth}{!}{%
\begin{tabular}{@{}lcccccccccc@{}}
\toprule
Method & \multicolumn{3}{c}{Input} & \multicolumn{3}{c}{Supervision} & Semantics & Continuous & 4D & Unsupervised \\
\cmidrule(r){2-4} \cmidrule(r){5-7}
 & \tiny Camera & \tiny LiDAR & \tiny RADAR & \tiny Camera & \tiny LiDAR & \tiny RADAR & & &  & \\
\midrule
ImplicitO \cite{agro2023implicit} & \xmark & \checkmark & \xmark & \xmark & \xmark  & \xmark & \checkmark & \checkmark & \xmark & \xmark \\
Occ4D \cite{khurana2023point} &  \xmark & \checkmark & \xmark & \xmark & \checkmark  & \xmark & \xmark & \checkmark & \checkmark & \checkmark \\
UnO \cite{Agro_2024_CVPR}     & \xmark & \checkmark & \xmark & \xmark & \checkmark  & \xmark & \xmark & \checkmark & \checkmark & \checkmark \\
DIO \cite{Diehl_2025_DIO}     & \xmark & \checkmark & \xmark & \xmark & \checkmark  & \xmark & \checkmark & \checkmark & \checkmark & (\xmark)$^*$ \\
GASP \cite{GASP_2025}          &  \xmark & \checkmark  & \xmark  & \checkmark  & \checkmark  & \xmark  & \xmark & \checkmark  & \checkmark & \xmark \\ 
MonoScene \cite{MonoScene} &  \checkmark & \xmark  & \xmark  & \xmark  & \xmark  & \xmark  & \checkmark & \xmark  & \xmark & \xmark \\ 
GaussianWorld  \cite{Zuo_2025_CVPR}                & \checkmark  & \xmark & \xmark & \xmark & \xmark & \xmark & \checkmark & (\checkmark)$^\diamond$ & \checkmark & \xmark \\
SceneDINO \cite{Jevtic_2025_ICCV} & \checkmark  & \xmark & \xmark  & \checkmark & \xmark & \xmark & \checkmark  & \checkmark  & \xmark  & \checkmark \\
OccNerf \cite{OccNerf23} & \checkmark & \xmark  & \xmark  & \checkmark  & \xmark  & \xmark  & \checkmark  & \checkmark  & \xmark  & (\checkmark)$^\ddagger$ \\
GaussianFlowOcc \cite{Boeder_2025_ICCV} & \checkmark & \xmark  & \xmark  & \checkmark  & \xmark  & \xmark  & \checkmark  & \xmark  & \xmark  & (\checkmark)$^\ddagger$ \\
QueryOcc \cite{QueryOcc_2025} & \checkmark & \xmark  & \xmark  & \checkmark  & $(\checkmark)^\dagger$  & \xmark  & \checkmark  & \checkmark  & \xmark  & (\checkmark)$^\ddagger$ \\
LRS4Fusion \cite{Palladin_2025_ICCV} &  \checkmark & \checkmark & \xmark & \xmark & \checkmark & \xmark  & \xmark & \checkmark & \checkmark & \checkmark  \\
RadarOcc \cite{RadarOcc2024} & \xmark  & \xmark & \checkmark  & \xmark & \xmark & \xmark & \checkmark  & \xmark  & \xmark  & \xmark \\
4DRolls \cite{4DRolls_IROS2025} & \xmark  & \xmark & \checkmark  & \xmark & \checkmark & \xmark & \xmark  & \xmark  & \checkmark  & \checkmark \\
RadarSplat \cite{Kung_2025_ICCV} & \xmark  & \xmark & \checkmark  & \xmark & \xmark & \checkmark & \xmark  & \xmark  & \xmark  & \checkmark \\
OccFusion \cite{OccFusion25} & \checkmark  & \checkmark & \checkmark  & \xmark & \xmark & \xmark & \checkmark  & \xmark  & \xmark  & \xmark \\
GaussianFusionOcc \cite{gaussianfusionoccseamlesssensorfusion} & \checkmark  & \checkmark & \checkmark  & \xmark & \xmark & \xmark & \checkmark  & (\checkmark)$^\diamond$  & \xmark  & \xmark \\
ShelfGaussian \cite{zhao2025shelfgaussian} & \checkmark  & \checkmark & \checkmark  & \checkmark & \checkmark & \xmark & \checkmark  & (\checkmark)$^\diamond$  & \xmark  & (\checkmark)$^\ddagger$ \\
\midrule
\textbf{TriO (Ours)} & \checkmark & \checkmark & \checkmark & \checkmark & \checkmark & \checkmark & \checkmark & \checkmark & \checkmark & \checkmark \\
\bottomrule
\end{tabular}%
}
\end{table}
This section provides an additional comparison to related work. We begin by revisiting occupancy-related approaches described in the main paper. \cref{tab:method_comparison_trio} compares the most closely related methods in terms of input modalities, supervision signals, and model properties. \ourmodel{} is the only method that uses all three modalities for both input and self-supervision, while remaining unsupervised and modeling continuous occupancy in 4D.

\subsubsection{Self-Supervised Flow Predictions.} 
\ourmodel{} is also related to self-supervised scene flow methods, which can be grouped into two categories. The first solves global optimization problems using test-time optimization techniques to estimate point flow across multiple frames \cite{EulerFlowNeurIPS, FastNSF_Li_2023_ICCV}. One downside of this approach is its high runtime, which often makes these methods unsuitable for real-time applications. The second category uses feed-forward neural networks to predict point flow across frames. This is achieved through self-supervised losses \cite{NFSP_Nips21, SeFlowECCV24, liu2024selfsupervisedmultiframeneuralscene, Flow4DRAL2025, Floxels_Hoffmann_2025_CVPR} or distillation from test-time optimization techniques \cite{ZeroFlowECCV24}. The previously mentioned methods primarily rely on multiple frames of LiDAR point cloud data. However, RADAR data can provide a partial supervision signal for velocity estimation via the Doppler effect. For example, \cite{SSL_Flow_4DRadar_RAL22} uses RADAR data, including radial velocity measurements from 4D RADARs.
Other methods also combine occupancy and flow estimation: \cite{Diehl_2025_DIO} proposes a occupancy consistency loss between 3D occupancy predictions to infer flow while relying solely on LiDAR. \cite{LetOccFlow_CoRL_25} estimates 3D occupancy and occupancy flow using only camera inputs, with self-supervision via differentiable rendering.

In contrast, our method is not only able to predict 4D occupancy and obstacle segmentations, but also uniquely combines multiple sensor modalities as input and self-supervised flow learning with RADAR velocity, while additionally predicting future 4D flow.

\subsubsection{Pre-Training for Perception.} 
Representation learning methods aim to leverage large-scale unlabeled data to enhance performance on downstream tasks. Beyond the occupancy-based approaches \cite{khurana2023point, Agro_2024_CVPR, GASP_2025, Diehl_2025_DIO, Palladin_2025_ICCV, QueryOcc_2025},  which utilize LiDAR or camera self-supervision, several general-purpose representation learning frameworks have emerged. For instance, VoxelMAE \cite{hess2023masked} employs masked auto-encoding on voxelized point clouds as an effective pre-training strategy for 3D object detection. UniPAD \cite{yang2023unipad} integrates both camera and LiDAR inputs within a 3D differentiable rendering framework to reconstruct color and depth. More recently, AD-L-JEPA \cite{adjepa2025} introduced a Joint Embedding Predictive Architecture (JEPA) for LiDAR, demonstrating superior downstream detection performance compared to traditional generative masked auto-encoders. In the RADAR domain, RaDICaL \cite{RadarSSL_CVPR24} proposed RADAR-to-RADAR and RADAR-to-VISION contrastive losses as a self-supervised pre-training objective.
LEROjD \cite{Leroij24} introduces a LIDAR-to-RADAR cross-modal distillation and LiDAR thin-out strategies for downstream RADAR-only object detection. 

In contrast to these single- or dual-modality approaches, \ourmodel{} utilizes pre-training across all three modalities simultaneously. Furthermore, while existing methods focus primarily on representations for instantaneous downstream tasks (e.g. detection), \ourmodel{} specifically targets representations optimized for temporal 4D forecasting.

\section{Implementation Details}
\subsection{Network Architecture}

In this subsection we revisit the architecture of \ourmodel{} and provide additional details on the implementation. 

\subsubsection{Encoder:}
We instantiate the camera encoder $\fcam$ as a ResNet50 \cite{he2016deep} operating on $N_C$ input images $\mathbf{C}$ of size $1920 \times 1080$. The LiDAR-RADAR branch voxelizes the multi-frame point measurements and encodes them into a sparse 3D feature grid.
The voxel ROI spans $[-50, 350]$ m along the longitudinal axis, $[-50, 50]$ m along the lateral axis, both with a resolution of $0.15625$m, and $[-2, 4]$ m in the vertical dimension, with a height resolution of $0.2$m. 
Beyond spatial coordinates relative to the center of their corresponding voxel cell, $(x, y, z)_r \in \mathbb{R}^3$, the features from the LiDAR points 
$\mathbf{l} = [x, y, z, i, \Delta t]^\intercal \in \mathbb{R}^5$ (including intensity $i$ and relative 
timestep $\Delta t$) and RADAR points 
$\mathbf{r} = [x, y, z, v_r, \sigma, \Delta t]^\intercal \in \mathbb{R}^6$ (including range rate $v_r$, RCS $\sigma$, and $\Delta t$) are passed 
through separate two-layer MLPs with group norm to encode the input features of dimension 5 or 6 outputting features of size 15. The LiDAR intensity is rasterized in BEV and concatenated such that the final output has size 16. We aggregate points in the same voxel with mean aggregation. The result is a sparse feature grid $\mathbf{Z}_{\text{voxel}} \in \mathbb{R}^{30\times 640 \times 2560}$. This tensor is passed through a series of sparse convolutions (LiDAR-RADAR backbone) \cite{SubmanifoldSparseConvNet} progressively halving the spatial resolution and doubling the channel dimension (16$\to$32$\to$64$\to$128 channels), yielding $\zlr \in \mathbb{R}^{320 \times 80 \times 4 \times 128}$ at $8{\times}$ spatial downsampling, which is ultimately fused with the image features $\zc$.

\subsubsection{Feature Fusion.} In order to fuse the features from the RADAR-LiDAR encoder and the image encoder $\zc, \zlr$, we employ an image to sparse voxel attention mechanism, inspired by \cite{Li_2023_CVPR} giving $\mathcal{Z}_0=[\mathbf{V}_{2x}, \mathbf{V}_{4x}, \mathbf{V}_{8x}, \mathbf{M}_{16x}]$, with feature volumes $\mathbf{V}_{2x}, \mathbf{V}_{4x}, \mathbf{V}_{8x}$ and BEV feature map $\mathbf{M}_{16x}$ having resolution downsampled by 2x, 4x, 8x, and 16x, respectively:
\begin{equation}
    \mathcal{Z}_0 = f_\text{voxelatt}(\zc, \zlr) 
\end{equation}
However, in our implementation, we allow each of the sparse features to attend using deformable attention back into the image features, using their projected location in image space as the center of the deformable attention mechanism. We also encode the position features in both the LiDAR and image space before running through the attention mechanism. 
$\zc \in \mathbb{R}^{N_\text{cam} \times 128 \times H_c \times W_c}$ denotes the camera feature maps from the ResNet50 image backbone, and $\mathbf{M}_{16x} \in \mathbb{R}^{256 \times 80 \times 320}$ is the resulting fused dense BEV feature map. We then get $\mathcal{Z}_0$ by using $\mathbf{M}_{16x}$ and the intermediate feature maps of the previously mentioned LiDAR-RADAR backbone.

\subsubsection{Deformable Attention Neck.}
The fused BEV feature map $\mathcal{Z}_0$ is passed through a multi-scale neck
$f_\text{deformatt}$ composed of a ResNet with deformable convolution blocks~\cite{zhu2018deformableconvnetsv2deformable}
followed by a feature pyramid network (FPN)~\cite{lin2017feature}, yielding:
\begin{equation}
    \mathcal{Z}_1 = f_\text{deformatt}(\mathcal{Z}_0).
\end{equation}
Each deformable convolution block learns offsets,
enabling the network to adaptively expand its receptive field beyond the
fixed local neighborhood of standard convolutions.
The FPN aggregates these features producing a rich BEV representation at the original spatial resolution of $\mathbf{M}_{16x}$.
This is reminiscent of what is done in DIO~\cite{Diehl_2025_DIO}, but with
the additional deformable convolution operations enabling larger effective
receptive fields, resulting in better forecasting of fast-moving entities.

\subsubsection{Decoder.} For the decoder $f_{\text{dec}}$, we use the multiscale deformable attention mechanism of \cite{Diehl_2025_DIO}, 
\begin{equation}
    (o, \mathbf{f}, s) = f_\text{dec}(\mathcal{Z}_1, \zlr, \mathbf{q}).
\end{equation}
Specifically, given a query point $\mathbf{q}$, the header attends to each level of the sparse feature maps ($\mathbf{V}_{2x}, \mathbf{V}_{4x}, \mathbf{V}_{8x}$) and the dense feature map in $\mathcal{Z}_1$ at the location of this query point in 3D space. We then perform deformable attention to attend to a series of different locations, querying each of them across all feature maps, and combining them using the same deformable attention mechanism. The final features are then passed through a shallow MLP which decodes the occupancy probability $o$, the instantaneous motion (flow) $\mathbf{f}$ of the point, and the probability $s$ of that point being classified as an obstacle. This operation can be done efficiently in parallel for a batch of query points.

\subsection{Obstacle Segmentation Model}
This section provides implementation details of the obstacle segmentation pipeline used to supervise $s$, including the per-modality pseudo-label construction and fusion rule.

\subsubsection{Image Obstacle Segmentation.}
\label{sec:method_image_seg_model}

For each camera image $\mathbf{I} \in \mathbb{R}^{H \times W \times 3}$ with known intrinsics $K \in \mathbb{R}^{3\times3}$ and extrinsic pose $T_{\text{cam} \to \text{veh}}$, we use Metric3D v2 \cite{hu2024metric3dv2} to obtain per-pixel metric depth $d \in \mathbb{R}^{H \times W}$, depth confidence $c \in \mathbb{R}^{H \times W}$, and surface normals $\hat{\mathbf{n}} \in \mathbb{R}^{H \times W \times 3}$ with associated confidence $c_n \in \mathbb{R}^{H \times W}$.

We then convert each pixel to a 3D point and transform it, together with the normals, into vehicle frame. Let $\mathbf{p}$ and $\mathbf{n}$ denote a resulting 3D point and surface normal vector, respectively. On a high-level we assume that obstacle regions are defined by points with upwards (in $z$) pointing normals and smooth normal gradients. 

\textit{Soft Probabilistic Obstacle Criteria.}
Rather than applying hard thresholds we convert each geometric criterion into a soft probability using a parameterized sigmoid:

\begin{equation}
    \sigma_{\tau,k}(x) = 1 - \frac{1 + e^{-k\tau}}{1 + e^{k(x - \tau)}},
\end{equation}
where $\tau$ is the transition point and $k$ controls the steepness. This function transitions smoothly from $0$ at $x{=}0$ to ${\approx}1$ for $x \gg \tau$. For criteria where high values indicate invalidity, we use the complement $\bar{\sigma}_{\tau,k}(x) = 1 - \sigma_{\tau,k}(x)$. We compute four independent per-pixel probabilities:
\begin{enumerate}
    \item \textbf{Normal orientation}: $p_{\text{norm}} = \sigma_{\tau_n, k_n}(n_z)$, assigning high probability to upward-facing surfaces.
    \item \textbf{Normal smoothness}: $p_{\text{grad}} = \bar{\sigma}_{\tau_g, k_g}\!\bigl(\lVert \nabla\, \mathbf{n} \rVert \bigr)$, where  $\lVert \nabla\, \mathbf{n} \rVert$ is the normalized spatial image gradient of the normal field. We assume ground regions exhibit spatially smooth normals, while object boundaries produce large normalized gradients. 
    \item \textbf{Depth confidence}: $p_{\text{conf}} = \sigma_{\tau_c, k_c}(c)$, assigning low probability for points where depth estimates are unconfident.
    \item \textbf{Range}: $p_{\text{dist}} = \bar{\sigma}_{\tau_d, k_d}(p_x)$, assigning low probability for distant points where monocular depth degrades.
\end{enumerate}

\textit{RANSAC Plane Refinement.}
To further disambiguate ground from elevated flat surfaces (\eg, vehicle rooftops), we fit a ground plane to candidate 3D points using RANSAC \cite{Ransac1981}. We select candidate ground points as those whose combined probability exceeds a threshold $\tau_\pi$:
\begin{equation}
    p_{\text{conf}} \cdot p_{\text{dist}} \cdot p_{\text{norm}} \cdot p_{\text{grad}} > \tau_\pi.
\end{equation}
After applying RANSAC we compute the \textit{point-to-plane distance} $d_{\text{plane}}$, and the soft probability:
\begin{equation}
    p_{\text{plane}} = \bar{\sigma}_{\tau_r, k_r}\left(d_{\text{plane}}\right).
\end{equation}

\textit{Final Obstacle and Validity Probabilities.}
We define the ground probability as the product of the normal-based criteria, and the obstacle probability as the inverse
\begin{align}
    p_{\text{ground}} = p_{\text{norm}} \cdot p_{\text{grad}} \label{eq:p_ground} \\
    p_{\text{obst}} = 1 - p_{\text{ground}} \label{eq:p_obst} 
\end{align}
and the validity probability, which captures whether a pixel carries reliable geometric information or belong to regions like the sky:
\begin{equation}
    p_{\text{valid}} = \sigma(c_n) \cdot p_{\text{conf}} \cdot p_{\text{dist}} \cdot \bigl(1 - (1 - p_{\text{plane}}) \cdot p_{\text{ground}}\bigr),
    \label{eq:validity}
\end{equation}
where $\sigma(c_n)$ is the standard sigmoid of the normal confidence. 

The resulting per-pixel distribution over three classes is:
\begin{align}
    p^{\text{image}}_{\text{invalid}} &= 1 - p_{\text{valid}}, \label{eq:seg_invalid} \\
    p^{\text{image}}_{\text{obst}} &= p_{\text{valid}} \cdot (1 - p_{\text{ground}}), \label{eq:seg_nonground} \\
    p^{\text{image}}_{\text{ground}} &= p_{\text{valid}} \cdot p_{\text{ground}}. \label{eq:seg_ground}
\end{align}

These soft image-space labels (\ourimagelabels) are projected onto 3D LiDAR points via camera--LiDAR calibration to produce dense supervisory signals for the model output $s$. The output is a set of LiDAR points with confidences and their probability of being associated with the ground class, defined by $p^{\text{image}}_{\text{valid}}$ and $p^{\text{image}}_{\text{ground}}$, respectively.

\subsubsection{LiDAR Obstacle Segmentation.} 
We take inspiration from Patchwork++ \cite{patchworkpp}; however, rather than using its final obstacle segmentation output, we utilize its core ideas for local ground patch estimation. We first discretize the point cloud into polar patches, where for each patch $j$, a ground plane is characterized by its centroid $\mathbf{c}_j \in \mathbb{R}^3$ and surface normal $\hat{\mathbf{n}}_j \in \mathbb{R}^3$. For a LiDAR return $\mathbf{l} \in \mathbb{R}^3$, we identify its corresponding patch in the $xy$-plane. We then get the elevation $e\in \mathbb{R}$ as the point to plane distance using the patch normal, which we finally convert into a probability using a parametrized sigmoid 
\begin{equation}
p^{\text{LiDAR}}_{\text{obst}} = \sigma_{\tau_e, k_e}(e_i),    
\end{equation}

To define the validity probability, we quantify the uncertainty of the obstacle classification using a scaled binary entropy $\mathcal{H}$. This ensures that points with ambiguous geometric features contribute less to the supervisory signal. The LiDAR validity is \begin{equation}
p^{\text{LiDAR}}_{\text{valid}} = -\lambda \mathcal{H}(p^{\text{LiDAR}}_{\text{obst}}) + \gamma,
\end{equation} where $\lambda$ and $\gamma$ scale this to range between 0 and 1, ensuring that a probability of 0.5 results in a low validity score. This provides us with the LiDAR segmentation probabilities.

\subsubsection{Fused Obstacle Segmentation.}
We use a confidence-based fusion rule to select the supervision source per LiDAR point from the image or LiDAR based segmentation.
Specifically, if the LiDAR segmentation validity is low and the image validity is higher, we use image labels; otherwise, we keep LiDAR labels:
\begin{equation}
\begin{aligned}
    \mathbf{p}^{\text{comb}} &= 
    \begin{cases} 
        \mathbf{p}^{\text{image}}, & \text{if } p^{\text{LiDAR}}_{\text{valid}} < 0.5 \;\wedge\; p^{\text{image}}_{\text{valid}} > p^{\text{LiDAR}}_{\text{valid}} \\
        \mathbf{p}^{\text{LiDAR}}, & \text{otherwise}
    \end{cases}
\end{aligned}
\end{equation}
with $\mathbf{p}^{\textsc{source}} = \bigl[p^{\textsc{source}}_{\text{valid}},\; p^{\textsc{source}}_{\text{obst}}\bigr]^\intercal$, $\textsc{source} \in \{\text{comb}, \text{image},\, \text{LiDAR}\}$.  The probability vector $\mathbf{p}^{\text{comb}}$ defines the final combined soft labels (our combined labels), which we use for supervision of obstacle semantics. \Cref{tab:segmentation_hyperparameters} summarizes the hyperparameters used in the obstacle segmentation pipeline. 
\begin{table}[t]
    \centering
    \footnotesize
    \setlength\tabcolsep{4pt}
    \caption{Obstacle segmentation hyperparameters.}
    \label{tab:segmentation_hyperparameters}
    \begin{tabular}{@{}l l c@{}}
    \toprule
    Symbol & Description & Value \\
    \midrule
    $\tau_n$ & Normal orientation transition threshold & $0.85$ \\
    $k_n$ & Normal orientation sigmoid steepness & $20.0$ \\
    $\tau_g$ & Normal smoothness transition threshold & $0.05$ \\
    $k_g$ & Normal smoothness sigmoid steepness & $70.0$ \\
    $\tau_c$ & Depth confidence transition threshold & $0.7$ \\
    $k_c$ & Depth confidence sigmoid steepness & $100.0$ \\
    $\tau_d$ & Distance transition threshold & $150.0$ \\
    $k_d$ & Distance sigmoid steepness & $0.05$ \\
    $\tau_\pi$ & Candidate ground probability threshold & $0.5$ \\
    $\tau_r$ & Plane-distance transition threshold & $0.5$ \\
    $k_r$ & Plane-distance sigmoid steepness & $2.0$ \\
    $\tau_e$ & LiDAR elevation transition threshold & $0.2$ \\
    $k_e$ & LiDAR elevation sigmoid steepness & $20.0$ \\
    $\lambda$ & Entropy scaling factor for $p^{\text{LiDAR}}_{\text{valid}}$ & $3$ \\
    $\gamma$ & Entropy offset for $p^{\text{LiDAR}}_{\text{valid}}$ & $1$ \\
    \bottomrule
    \end{tabular}
\end{table}

\subsection{Occupancy World Model Loss Functions}
We provide additional loss function details for the occupancy ($\occloss$) and obstacle segmentation ($\groundloss$) loss: 
The occupancy loss is a binary cross-entropy loss summed across all query points:
\begin{align}
    \occloss = - \frac{1}{|\mathcal{Q^+} + \mathcal{Q^-}|}\left(\sum_{\mathbf{q} \in \mathcal{Q}^-} \log(1 - f_{\theta, \text{occ}}(\mathbf{q})) + \sum_{\mathbf{q} \in \mathcal{Q}^+} \log(f_{\theta, \text{occ}}(\mathbf{q})) \right).
    \label{eq:occ_loss_appendix}
\end{align}
The groundness probability loss is similarly defined as a \textit{soft} focal loss function, weighted by the validity probability: 
\begin{equation}
\begin{split}
\groundloss
= -\frac{1}{|\mathcal{P}_{\text{sweep}}|}
&\sum_{\mathbf{q} \in \mathcal{P}_{\text{sweep}}}
p^{\text{comb}}_{\text{valid}}(\mathbf{q})
\Big(
\alpha \, p^{\text{comb}}_{\text{ground}}(\mathbf{q})
\big(1-f_{\theta,\text{seg}}(\mathbf{q})\big)^{\gamma}
\log\!\big(f_{\theta,\text{seg}}(\mathbf{q})\big) \\
&+ (1-\alpha)\big(1-p^{\text{comb}}_{\text{ground}}(\mathbf{q})\big)
\big(f_{\theta,\text{seg}}(\mathbf{q})\big)^{\gamma}
\log\!\big(1-f_{\theta,\text{seg}}(\mathbf{q})\big)
\Big).
\end{split}
\end{equation}
with $\alpha = 0.25$ and $\gamma=2.0$.

\subsection{Training of Downstream Task}
\subsubsection{LiDAR Forecasting.}
For LiDAR forecasting, we apply techniques from prior work \cite{khurana2022differentiable, Agro_2024_CVPR, Diehl_2025_DIO}. We first train the occupancy world model $f_{\theta}$, followed by a training of a LiDAR forecasting header $f_{\psi}$ with frozen occupancy parameters $\theta$ and new parameters $\psi$. This model is specifically trained to predict the expected depth of a LiDAR for the task of lidar forecasting.  We use an $L_1$ loss to supervise the depth of the model under the assumption of previous works that the direction of the LiDAR ray can be assumed to be known, and that our goal is to render where the ray terminates. Therefore, given a set of LiDAR rays $\mathcal{P}_\text{ray}$ with LiDAR ray returns $\mathbf{l}$ we wish to sample, we formulate the loss as:

\begin{equation}
    \lidarloss = \frac{1}{|\mathcal{P}_{\text{ray}}|} \sum_{\mathbf{l} \in \mathcal{P}_\text{ray}} \left| f_{\psi}(f_{\theta, \text{occ}}(\mathbf{l})) - \lVert \mathbf{l} \rVert_2 \right|.
\end{equation}

In practice, during training we randomly sample 50 LiDAR rays, then further sample 2000 points per ray at 0.1m resolution to calculate our LiDAR forecasting loss.

\subsubsection{BEV/ 3D Supervised Semantic Occupancy.}

For supervised occupancy training, we simply use binary cross entropy to train the model using the rasterized box labels of the objects of interest. For the sake of evaluating how well the model forecasts the occupancy of fast moving objects, we focus on supervising exclusively on the vehicle class. Therefore, for a set of points $\mathcal{U}$ uniformly sampled in 2D BEV (LRR) or 3D (AV2) over the region of interest, and a label function $\text{Ras}(\mathbf{q})$ that returns 1 if the query point lies within a bounding box and 0 if not, with the loss function defined in \Cref{eq:occ_loss_appendix}.

\subsection{Training Hyperparameters} 
We train \ourmodel{} on 16 GPUs using the AdamW optimizer with a cosine learning rate schedule. The learning rate begins at $8.0 \times 10^{-5}$ and follows a 1,000-iteration warmup period to reach a peak of $8.0 \times 10^{-4}$. We set $N = |\mathcal{Q}^+| = |\mathcal{Q}^-| = 7500$, $|\mathcal{P}_{\text{sweep}}| = 10000$, and $|\mathcal{R}_{\text{clean}}| = 2000$, with loss weights configured as $\lambda_1 = 1500, \lambda_2 = 10$, and $\lambda_3 = 125$. Note for a fair comparison, hyperparameters of baselines are also tuned.

\subsection{Extended Discussion of Unsupervised Framing}
Throughout this paper, we use \textit{unsupervised} to denote that \ourmodel{}'s training pipeline requires no human-provided annotations (e.g., bounding boxes, semantic labels, or text prompts). The pre-trained model used for pseudo-label generation (Metric3Dv2~\cite{hu2024metric3dv2}) was supervised on geometric properties (depth and normals), which can be derived from LiDAR and reconstructions. Nevertheless, some datasets (e.g.,~\cite{ScanNet2017}) used by~\cite{hu2024metric3dv2} (see Tab. 5 in the reference)
contain further human-in-the-loop work like CAD model alignment; hence, we
restrict our unsupervised claim to \ourmodel{}'s \textit{own training pipeline and not the entire system}. 

\section{Experimental Setup}

In this section we provide additional details for our experimental setup  on Argoverse 2 (AV2) \cite{wilson2023argoverse}, Spotting the Unexpected (STU) \cite{STU_cvpr2025} and the unreleased dataset Long Range Radar (LRR).

\subsection{Occupancy World Model Pre-training (AV2/LRR)}
\label{sec:supp_exp_setupocc_world_model_pretraining}

AV2 is a dataset collected in six cities in the United States, especially mined for interesting urban scenarios. We use the official split with 750 train and 150 val sequences. Each sequence spans \SI{15}{\second} with 150 frames. We train \ourmodel{} on a ROI of $[-100, 150]$m in the $x$ dimension  $[-100, 100]$ in the y dimension and $[-4.5, 4.5]$m in $z$ dimension.
For generating the segmentation labels, we use all eight cameras of the camera rig (\cite{wilson2023argoverse}). 
We train our model with the front left and front right cameras ($N_{\text{C}}=2)$ for memory efficiency and use one LiDAR ($N_{\text{L}}=1$). We use $H=\SI{3.0}{\second}$ of past data sampled at 5Hz and pre-train to predict $T=\SI{3.0}{\second}$ into the future. We also ablated using four cameras (\ref{sec:cam_ablations}), which led to slight performance gains.

LRR is a dataset with urban and highway driving with 2280 sequences each 20 seconds long, whereas we use 19 sequences for evaluation. 
\ourmodel{} uses $N_{\text{C}} = 5$ cameras, $N_{\text{L}}=6$ LiDAR and $N_{\text{R}}=4$ RADAR. We use $H=\SI{0.5}{\second}$ of past data sampled at 10 Hz and pre-train with to predict $T=\SI{1.5}{\second}$ into the future. We use an ROI of $[-50, 350]$ m on the $x$ dimension  $[-50, 50]$ on the y dimension and $[-2, 4]$ m in $z$ dimension.

\subsection{Zero-Shot Anything Perception Evaluation (STU)}
We evaluate \textit{zero-shot} obstacle segmentation on long-tail obstacles using the Spotting the Unexpected (STU) dataset, where we use one LiDAR ($N_{\text{L}}=1$) and one front camera ($N_{\text{C}}=1$) featuring sensor characteristics that differ from those on AV2. The STU dataset is split into three parts: a train set, a validation set, and a non-public test set.  The train set consists of two sequences and represents naturalistic driving data with fully labeled scenes. The validation set contains 19 sequences and features a variety of unexpected obstacles placed in the road, but does not contain labels for the ground.
We combine the train and validation sets into a single dataset for our evaluations in the following manner: first we retain only points with negative (ground) labels from the train set, second we retain only points with positive (obstacle) labels from the validation set. This combined set is only used for evaluation and not for training or fine-tuning the models trained on AV2. The combination allows for an evaluation focused on how well rare obstacles can be segmented from the road.

Next we detail pre-processing steps used in our evaluation process. First, labeled LiDAR points are filtered to the height range $z \in [-3, 5]$\,m and to those projecting into the front camera's field of view via pinhole projection with the provided intrinsics, extrinsics, and distortion parameters. 
For the implicit world model evaluations (TriO), each LiDAR point $\mathbf{l} = (l_x,l_y,l_z)$ is augmented with a zero timestamp to form a query point $\mathbf{q}=(l_x,l_y,l_z,l_t{=}0)$. Then, the decoder is queried with these LiDAR points to obtain predictions for occupancy and groundness. Points with predicted occupancy of less than 0.5 are assigned to ground.
For image-based baselines, obstacle segmentation prediction is obtained by projecting LiDAR points onto the segmentation image associated with the front camera. For image methods with binarized outputs, ignore/invalid classifications are mapped to ground. For image methods with probabilistic outputs, the valid probability channel is ignored when determining the pixel label (leaving only $p^{\text{image}}_{\text{obst}}$ and 
    $p^{\text{image}}_{\text{ground}}$ \label{eq:seg_ground}).  Thus, we produce a set of predictions and their associated labels that can be used for evaluation of both implicit models and image space models.

\subsubsection{Metrics.}
Given these predictions and their associated labels, a precision--recall curve is computed over the combined dataset. The operating-point threshold is selected to maximize the $F_1$ score. All metrics requiring an operating point are reported using this optimal threshold. Models that output already binarized have their binarized output used directly.
We report four metrics: (i)~mean Intersection over Union (mIoU) (ii)~$F_1$ score; (iii)~recall; and (iv)~Average Precision (AP).
\begin{itemize}
    \item \textbf{mIoU} [$\%$]: This metric is the mean Intersection over Union, and represents the number of points that are correctly predicted to be the desired class, divided by the union of the prediction and the labels for said class. This is averaged over the two classes equally and is generally a standard metric for semantic occupancy \cite{OpenOccupancy_2023_ICCV}. 

    \item \textbf{F1} [$\%$]: This formally represents the harmonic mean over precision and recall. It represents how well the model can balance both of these metrics at a single optimal operating point. This metric is more suitable for evaluating the segmentation pseudo-labels than other metrics such as mean average precision, due to the fact that they evaluate at a single operating point, and therefore do not unfairly advantage soft metrics that return the probability of each class. 

    \item \textbf{Recall} [$\%$]: Recall tracks the proportion of ground-truth obstacle points that are correctly classified as obstacle points. This is important for evaluating the safety of autonomous driving systems, as missing even a single obstacle could lead to a critical failure of the overall system.

    \item \textbf{AP}[$\%$]: Average Precision represents the area under the precision recall curve, and is often used to determine how well a model is able to interplay between true positives, false positives, and false negatives. However, the precision recall curve is computed over a series of different thresholds, and therefore this metric is designed with continuous probability models in mind. As a result, we only use this metric on our occupancy baselines, where a direct comparison is fair.
\end{itemize}

\begin{table}[t]
\centering
\caption{\textbf{Default} terms/ text prompts used for open-vocabulary label mapping. Those colored as \textcolor{orange}{orange} correspond to an obstacle class, and those colored as \textcolor{blue}{blue} correspond to a ground class. Table design inspired from \cite{sal2024eccv}.}
\label{tab:text_prompts_sal}
\begin{tabular}{l|l}
\toprule
Class & Text prompts \\
\midrule
\multicolumn{2}{c}{Super classes} \\
\midrule
\multirow{2}{*}{\textcolor{orange}{vehicle}}   & vehicle, car, truck, bicycle, motorcycle, other-vehicle, jeep, SUV, van, \\
                            & bike, moped, pickup truck, caravan, trailer, bus, tram, train, \\
                            & construction vehicle, crane, excavator \\
\multirow{2}{*}{\textcolor{orange}{human}}     & human, person, bicyclist, motorcyclist, pedestrian, bicycle rider, \\
                            & motorcycle rider \\
\multirow{2}{*}{\textcolor{blue}{ground}}    & ground, road, sidewalk, parking, other-ground, driveable area, \\
                            & service lane, bike lane, parking lot, curb, driveway, traffic island \\
\textcolor{orange}{structure}                   & structure, building, garage, wall, window, stair \\
\multirow{2}{*}{\textcolor{orange}{nature}}    & nature, vegetation, trunk, terrain, bush, shrub, foliage, treetop, \\
                            & tree trunk, grass, soil \\
\multirow{2}{*}{\textcolor{orange}{object}}    & object, fence, pole, traffic-sign, lamp post, traffic-sign pole, \\
                            & traffic-sign mounting, separator, small wall, crash barrier, \\
                            & traffic cone, bench \\
\bottomrule
\end{tabular}%
\end{table}
\begin{table}[t]
\centering
\small
\setlength{\tabcolsep}{4pt}
\caption{\textbf{Oracle} \textcolor{orange}{obstacle} text prompts/ terms hand picked for STU, separated by category. These prompts are added to $\mathcal{P}_{\text{obstacle}}$ got the oracle baselines.}
\begin{tabular}{@{}ll@{}}
\toprule
Category & Oracle term/ text prompt \\
\midrule
Seating & Chair, Office chair, Roller chair, Gaming chair \\
Bags & Dior bag, Shopping bag, Paper bag, Plastic bag, Polkadot bag, \\
      & Banana bag, Zebra bag, Inflatable bag, Briefcase, Sleeping bag, \\
      & Workout bag, Backpack \\
Inflatables & Inflatable toy, Inflatable ball, Inflatable, Pool noodle \\
Balls & Ball, Beach ball \\
Mats \& Foam & Yoga mat, Foam mat, Foam, Noodle \\
Animals \& Toys & Kangaroo, Panda, Model animal, Road kill, RC car, \\
               & Model car, Toy, Donuts, Banana \\
Umbrellas & Umbrella, Open umbrella \\
Vegetation & Tree branch, Branch, Foliage, Leaves \\
Misc.\ Debris & Cardboard, Debris, Toolbox \\
\bottomrule
\end{tabular}

\label{tab:object-list}
\end{table}

\subsubsection{Baselines}
We benchmark against several state-of-the-art open-vocabulary segmentation methods: GroundedSAM \cite{groundedsam24} employs an open-set object detector, Grounding DINO \cite{groundingdino2024eccv}, with various text prompts to query SAM \cite{Kirillov_2023_ICCV}. Similarly, we introduce an OWLv2+SAM baseline that replaces the Grounding DINO detector with OWLv2 \cite{OWLv2neurips23}. These methods follow a \textit{detect-then-segment} paradigm. Conversely, we also evaluate \textit{segment-then-classify} approaches, where the image is first partitioned into masks which are subsequently classified. We utilize the state-of-the-art MaskCLIP++ \cite{maskclippp25} framework, which we combine with various mask predictors. Specifically, we evaluate mask predictors from FC-CLIP \cite{FCCLIP_NIPS23} and MAFT+ \cite{MAFTp_ECCV2024} and SAM. We also compare against more  heuristic-based ground segmentation models, such as Patchwork++ \cite{patchworkpp}, which uses estimated LiDAR elevation to compare against what is and what isn't ground.

For the second set of experiments, we also train \ourmodel{}'s segmentation objective ($\mathcal{L}_{\text{seg}}$) using each of the pseudo-labels from the baselines. The goal is to determine how well this information can be distilled into a downstream world model, and in turn how well this model generalizes when evaluated on a different domain. Notably, our model under GroundedSAM supervision can be viewed as an extension of the recent  work QueryOcc+ \cite{QueryOcc_2025}, albeit with a distinct architecture and additional LiDAR input. Finally, we also add baselines training the LiDAR-only based models DIO \cite{Diehl_2025_DIO} and UnO \cite{Agro_2024_CVPR} with the pseudo-labels from our method (\ourimagelabels{}, \ourcombinedlabels{}).

\subsubsection{Text Prompts.}
For all open-vocabulary methods, we utilize a consistent set of prompts for ground ($\mathcal{P}_{\text{ground}}$) and obstacle ($\mathcal{P}_{\text{obst}}$) classes based on \cite{sal2024eccv}. 
Simply put, we start with those prompts and then map them to either the obstacle or the ground class as shown in \Cref{tab:text_prompts_sal}. 
Following common practices, the baselines also wrap each text prompt into a full sentence, resulting in prompts like \textit{a photo of a pedestrian}. 
The MaskCLIP++ (MCPP) baselines first generate a set of mask, which are later classified. To avoid assigning regions like the sky to either ground or obstacle, we prompt those models additionally with the prompt \textit{sky} and map this later to an ignore class, which enables a more fair comparison.

Finally, recognizing that a closed-set of predefined prompts are inherently limited and risks omitting rare object classes, to establish an upper bound for open-vocabulary performance, we include an \textbf{(Oracle)} variant where scenes are manually inspected and specific unknown object classes are appended to $\mathcal{P}_{\text{obstacle}}$.  We show these prompts in \Cref{tab:object-list}.

\definecolor{groundblue}{RGB}{31,119,180}
\definecolor{otherorange}{RGB}{255,127,14}

\subsection{LiDAR Forecasting (AV2)}

Following the Argoverse 2 leaderboard, we report the following four metrics using the same evaluation protocol:
\begin{itemize}
    \item \textbf{L1} [$\SI{}{\metre}$] The average absolute error between the depth predicted by the occupancy model and the ground truth LiDAR ray depth. This is the primary metric for the AV2 leaderboard rankings.
    \item \textbf{AbsRel} [$\%$] The absolute error of the LiDAR ray normalized by the ground truth depth. 
    \item \textbf{CD} [$\SI{}{\metre}^2$]: The Chamfer Distance between the predicted LiDAR point cloud and the ground truth. It is defined as the mean squared distance from every point in the predicted point cloud to its nearest neighbor in the true point cloud.
    \item \textbf{NFCD} [$\SI{}{\metre}^2$]: The Near Field Chamfer Distance, which measures the Chamfer distance between the predicted and ground truth point clouds within a restricted 70m$ \times$ 70m range.
\end{itemize}

\subsection{Downstream Occupancy Fine-tuning (AV2/LRR)}

On AV2, the evaluation ROI is same as training except only 25m to the sides due to memory constraints and at a resolution of \SI{0.5}{\metre} every \SI{0.1}{\second}.
LRR uses the same evaluation ROI $(x,y)$ as described in \Cref{sec:supp_exp_setupocc_world_model_pretraining}. We evaluate on a grid of query points with resolution \SI{0.3125}{\metre} every \SI{0.3}{\second}.

\subsubsection{Metrics.} The fine-tuning experiments for BEV/3D semantic occupancy/flow forecasting use the following metrics:  
\begin{itemize}
    \item \textbf{mAP} [$\%$]: This mean average precision metric is the average area under the precision recall curve when the predicted occupancy is evaluated against the rasterized bounding box labels. This can be done in both 2D BEV and in 3D. 

    \item \textbf{IoU} [$\%$]: Soft Intersection Over Union is a metric originally proposed by \cite{mahjourian2022occupancy} as a more calibrated soft evaluation method for occupancy. It is the probabilistic equivalent of IoU, previously introduced. 

    \item \textbf{EPE} [$\%$]: End point error represents the $L_2$ distance between the ground truth flow vector, and the predicted flow vector. This can be computed in 2D for 2D BEV models, and in 3D. 
\end{itemize}

\section{Additional Experiments}

\subsection{Quantitative Results}

\subsubsection{Runtime and Memory.} In \Cref{tab:runtime_memory}, we provide the runtime and memory usage for \ourmodel{} and the baselines on an NVIDIA RTX 5000 GPU. We also report the number of parameters for each method. We distinguish between two settings: The first setting (eval) is evaluated on a full grid of query points $(x,y,t)$, totaling 2,048,000 points; this setting is also used to compute our metrics. We observe that UnO and DIO achieve lower runtime and memory usage, which can be attributed to the addition of the new camera and RADAR sensors in \ourmodel{}. Specifically, the camera encoding contributes to an increased parameter count, motivating the need for more efficient architectures in future work. In the second setting, we query the model with 20,000 points and measure the resulting runtime and memory. This is motivated by the number of query points used by a state-of-the-art planner, as reported by QuAD \cite{QuAD2024}. We observe a significant reduction in runtime across all methods when transitioning from the grid (eval: 2,048k) to the planner (20k) query setting. Notably, TriO (no camera) achieves the best runtime, rendering the method real-time capable.

\begin{table*}[th]
    \centering
    \footnotesize
    \setlength\tabcolsep{3pt}
    \caption{Model parameters, runtime, and GPU memory for two query-point settings (Planner~\cite{QuAD2024}: 20k points; Eval: 2048k points), and fine-tuning downstream performance on \nameourdataset{}. }
    \label{tab:runtime_memory}
    \resizebox{\linewidth}{!}{%
    \begin{tabular}{@{}l ccc ccc r r r r@{}}
    \toprule
     & \multicolumn{3}{c}{All} & \multicolumn{3}{c}{200m+} & & & \multicolumn{1}{c}{Planner~\cite{QuAD2024}} & \multicolumn{1}{c}{Eval} \\
    \cmidrule(lr){2-4} \cmidrule(lr){5-7} \cmidrule(lr){10-10} \cmidrule(lr){11-11}
    Method & mAP $\uparrow$ & Soft-IoU $\uparrow$ & EPE $\downarrow$ & mAP $\uparrow$ & Soft-IoU $\uparrow$ & EPE $\downarrow$ & \# Params & Mem (MB) $\downarrow$ & Runtime (ms) $\downarrow$ & Runtime (ms) $\downarrow$ \\
    \midrule
    UnO       & 51.6 & 21.2 & 4.92 & 27.2 & 8.30 & 8.29 & 39,742,221 & \textbf{2,493} & 63.92 & 1774.41 \\
    DIO       & 58.0 & 31.5 & 3.01 & 36.2 & 13.8 & 4.40 & 8,205,109 & 2,930 & 68.21 & 1776.86 \\
    \rowcolor{tablecolor}
    \ourmodel{} (no camera) & \textbf{60.5} & \textbf{35.7} & \textbf{2.68} & 42.2 & 20.9 & \textbf{3.64} & 12,930,483 & 3,449 & \textbf{60.81} & \textbf{1622.54} \\
    \rowcolor{tablecolor}
    \ourmodel{} (no radar)  & 57.7 & 31.9 & 3.04 & 34.5 & 14.3 & 4.59 & 54,224,584 & 9,907 & 488.06 & 2048.88 \\
    \rowcolor{tablecolor}
    \ourmodel{} & \textbf{60.5} & \textbf{35.7} & 2.86 & \textbf{42.6} & \textbf{21.4} & 4.01 & 54,224,967 & 9,915 & 489.32 & 2054.63 \\
    \bottomrule
    \end{tabular}}
\end{table*}

\subsubsection{Additional Camera Ablations}
\label{sec:cam_ablations}
\cref{tab:4cam_comparison} shows marginal gains from 4 cameras on AV2 in occupancy, flow, and LiDAR metrics. However, these were not enough to justify the added cost during our main experiments, which aligns with our no-camera ablations in the main paper. 

Expanding on our no camera ablation study in the main paper on LRR, we find camera inputs increases occupancy performance at range lending to the idea that camera features are more dense than other sensor modalities such as LiDAR at these distances. Howeve, since we only pass in one timestep of camera data as input to the model, we hypothesize this may be responsible for the slight regression in flow estimation since a single timestep of camera is insufficient to estimate flow by itself. The improvement in flow in AV2, contrasting with the decline seen in the LRR "no-camera" ablation, is likely because AV2 (urban) contains significantly more static vehicles. In contrast, LRR (highway/urban) includes many high-velocity objects for which single-frame camera input is less informative. Furthermore, LiDAR provides a dominant signal for geometric labels (occupancy, obstacle segmentation) except at long range, where sparsity increases; this aligns with the observed occupancy improvements at range when incorporating camera data. Future work could explore camera pre-training to better leverage multi-modal features.

\begin{table}[h]
\centering
\small
\caption{Camera Ablation on AV2} \label{tab:4cam_comparison}
\resizebox{\columnwidth}{!}{
\begin{tabular}{@{}lccccccc@{}}
\toprule
& \multicolumn{3}{c}{4D Semantic Occupancy} & \multicolumn{4}{c}{LiDAR Forecasting} \\
\cmidrule(lr){2-4} \cmidrule(lr){5-8}
Model & mAP (\%) $\uparrow$ & SoftIoU (\%) $\uparrow$ & EPE (m) $\downarrow$ & L1 (m) $\downarrow$ & AbsRel (\%) $\downarrow$ & NFCD (m$^2$) $\downarrow$ & CD (m$^2$) $\downarrow$ \\
\midrule
TRiO (2 Cam) & \textbf{63.3} & 31.9 & 1.97 & \textbf{1.82} & 10.63 & 0.93 & \textbf{10.90} \\
TRiO (4 Cam) & \textbf{63.3} & \textbf{32.3} & \textbf{1.91} & \textbf{1.82} & \textbf{10.48} & \textbf{0.90} & 11.06 \\
\bottomrule
\end{tabular}
}
\end{table}

\subsubsection{Unsupervised Flow Estimation}
\label{supp:experiments_unsupervised_flow}
In \cref{tab:unsupervised_flow}, we compare the unsupervised flow estimation of TriO against that of DIO \cite{Diehl_2025_DIO}, which uses a warping consistency loss.
We observe that our method, directly supervised with RADAR, provides the best unsupervised flow estimation. We find that DIO only learns a minimal amount of flow on the surface of objects, leaving the occupancy in the interior stationary. This is also reflected in the EPE (static) metric, where DIO performs well due to predicting zero flow. As a result, our RADAR supervised flow results in significantly more accurate predictions, especially for dynamic obstacles as shown by the EPE (dynamic).

\begin{table*}[ht]
    \centering
    \footnotesize
    \setlength\tabcolsep{2pt}
    \caption{Unsupervised flow forecasting results on \nameourdataset{}.}
    \label{tab:unsupervised_flow}
    \resizebox{\linewidth}{!}{%
    \begin{tabular}{@{}l ccc ccc@{}}
    \toprule
    \cmidrule(lr){2-7}
     & \multicolumn{3}{c}{All} & \multicolumn{3}{c}{200m+} \\
    \cmidrule(lr){2-4} \cmidrule(lr){5-7}
    Method & EPE $\downarrow$ & EPE (static) $\downarrow$ & EPE (dynamic) $\downarrow$ & EPE $\downarrow$ & EPE (static) $\downarrow$ & EPE (dynamic) $\downarrow$ \\
    \midrule
    DIO & 27.8 & \textbf{0.01} & 28.8 & 28.4 & \textbf{0.01} & 29.3 \\
    \rowcolor{tablecolor}
    \ourmodel{} & \textbf{19.4} & 0.17 & \textbf{20.3} & \textbf{19.6} & 0.16 & \textbf{20.5} \\
    \bottomrule
    \end{tabular}}
\end{table*}

\subsubsection{Breakdown of AV2 Prediction Metrics}
\cref{tab:lidar_forecasting_flat} shows a metrics breakdown in near and far field. We contacted the leaderboard creators and confirmed the published L1 and AbsRel metrics are implicitly near field.  
\cref{tab:lidar_forecasting_flat} shows our re-implemented UnO and DIO baselines, have slightly different metrics (better L1, worse CD) than the results from the leaderboard in the main paper, but in general we do better close by and worse at range. 
\definecolor{lbblue}{RGB}{130,180,230}
\newcommand{\lb}[1]{{\color{lbblue}#1}}

\vspace{-0.3cm}
\begin{table}[h]
\centering
\small
\caption{AV2 LiDAR forecasting. \lb{Leaderboard metrics}.} \label{tab:lidar_forecasting_flat}
\resizebox{\columnwidth}{!}{
\begin{tabular}{@{}lccccccc@{}}
\toprule
Model & \lb{NF-L1} $\downarrow$ & FF-L1 $\downarrow$ & \lb{NF-AbsRel (\%)} $\downarrow$ & FF-AbsRel (\%) $\downarrow$ & \lb{NFCD} $\downarrow$ & FFCD $\downarrow$ & \lb{CD} $\downarrow$ \\
\midrule
UnO   & 2.07 & \textbf{2.07} & 11.69 & \textbf{3.01} & 1.03 & 79.28 & 11.41 \\
DIO   & 1.87 & 2.20 & 11.28 & 3.28 & 0.97 & 87.07 & 12.43 \\
TRiO  & \textbf{1.82} & 2.13 & \textbf{10.63} & 3.31 & \textbf{0.93} & \textbf{76.52} & \textbf{10.90} \\
\bottomrule
\end{tabular}
}
\end{table}

\subsubsection{Sensitivity Analysis}
\begin{figure}[h]
\begin{tikzpicture}
\begin{axis}[
    xlabel={$\alpha_{\text{fusion}}$},
    ylabel={Score (\%)},
    xmin=0.05, xmax=0.95,
    ymin=30, ymax=92,
    xtick={0.1, 0.2, 0.3, 0.4, 0.5, 0.6, 0.7, 0.8, 0.9},
    ytick={40, 60, 80},
    ymajorgrids=true,
    grid style={gray!30},
    width=\columnwidth,
    height=5.0cm,
    tick label style={font=\scriptsize},
    label style={font=\small},
    legend columns=2,
    legend style={
        at={(0.68,0.60)},
        anchor=west,
        font=\tiny,
        fill=white, fill opacity=0, text opacity=1,
        draw=none,
        column sep=4pt,
        row sep=-2pt,
        /tikz/column 2/.style={column sep=4pt},
    },
    legend cell align={left},
    legend image post style={xscale=0.5},
    every axis plot/.append style={thick},
]

\addplot[color=blue, mark=square*, mark size=1.5pt] coordinates {
    (0.1, 39.7) (0.2, 40.6) (0.3, 41.7) (0.4, 43.2) (0.5, 44.6) 
    (0.6, 45.8) (0.7, 47.1) (0.8, 45.4) (0.9, 45.3)
};
\addlegendentry{mIOU}

\addplot[color=red!80!black, mark=triangle*, mark size=2pt] coordinates {
    (0.1, 56.6) (0.2, 57.2) (0.3, 57.6) (0.4, 58.0) (0.5, 57.8) 
    (0.6, 56.5) (0.7, 52.7) (0.8, 41.5) (0.9, 39.1)
};
\addlegendentry{F1}

\addplot[color=green!60!black, mark=diamond*, mark size=2pt] coordinates {
    (0.1, 71.1) (0.2, 73.0) (0.3, 75.0) (0.4, 77.5) (0.5, 79.9) 
    (0.6, 82.1) (0.7, 84.8) (0.8, 84.5) (0.9, 84.5)
};
\addlegendentry{Recall}

\end{axis}
\end{tikzpicture}
\caption{Sensitivity to fusion parameter $\alpha_{\text{fusion}}$ (STU)}. 
\label{fig:sensitivity_fusion}
\end{figure}
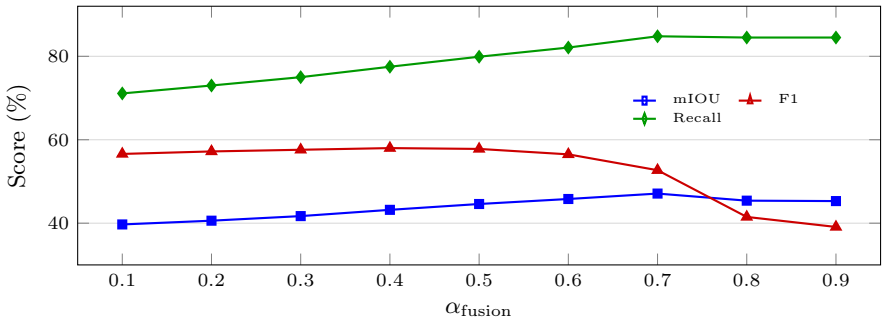

\cref{fig:sensitivity_fusion} shows the current 0.5 threshold is a good balance between recall and F1 score. The method is robust balance across all metrics; performance only degrades at extreme values. 
We also perform loss sensitivity analysis on LRR. Scaling each loss weight by $4\times$ and $0.25\times$ shows that TriO is robust to this, with min/max deviations of $0.1/0.2\%$ AP, $0.0/0.4\%$ SoftIoU, $0.03/0.44$\,m EPE. 

\subsection{Qualitative Results}
We provide additional qualitative visualizations to highlight different aspects of \ourmodel{} and the baselines. \Cref{fig:supp_av2_qual,fig:supp_av2_qual_2} show diverse AV2 scenes, illustrating the model's generalization across a wide range of scenarios and obstacle types, including VRUs, construction vehicles, animals, infrastructure, and challenging weather conditions like snow. 

\Cref{fig:supp_seg_qual,fig:supp_seg_qual_2} compare our pseudo-labels against several open-vocabulary VLM baselines on a broad range of uncommon obstacles, emphasizing both the strong generalization of our model and representative failure cases of current state-of-the-art VLMs, even when supplied with 'ground-truth' text prompts (Oracle).

\Cref{fig:stu_rollout_1,fig:stu_rollout_2} present additional zero-shot results of our model, trained on AV2, via roll-outs on STU that showcase obstacle segmentation for rare objects. In addition \Cref{fig:av2_rollout_vru} shows an AV2 rollout containing a vulnerable road user on a scooter. \Cref{fig:supp_lidar} visualizes forecasted LiDAR vs ground-truth LiDAR on a challenging scenario with different traffic participants.  \Cref{fig:av2_failure_cases,fig:stu_failure_case} summarize typical failure modes on AV2 and STU, respectively.

\begin{figure}[h]
    \centering
    \begin{tikzpicture}
        \pgfmathsetlengthmacro{\imw}{\columnwidth / 4 - 3pt}
        \pgfmathsetlengthmacro{\xstep}{\imw + 3pt}

        \node[inner sep=0pt, outer sep=0, anchor=north west] (img2) at (0,0) {\includegraphics[width=\imw, height=2cm, keepaspectratio]{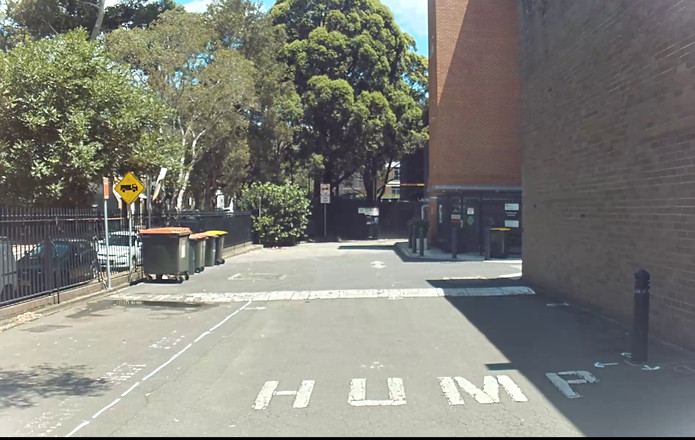}};
        \node[inner sep=0pt, outer sep=0, anchor=north west] (seg2) at (\xstep,0) {\includegraphics[width=\imw, height=2cm, keepaspectratio]{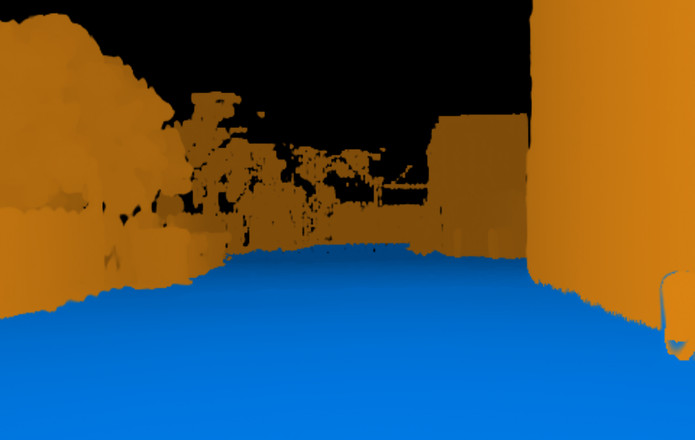}};
        \node[inner sep=0pt, outer sep=0, anchor=north west] (img3) at (2*\xstep,0) {\includegraphics[width=\imw, height=2cm, keepaspectratio]{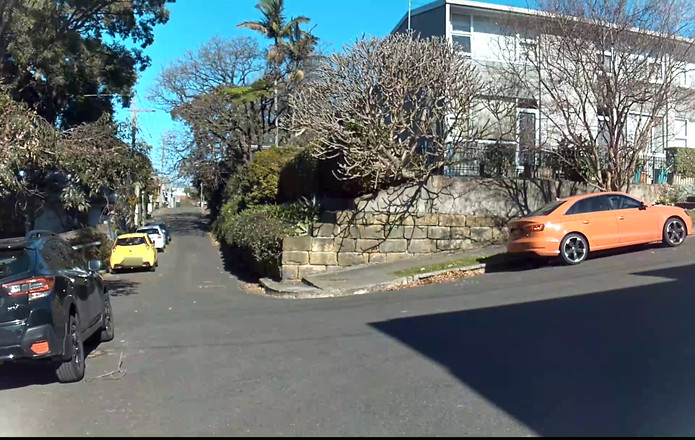}};
        \node[inner sep=0pt, outer sep=0, anchor=north west] (seg3) at (3*\xstep,0) {\includegraphics[width=\imw, height=2cm, keepaspectratio]{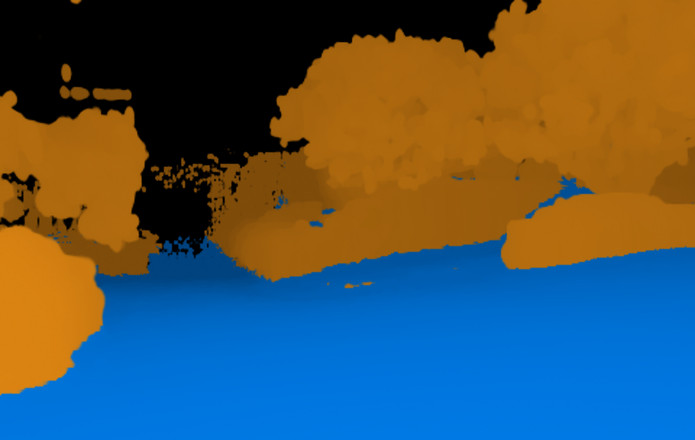}};

    \end{tikzpicture}
    \vspace{-0.2cm}
    \caption{
        TriO inference results for 1. speed bump, 2. hilly roads.
    }
    \label{fig:add_visualisations}
\end{figure}

Meanwhile, \Cref{fig:add_visualisations} showcase how our model successfully extends to edge cases such as speed bumps and hilly roads.

\section{Extended Limitations and Future Work}
For obstacle segmentation pseudo-labels, we use a late-fusion approach that combines image- and LiDAR-based segmentation, where pseudo-labels are first generated independently for each modality. Since both segmentation approaches rely primarily on surface normals, future work could explore unified early-fusion methods such as \cite{keetha2026mapanything}. Although including cameras as an input improves long-range occupancy performance, it also increases runtime and memory usage. Future work should therefore investigate more efficient camera encoding approaches.

While \ourmodel{} also generalizes to other weather conditions, for example as shown in \Cref{fig:supp_av2_qual}, we sometimes observe false-positive occupancy and obstacle segmentation predictions caused by snow and exhaust (\Cref{fig:av2_failure_cases}). Incorporating more occupancy self-supervision from cameras and RADAR could help mitigate these failures.
We acknowledge that open-loop metrics may not fully capture final closed-loop driving performance. Although this is beyond the scope of the present paper, we are interested in studying how \ourmodel{} can be used directly in downstream motion planning tasks, since the ability to detect and avoid rarely seen objects is vital for self-driving.

\begin{figure}[t]
    \centering
    \begin{tikzpicture}
        \pgfmathsetlengthmacro{\imw}{\columnwidth / 2 - 3pt}
        \pgfmathsetlengthmacro{\xstep}{\imw + 3pt}
        
        \node[inner sep=0pt, outer sep=0, anchor=north west] (img2) at (0,0) {\includegraphics[width=\imw, height=2cm, keepaspectratio]{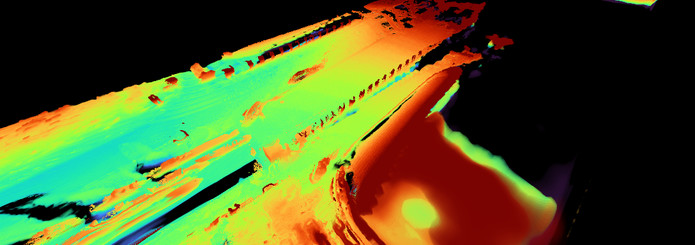}};
        \node[inner sep=0pt, outer sep=0, anchor=north west] (seg2) at (\xstep,0) {\includegraphics[width=\imw, height=2cm, keepaspectratio]{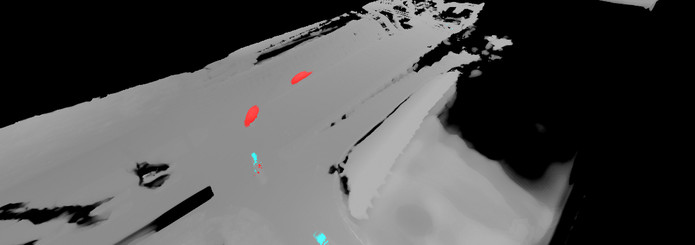}};

        \node[anchor=north west] at (img2.north west) {\textbf{(a)}};
        \node[anchor=north west] at (seg2.north west) {\textbf{(b)}};
        
    \end{tikzpicture}
    \vspace{-0.2cm}
    \caption{
a) Warped occupancy by flow (hard scenario). b) Flow degeneracy: longitudinal bias failing to resolve orthogonal motion.
    }
    \vspace{-0.3cm}
    \label{fig:flow_viz}
\end{figure}

Another area for future work is improving upon the flow representation. 
Forecasted occupancy warped via 3D flow (\cref{fig:flow_viz} a) reveals inconsistencies between both. While RADAR flow is a strong pretraining signal (Tab. 3) outperforming consistency-baselines (Tab. 6, Supp.), it is insufficient for standalone metric flow.
TriO resolves 3D flow from 1D radial Doppler with spatially and temporally diverse measurements: Multiple RADAR sensors capture distinct radial components of the same velocity vector from different viewpoints, resolving the full 3D flow via geometric triangulation. A single object can also yield multiple returns across its spatial extent, and we supervise with future sensor frames, making learning only radial flow non-trivial. We hypothesize that the location invariance in the CNN backbone also helps. However, tangential failures (\cref{fig:flow_viz} b) persist, likely due to highway motion biases.

\clearpage
\newpage 

\input{figures/supp_av2_composite}

\input{figures/supp_seg_composite}

\input{figures/supp_stu_rollout_1}

\input{figures/supp_stu_rollout_2}

\input{figures/supp_av2_rollout_vru}

\input{figures/supp_lidar_render}

\input{figures/supp_av2_failure_cases}

\input{figures/supp_stu_failure_case}

\end{document}